\documentclass[acmsmall,nonacm]{acmart}

\AtBeginDocument{%
  \providecommand\BibTeX{{%
    \normalfont B\kern-0.5em{\scshape i\kern-0.25em b}\kern-0.8em\TeX}}}

\acmJournal{JACM}

\usepackage{color}
\usepackage{xcolor}
\usepackage{amsthm}
\usepackage{ulem}
 
\usepackage{amsmath,amssymb,amsfonts}
\usepackage{pifont}

\usepackage{subfigure}
\usepackage{graphicx}
\usepackage{textcomp}
\usepackage{setspace}
\usepackage{mathtools}
\usepackage{cases}
\usepackage{subeqnarray}
\usepackage{makecell}
\usepackage{subfigure}
\usepackage{multirow}
\useunder{\uline}{\ul}{}
\usepackage{makecell}
\usepackage{adjustbox}

\usepackage{tcolorbox}
\definecolor{color1}{rgb}{0.1791464821222607, 0.49287197231833907, 0.7354248366013072}
\definecolor{crimson}{rgb}{0.0, 0.0, 0.0}

\usepackage{wrapfig}
\usepackage[edges]{forest}
\usepackage{tikz}

\usepackage{booktabs}
\usepackage[ruled,linesnumbered,noend]{algorithm2e}

\begin{document}

\title{Efficient Deep Learning Infrastructures for Embedded Computing Systems: A Comprehensive Survey and Future Envision}

\author{Xiangzhong Luo}
\email{xiangzho001@e.ntu.edu.sg}
\affiliation{%
  \institution{Nanyang Technological University}
  \country{Singapore}
}

\author{Di Liu}
\affiliation{%
  \institution{Norwegian University of Science and Technology}
  \country{Norway}
}
\email{di.liu@ntnu.no}

\author{Hao Kong}
\affiliation{%
 \institution{Nanyang Technological University}
 \country{Singapore}
}
\email{kong.hao@ntu.edu.sg}

\author{Shuo Huai}
\affiliation{%
 \institution{Nanyang Technological University}
 \country{Singapore}
}
\email{huai.shuo@ntu.edu.sg}

\author{Hui Chen}
\affiliation{%
    \institution{Nanyang Technological University}
    \country{Singapore}
}
\email{chen.hui@ntu.edu.sg}

\author{Guochu Xiong}
\affiliation{%
 \institution{Nanyang Technological University}
 \country{Singapore}
}
\email{guochu.xiong@ntu.edu.sg}

\author{Weichen Liu}
\authornote{The corresponding author is Weichen Liu (Email: liu@ntu.edu.sg). \\
This research is partially supported by the Ministry of Education, Singapore, under its Academic Research Fund Tier 1 (RG94/23), and partially supported by Nanyang Technological University, Singapore, under its NAP (M4082282/04INS000515C130).}
\affiliation{%
 \institution{Nanyang Technological University}
 \country{Singapore}
}
\email{liu@ntu.edu.sg}

\renewcommand{\shortauthors}{}

\begin{abstract}

Deep neural networks (DNNs) have recently achieved impressive success across a wide range of real-world vision and language processing tasks, spanning from image classification to many other downstream vision tasks, such as object detection, tracking, and segmentation. However, previous well-established DNNs, despite being able to maintain superior accuracy, have also been evolving to be deeper and wider and thus inevitably necessitate prohibitive computational resources for both training and inference. This trend further enlarges the computational gap between computation-intensive DNNs and resource-constrained embedded computing systems, making it challenging to deploy powerful DNNs upon real-world embedded computing systems towards ubiquitous embedded intelligence. To alleviate the above computational gap and enable ubiquitous embedded intelligence, we, in this survey, focus on discussing recent efficient deep learning infrastructures for embedded computing systems, spanning \textbf{from training to inference}, \textbf{from manual to automated}, \textbf{from convolutional neural networks to transformers}, \textbf{from transformers to vision transformers}, \textbf{from vision models to large language models}, \textbf{from software to hardware}, and \textbf{from algorithms to applications}. Specifically, we discuss recent efficient deep learning infrastructures for embedded computing systems from the lens of (1) efficient manual network design for embedded computing systems, (2) efficient automated network design for embedded computing systems, (3) efficient network compression for embedded computing systems, (4) efficient on-device learning for embedded computing systems, (5) efficient large language models for embedded computing systems, (6) efficient deep learning software and hardware for embedded computing systems, and (7) efficient intelligent applications for embedded computing systems. Furthermore, we also envision promising future directions and trends, which have the potential to deliver more ubiquitous embedded intelligence. We believe this survey has its merits and can shed light on future research, which can largely benefit researchers to quickly and smoothly get started in this emerging field.

\end{abstract}



\maketitle
\pagestyle{plain}

\section{Introduction}
\label{sec:introduction}

With the increasing availability of large-scale datasets and advanced computing paradigms, deep neural networks (DNNs)\footnote{In this work, we may interchangeably use some technical terms, such as deep learning models, machine learning models, DL models, ML models, deep neural networks (DNNs), and convolutional neural networks (CNNs).} have empowered a wide range of intelligent applications and have demonstrated strong performance \cite{simonyan2014vggnet, he2016deep, huang2017densely}. These intelligent applications may span from image classification \cite{he2016deep} to downstream vision tasks, such as object detection \cite{liu2016ssd}, tracking \cite{kristan2015visual}, and segmentation \cite{li2014secrets}, to natural language processing (NLP) tasks, such as automatic speech recognition \cite{yu2016automatic}, machine translation \cite{wu2016google}, and question answering \cite{choi2018quac}. In the subsequent years, deep neural networks have been evolving deeper and deeper with more and more layers in order to maintain state-of-the-art accuracy on target task \cite{simonyan2014vggnet, he2016deep, huang2017densely}. In the meantime, novel network structures and advanced training techniques have also emerged, which further push forward the attainable accuracy \cite{xie2017aggregated, hinton2015distilling, zhang2017mixup}. These powerful deep learning (DL) networks and advanced training techniques, starting from VGGNet \cite{simonyan2014vggnet} and ResNet \cite{he2016deep}, mark the emergence of the deep learning era.

The tremendous breakthroughs of DNNs have subsequently attracted a huge amount of attention from both academia and industry to deploy powerful DNNs upon real-world embedded computing systems, including mobile phones \cite{ignatov2018ai, tan2021deep}, autonomous vehicles \cite{kisavcanin2017deep, fayyad2020deep}, and healthcare \cite{norgeot2019call, esteva2019guide}, to enable intelligent embedded applications towards embedded intelligence \cite{wu2019machine}. In practice, this may bring significant benefits. For example, embedded computing systems explicitly allow real-time on-device data processing, which significantly improves the processing efficiency and thus delivers enhanced user experience. This also protects data security and privacy since everything can be locally processed without being uploaded to the remote server \cite{wu2019machine}. Despite the above promising benefits, deploying powerful DNNs upon real-world embedded computing systems still suffers from several critical limitations. On the one hand, in order to maintain competitive accuracy, recent representative networks have been evolving deeper and deeper with hundreds of layers \cite{he2016deep, huang2017densely}, and as a result, lead to prohibitive computational complexity \cite{wu2019machine, liu2022bringing, huai2023hardware}. For example, ResNet50 \cite{he2016deep}, as one of the most representative deep networks, consists of over 4 billion floating-point operations (FLOPs) and 25 million parameters, which requires over 87\,MB on-device storage to deal with one single input image. On the other hand, real-world embedded computing systems like mobile phones and autonomous vehicles typically feature limited available computational resources in order to optimize the on-device power and energy consumption. In sight of the above, the evolving network complexity continues to enlarge the computational gap between computation-intensive deep neural networks and resource-constrained embedded computing systems \cite{liu2024edge}, inevitably making it increasingly challenging to embrace ubiquitous embedded intelligence.

\begin{figure}[t]
    \begin{center}
    \includegraphics[width=1.0\columnwidth]{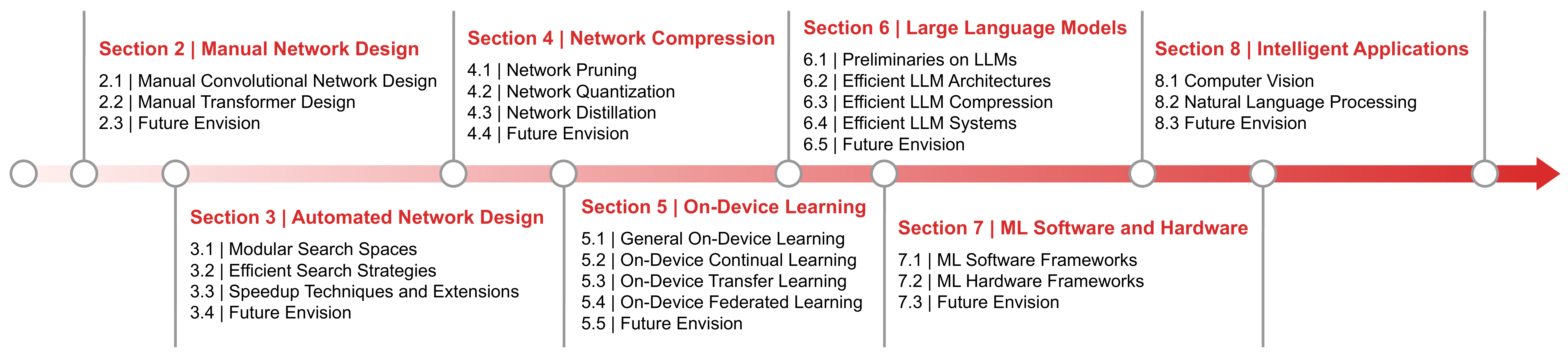}
    \end{center}
    \vspace{-5pt}
    \caption{The organization of this paper, in which we ignore Section~\ref{sec:introduction} and Section~\ref{sec:conclusion} for the sake of simplicity.}
    \vspace{-5pt}
    \label{fig:survey-structure}
\end{figure}

To bridge the aforementioned computational gap towards ubiquitous embedded intelligence, a plethora of model compression techniques have been recently proposed, including network pruning \cite{li2016pruning, he2018soft, he2019filter}, network quantization \cite{courbariaux2015binaryconnect, hubara2016binarized, rastegari2016xnor}, and network distillation \cite{ba2014deep, hinton2015distilling, romero2014fitnets}, which strive for better accuracy-efficiency trade-offs to accommodate the limited available computational resources in real-world embedded scenarios. For example, network pruning focuses on removing the redundancy network units, such as weights \cite{han2015learning}, channels \cite{li2016pruning}, and layers \cite{jordao2020discriminative}, to trim down the network redundancy, which can boost the efficiency on target hardware with minimal accuracy loss on target task. In addition to network compression, another parallel alternative is to manually design resource-efficient networks instead, such as SqueezeNet \cite{hu2018squeeze}, MobileNets \cite{howard2017mobilenets, molchanov2022lana}, ShuffleNets \cite{ma2018shufflenet, zhang2018shufflenet}, and GhostNets \cite{han2020ghostnet, tang2022ghostnetv2}, which have dominated the early progress from the lens of efficient network design. These efficient networks, despite being able to exhibit superior efficiency, highly rely on human expertise to explore novel network structures through trial and error, which also involve non-trivial engineering efforts and prohibitive computational resources \cite{tan2019mnasnet, wu2019fbnet, cai2019proxylessnas}. To overcome such limitations, recent network design practices have shifted from manual to automated, also referred to as neural architecture search (NAS) or automated machine learning (AutoML), which focuses on automatically exploring novel network structures \cite{white2023neural}. The tremendous success of NAS has subsequently sparked rich hardware-aware NAS works, such as MnasNet \cite{tan2019mnasnet}, ProxylessNAS \cite{cai2019proxylessnas}, FBNet \cite{wu2019fbnet}, and Once-for-All \cite{cai2020once}, to automate the design of accurate yet hardware-efficient network solutions, which have shown strong accuracy-efficiency trade-offs and have been widely upon real-world embedded computing systems to deliver intelligent services \cite{benmeziane2021comprehensive}.

Apart from the above efficient networks and techniques that typically focus on improving the on-device inference efficiency, recent research also turns back to the on-device training efficiency \cite{cai2020tinytl, lin2022device}. The rationale here is that previous representative networks, despite being able to exhibit superior accuracy, have to be trained for hundreds of epochs, which may require multiple days on powerful GPUs \cite{cai2020tinytl}. And even worse, the expensive training process on remote GPUs does not allow on-device customization on local hardware, especially in resource-constrained embedded scenarios \cite{lin2022device}. Note that local on-device customization has the potential to further improve the attainable accuracy using newly collected data since local sensors continue to collect new data from users over time. To overcome such limitations, several efficient on-device learning techniques have been recently established, such as on-device continual learning \cite{van2019three}, on-device transfer learning \cite{cai2020tinytl}, and on-device federated learning \cite{qiu2022zerofl}, making it possible to train and fine-tune powerful deep networks on local hardware for further performance improvement.

More recently, large language models (LLMs), such as GPT-3 \cite{brown2020language} and GPT-4 \cite{openai2023gpt4}, have demonstrated impressive success across various real-world language processing tasks \cite{bai2024beyond}. However, the strong learning capability of these powerful LLMs also comes at the cost of excessive computational complexity. For example, OpenAI's GPT-3 \cite{brown2020language}, as one of the most representative LLMs, consists of 175 billion parameters. Furthermore, in order to achieve state-of-the-art performance, recent LLMs continue to evolve to be larger and larger with ever-increasing model sizes \cite{chowdhery2023palm, le2023bloom}. These make it increasingly challenging to deploy recent powerful LLMs on modern embedded computing systems towards intelligent language processing services. To overcome such limitations, a series of effective techniques have been recently proposed, which focus on alleviating the prohibitive computational complexity of LLMs to explore computation-efficient LLMs, including efficient LLM architecture design \cite{kwon2023efficient, dao2022flashattention, dao2023flashattention, xiao2023efficient}, efficient LLM compression techniques (i.e., pruning \cite{ma2023llm, sun2023simple}, quantization \cite{xiao2023smoothquant, lin2023awq}, and knowledge distillation \cite{wang2022self, gu2023minillm}), and efficient LLM system design \cite{sheng2023flexgen, borzunov2022petals, wang2023tabi}.

In parallel to the booming emergence of powerful deep networks and advanced training techniques, a plethora of representative deep learning software frameworks and hardware accelerators have been tailored to facilitate the development of efficient deep learning solutions for embedded computing systems, such as TensorFlow \cite{tensorflow2015-whitepaper}, PyTorch \cite{paszke2019pytorch}, Google edge TPUs \cite{google-edgetpu}, Nvidia edge GPUs \cite{nvidia-jetson}, and Intel Neural Compute Stick \cite{intel-ncs}. These deep learning software and hardware have been extensively adopted in the deep learning era and bring two main benefits. On the one hand, these deep learning software and hardware lift the roadblock for both software and hardware engineers and thus allow them to quickly develop intelligent embedded applications, such as on-device object detection \cite{liu2016ssd}, tracking \cite{kristan2015visual}, and segmentation \cite{li2014secrets}, with less domain-specific expertise. On the other hand, these deep learning software and hardware typically feature domain-specific optimization and thus can achieve superior accuracy-efficiency trade-offs with minimal engineering efforts. For example, Nvidia Jetson AGX Xavier, as one representative Nvidia Jetson edge GPU, supports the development of intelligent embedded applications with the precision of INT8 (i.e., 8-bit weights), which can deliver significant efficiency improvement over its full-precision counterpart (32-bit weights) without degrading the accuracy on target task \cite{nvidia-jetson}.

\subsection{Organization of This Paper}
\label{sec:organization-of-this-paper}

In this survey, we focus on summarizing recent efficient deep learning infrastructures that may benefit current and future embedded computing systems towards ubiquitous embedded intelligence. In practice, some existing surveys \cite{menghani2023efficient, cheng2017survey, choudhary2020comprehensive, li2023model} typically focus on efficient deep learning algorithms, which, however, may be out-of-date since recent deep learning infrastructures have been rapidly evolving, especially from the perspective of large language models. In contrast to \cite{menghani2023efficient, cheng2017survey, choudhary2020comprehensive, li2023model}, we focus on providing a more comprehensive and holistic view of recent efficient deep learning infrastructures for embedded computing systems, spanning \textbf{from training to inference}, \textbf{from manual to automated}, \textbf{from convolutional neural networks to transformers}, \textbf{from transformers to vision transformers}, \textbf{from vision models to large language models}, \textbf{from software to hardware}, and \textbf{from algorithms to applications}. Specifically, we discuss recent efficient deep learning infrastructures for embedded computing systems from the lens of (1) efficient manual network design for embedded computing systems, (2) efficient automated network design for embedded computing systems, (3) efficient network compression for embedded computing systems, (4) efficient on-device learning for embedded computing systems, (5) efficient large language models for embedded computing systems, (6) efficient deep learning software and hardware for embedded computing systems, and (7) efficient intelligent applications for embedded computing systems. We believe this survey has its merits and can shed light on future research, which can largely benefit researchers to quickly and smoothly get started in this emerging field. Finally, we demonstrate the organization of this survey in Fig.~\ref{fig:survey-structure}, which is also summarized as follows:
\begin{itemize}
    \item {
    Section~\ref{sec:manual-network-design-for-embedded-computing-systems} extensively discusses recent representative efficient manual networks.
    }
    \item {
    Section~\ref{sec:automated-network-design-for-embedded-computing-systems} extensively discusses recent representative efficient automated networks.
    }
    \item {
    Section~\ref{sec:network-compression-for-embedded-computing-systems} extensively discusses recent representative network compression techniques.
    }
    \item {
    Section~\ref{sec:efficient-on-device-learning-for-embedded-computing-systems} extensively discusses recent representative on-device learning techniques.
    }
    \item {
    Section~\ref{sec:large-language-models-for-embedded-computing-systems} extensively discusses recent representative large language models.
    }
    \item {
    Section~\ref{sec:deep-learning-frameworks-for-embedded-computing-systems} extensively discusses recent representative deep learning software and hardware.
    }
    \item {
    Section~\ref{sec:deep-learning-applications-for-embedded-computing-systems} extensively discusses recent representative intelligent embedded applications.
    }
\end{itemize}
Furthermore, at the end of each section, we also envision possible future directions in the respective field, which have the potential to pave the way for future ubiquitous embedded intelligence.

\section{Manual Network Design for Embedded Computing Systems}
\label{sec:manual-network-design-for-embedded-computing-systems}

The tremendous success of DNNs highly relies on the prohibitive network complexity, leading to the computational gap between computation-intensive DNNs and resource-constrained embedded computing systems \cite{liu2022bringing}. To bridge the above computational gap, one of the most representative solutions is to design computation-efficient DNNs to accommodate the limited computational resources on embedded computing systems. To this end, we, in this section, systematically discuss recent state-of-the-art efficient manual networks. For better understanding, we divide these efficient networks into two main categories and sub-sections, including efficient convolutional networks in Section~\ref{sec:manual-convolutional-neural-networks} and efficient transformers in Section~\ref{sec:manual-transformers}, since these efficient networks may feature different network structures and also target different intelligent embedded applications.

\begin{figure}[t]
    \begin{center}
    \includegraphics[width=0.9\columnwidth]{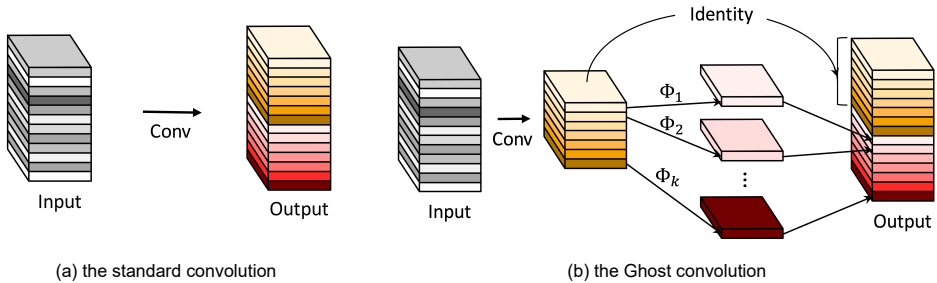}
    \end{center}
    \vspace{-5pt}
    \caption{Comparisons between the standard convolution (\textit{left}) and the Ghost convolution (\textit{right}) of GhostNets \cite{han2020ghostnet, tang2022ghostnetv2, han2022ghostnets}. In particular, compared with the standard convolutional layer, the Ghost convolutional layer can generate rich features using simple and cheaper linear operations. \textbf{(figure from \cite{han2020ghostnet})}}
    \vspace{-5pt}
    \label{fig:ghostnet-overview}
\end{figure}

\subsection{Manual Convolutional Neural Network Design}
\label{sec:manual-convolutional-neural-networks}

As shown in previous state-of-the-art deep convolutional networks, such as AlexNet \cite{krizhevsky2012alexnet}, VGGNet \cite{simonyan2014vggnet}, GoogleNet \cite{szegedy2015googlenet}, ResNet \cite{he2016deep}, DenseNet \cite{huang2017densely}, and EfficientNets \cite{tan2019efficientnet, tan2021efficientnetv2}, despite being able to push forward the attainable accuracy on ImageNet \cite{deng2009imagenet} from 57.2\% \cite{iandola2016squeezenet} to 87.3\% \cite{tan2021efficientnetv2}, the network complexity has increased over time. We note that the convolutional network consists of convolutional layers, pooling layers, and fully-connected layers, where most of the network complexity comes from convolutional layers \cite{cai2022enable}. For example, in ResNet50 \cite{he2016deep}, more than 99\% floating-point operations (FLOPs) are from convolutional layers. In sight of this, designing efficient convolutional layers is critical to innovating computation-efficient convolutional networks. In practice, there are five typical efficient convolutional layers, including pointwise convolution, groupwise convolution, depthwise convolution, dilated convolution, and Ghost convolution:
\begin{itemize}
    \item {
    \textbf{Pointwise Convolution.} 
    Pointwise convolution is a type of convolutional layer with the fixed kernel size of $1 \times 1$, which performs an element-wise multiplication and addition along the depth dimension. On the one hand, compared with the standard $K \times K$ convolutional layer, the pointwise convolutional layer is able to reduce the number of FLOPs and parameters by $K^2$ times, which therefore significantly improves the efficiency. On the other hand, we note that the output from the pointwise convolutional layer typically has the same spatial dimensions as the input but may have a different number of channels. As such, the pointwise convolutional layer can be used to adjust the intermediate feature maps in terms of the number of channels. Specifically, it can reduce or increase the number of channels, making it a practical technique for compressing or expanding convolutional networks. 
    }
    \item {
    \textbf{Groupwise Convolution.}
    Groupwise convolution is a type of convolutional layer that (1) divides the input feature map into $G$ groups along the depth dimension, (2) performs convolution in terms of each group, respectively, and (3) concatenates the outputs along the depth dimension to derive the final output. For example, given an input feature map with the size of $B \times C \times H \times W$, each kernel in the $K \times K$ groupwise convolutional layer is with the size of $(C/G) \times K \times K$, which convolves the above $G$ groups of feature maps, respectively. Therefore, compared with the standard $K \times K$ convolutional layer, the groupwise convolutional layer is able to reduce the number of FLOPs and parameters by $G$ times.
    }
    \item {
    \textbf{Depthwise Convolution.}
    Depthwise convolution is a type of convolutional layer that has gained popularity due to its ability to significantly reduce the number of FLOPs and parameters in convolutional networks. It is a special case of groupwise convolutional layer, in which the number of groups $G$ is equal to the number of input channels. Specifically, each input channel is convolved with a unique kernel of $1 \times K \times K$, after which the outputs from all input channels are concatenated along the depth dimension to derive the final output. In practice, this has the potential to achieve significant reduction in the number of FLOPs and parameters because the intermediate feature maps may consist of thousands of channels as shown in previous state-of-the-art convolutional networks \cite{he2016deep, huang2017densely, tan2019efficientnet, tan2021efficientnetv2}. 
    }
    \item {
    \textbf{Dilated Convolution.}
    Dilated convolution \cite{dilated-conv}, also referred to as atrous convolution, is a type of convolutional layer that is designed to increase the receptive field size. Specifically, in the dilated convolutional layer, there is an adjustable parameter called dilation rate, which determines the spacing between different elements and can be varied to adjust the size of the receptive field. For example, the $3 \times 3$ dilated convolutional layer with the dilation rate of 1 maintains the same receptive field as the standard $5 \times 5$ convolutional layer. This further allows us to increase the receptive field size to unlock better accuracy without introducing additional computational overheads, such as FLOPs and parameters.
    }
    \item {
    \textbf{Ghost Convolution.}
    The Ghost convolution \cite{han2020ghostnet, tang2022ghostnetv2, han2022ghostnets} is a type of convolutional layer that is designed to generate rich feature maps using cheaper computational resources as illustrated in Fig.~\ref{fig:ghostnet-overview}. Specifically, the Ghost convolutional layer consists of two sequential parts. The first part corresponds to the standard convolutional layer, in which the number of output channels is rigorously controlled. Subsequently, in the second part, to generate rich feature maps, a series of simple linear operations are applied to the output feature maps from the first part. As a result, the size of the output feature maps still remains the same as the standard convolutional layer, but the total required computational resources, such as the number of FLOPs and parameters, are significantly reduced as shown in Fig.~\ref{fig:ghostnet-overview}.
    }
    \item {
    \textbf{Partial Convolution.}
    The partial convolution \cite{chen2023run} is to reduce the computational redundancy and memory access simultaneously. Specifically, the partial convolution is built upon the regular convolution, in which only a small number of input channels are convolved with the regular convolution to extract representative spatial features and the remaining input channels are unchanged. Similar to the Ghost convolution, the resulting output channels are further concatenated along the depth dimension to produce the final output channels. In practice, the partial convolution brings significant computational efficiency and memory efficiency since only a small number of input channels are convolved, which also maintains better on-device resource utilization than the Ghost convolution.
    }
\end{itemize}

\begin{figure}[t]
    \begin{center}
    \includegraphics[width=0.9\columnwidth]{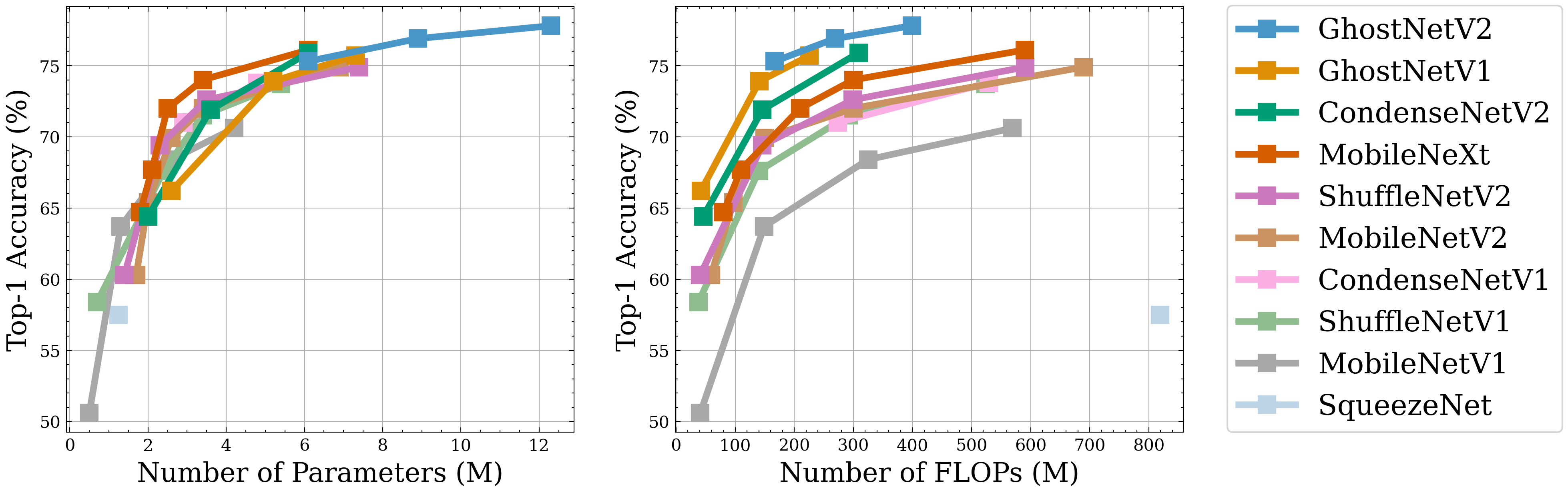}
    \end{center}
    \vspace{-5pt}
    \caption{Comparisons of efficient convolutional networks that have been discussed in Section~\ref{sec:manual-convolutional-neural-networks}, including SqueezeNet \cite{iandola2016squeezenet}, MobileNets \cite{howard2017mobilenets, sandler2018mobilenetv2, zhou2020mobilenext}, ShuffleNets \cite{zhang2018shufflenet, ma2018shufflenet}, CondenseNets \cite{huang2018condensenet, yang2021condensenet}, and GhostNets \cite{han2020ghostnet, han2022ghostnets, tang2022ghostnetv2}, in which the accuracy is evaluated on ImageNet \cite{deng2009imagenet} and is taken from the respective paper. Note that the convolutional networks in this figure may be trained under different training recipes.}
    \vspace{-5pt}
    \label{fig:manual-conv-networks}
\end{figure}

Built on top of the aforementioned efficient convolutional layers and structures, there are several representative families of manually designed efficient convolutional networks, including SqueezeNet \cite{iandola2016squeezenet}, MobileNets \cite{howard2017mobilenets, sandler2018mobilenetv2, zhou2020mobilenext}, ShuffleNets \cite{zhang2018shufflenet, ma2018shufflenet}, CondenseNets \cite{huang2018condensenet, yang2021condensenet}, GhostNets \cite{han2020ghostnet, tang2022ghostnetv2, han2022ghostnets}, and FasterNet \cite{chen2023run}. We compare the above representative efficient convolutional networks in Fig.~\ref{fig:manual-conv-networks}\footnote{We do not include FasterNet \cite{chen2023run} in Fig.~\ref{fig:manual-conv-networks} for comparisons since FasterNet does not optimizes the number of FLOPs.}, which are also discussed in the remainder of this section.

\textbf{SqueezeNet} \cite{iandola2016squeezenet} is stacked using a series of \textit{Fire} modules, which aims to achieve AlexNet-level accuracy with fewer parameters. Specifically, each \textit{Fire} module consists of two convolutional layers, including one \textit{squeeze} layer and one \textit{expand} layer. In the \textit{squeeze} layer, only pointwise convolutional layers are used to reduce the number of input channels for the subsequent \textit{expand} layer. Next, the \textit{expand} layer performs feature expansion using a pair of $1 \times 1$ and $3 \times 3$ convolutional layers. In particular, SqueezeNet is able to achieve slightly better accuracy on ImageNet than AlexNet (i.e., 57.5\% in SqueezeNet vs. 57.2\% in AlexNet) using $\times$50 smaller model size. Meanwhile, SqueezeNet is more compression-friendly than AlexNet. For example, we are allowed to further compress SqueezeNet using \cite{han2015deep}, which delivers more compact network variants with $\times$363$\sim$$\times$510 smaller model size, and more importantly, without degrading the accuracy on ImageNet.

\textbf{MobileNets} \cite{howard2017mobilenets, sandler2018mobilenetv2, zhou2020mobilenext} are a family of lightweight convolutional networks, including MobileNetV1 \cite{howard2017mobilenets}, MobileNetV2 \cite{sandler2018mobilenetv2}, and MobileNeXt \cite{zhou2020mobilenext}, which are tailored for mobile devices with limited computational resources. Specifically, MobileNetV1 is built upon a series of building blocks, where each building block consists of two convolutional layers, including one $3 \times 3$ depthwise convolutional layer and one $1 \times 1$ pointwise convolutional layer. With 569\,M FLOPs and 4.2\,M parameters, MobileNetV1 achieves 70.6\% top-1 accuracy on ImageNet. In addition, MobileNetV2 is an improved version of MobileNetV1, which aims to unlock higher accuracy with fewer FLOPs and parameters. Specifically, MobileNetV2 introduces the inverted residual building block that consists of three convolutional layers, including one $1 \times 1$ pointwise convolutional layer, one $3 \times 3$ depthwise convolutional layer, and one $1 \times 1$ pointwise convolutional layer. Here, the inverted residual building block also borrows the residual connection from ResNet \cite{he2016deep} to stabilize the training process and improve the accuracy. With 300\,M FLOPs and 3.4\,M parameters, MobileNetV2 achieves 72.0\% top-1 accuracy on ImageNet. Furthermore, MobileNeXt investigates the inverted residual building block in MobileNetV2 and introduces the sandglass block to enhance the accuracy without increasing the network complexity. Specifically, the sandglass block consists of four convolutional layers, including one $3 \times 3$ depthwise convolutional layer, one $1 \times 1$ pointwise convolutional layer, one $1 \times 1$ pointwise convolutional layer, and one $3 \times 3$ depthwise convolutional layer. With 300\,M FLOPs and 3.4\,M parameters, MobileNeXt achieves 74.0\% top-1 accuracy on ImageNet.

\textbf{ShuffleNets} \cite{zhang2018shufflenet, ma2018shufflenet} are a family of efficient convolutional networks, including ShuffleNetV1 \cite{zhang2018shufflenet} and ShuffleNetV2 \cite{ma2018shufflenet}, which exploit channel shuffling to reduce the network complexity while maintaining competitive accuracy. Specifically, ShuffleNetV1, for the first time, introduces channel shuffling to enhance the information flow across different channels. In practice, the channel shuffling operation is inserted after the $3 \times 3$ depthwise convolutional layer to shuffle the feature maps from different groups, which is capable of generating richer and more diverse feature maps while not increasing the number of FLOPs and parameters. With 292\,M FLOPs and 3.4\,M parameters, ShuffleNetV1 achieves 71.5\% top-1 accuracy on ImageNet, which is $+$0.9\% higher than MobileNetV1 under comparable settings of FLOPs. Furthermore, ShuffleNetV2 improves the accuracy and efficiency of ShuffleNetV1 with several architectural modifications. Specifically, ShuffleNetV2 first leverages channel splitting to divide the input feature maps into two parallel branches, one of which is fed into three convolutional layers, including one $1 \times 1$ pointwise convolutional layer, one $3 \times 3$ depthwise convolutional layer, and one $1 \times 1$ pointwise convolutional layer. After that, the above two branches of feature maps are concatenated along the depth dimension, which are then shuffled using the channel shuffling operation. In particular, with 299\,M FLOPs and 3.5\,M parameters, ShuffleNetV2 is able to achieve 72.6\% top-1 accuracy on ImageNet, which is $+$1.1\% higher than ShuffleNetV1 under comparable settings of FLOPs.

\textbf{CondenseNets} \cite{huang2018condensenet, yang2021condensenet} are a family of efficient convolutional networks, including CondenseNetV1 \cite{huang2018condensenet} and CondenseNetV2 \cite{yang2021condensenet}, which are built upon another representative convolutional network named DenseNet \cite{huang2017densely}. Specifically, CondenseNetV1 enhances the dense connection with a novel module called learned group convolution. Note that the dense connection re-uses the features from preceding convolutional layers to enhance the information flow as seen in DenseNet. In contrast, the learned group convolution removes the redundant dense connection between different convolutional layers to reduce network redundancy. With 274\,M FLOPs and 2.9\,M parameters, CondenseNetV1 achieves 71.0\% top-1 accuracy on ImageNet. Furthermore, CondenseNetV2 introduces an alternative named sparse feature re-activation (SFR) to increase the feature re-using. In particular, integrated with SFR, each convolutional layer can learn to (1) selectively re-use a set of most important features from preceding convolutional layers and (2) actively update a set of preceding features to increase their re-using in subsequent convolutional layers. With 146\,M FLOPs and 3.6\,M parameters, CondenseNetV2 achieves 71.9\% top-1 accuracy on ImageNet.

\begin{figure}[t]
    \begin{center}
    \includegraphics[width=1.0\columnwidth]{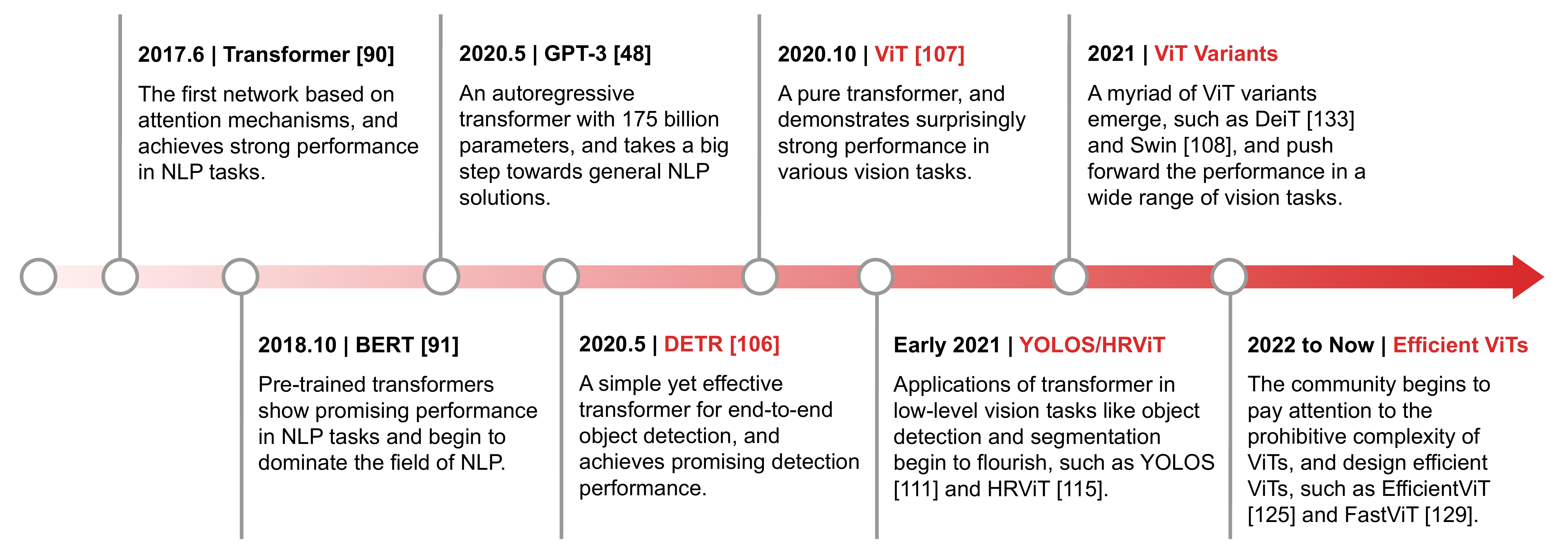}
    \end{center}
    \vspace{-5pt}
    \caption{Illustration of the key milestones of transformer, which is originally applied to NLP tasks and has recently gained increasing popularity in the vision community. Here, we mark the vision transformers in red.}
    \vspace{-5pt}
    \label{fig:transformer-milestones}
\end{figure}

\textbf{GhostNets} \cite{han2020ghostnet, han2022ghostnets, tang2022ghostnetv2} are a family of efficient deep convolutional networks, including GhostNetV1 \cite{han2020ghostnet, han2022ghostnets} and GhostNetV2 \cite{tang2022ghostnetv2}, which focus on generating rich feature maps using computationally cheap and simple yet powerful operations. To this end, GhostNetV1 introduces a powerful yet computation-efficient convolution dubbed Ghost convolution as shown in Fig.~\ref{fig:ghostnet-overview}, which consists of two sequential parts. The first part corresponds to the standard convolutional layer, where the number of output channels is rigorously controlled. And next, in the second part, a series of computationally cheap and simple linear operations are applied to the output feature maps from the first part to generate rich feature maps. In particular, with only 141\,M FLOPs and 5.2\,M parameters, GhostNetV1 achieves 73.9\% top-1 accuracy on ImageNet. Furthermore, GhostNetV2 introduces a novel hardware-friendly attention mechanism, namely DFC attention, to enhance the learned feature maps to boost the expressiveness ability, which is seamlessly integrated into GhostNetV1 to push forward the accuracy and efficiency. For example, with 167\,M FLOPs and 6.1\,M parameters, GhostNetV2 is able to achieve 75.3\% top-1 accuracy on ImageNet.

\textbf{FasterNets} \cite{chen2023run} are built upon the partial convolution. In contrast to the above efficient networks that typically optimize the number of FLOPs, FasterNet pioneers to design efficient networks with optimized FLOPS (i.e., FLOPs per second). The motivation behind FasterNet is that the on-device latency is determined by both FLOPs and FLOPS (i.e., Latency=FLOPs/FLOPS). To this end, FasterNet focuses on increasing the number of FLOPs to maintain competitive accuracy on target task, while at the same time optimizing FLOPS to maintain competitive efficiency on target hardware. For example, compared with GhostNetV1x1.3 that involves 0.24\,G FLOPs and exhibits 75.7\% top-1 accuracy on ImageNet, FasterNet-T1 achieves $+$0.5\% higher top-1 accuracy with much more FLOPs (i.e., 0.85\,G), and more importantly, achieves $\times$1.7 speedup on ARM processors.

\subsection{Manual Transformer Design}
\label{sec:manual-transformers}

\subsubsection{Transformer for NLP}
In parallel to convolutional networks, transformer \cite{vaswani2017attention} is another well-established branch of DNNs, which exploits multi-head self-attention mechanisms. In practice, transformer is first designed and applied to natural language processing (NLP) tasks, where it has achieved tremendous success. For example, BERT \cite{devlin2018bert}, as one of the most representative transformers in the field of NLP, is able to achieve state-of-the-art performance across 11 downstream NLP tasks, such as language translation, question answering, language generation, etc., at the moment of BERT being proposed. Furthermore, GPT-3 \cite{brown2020language}, also known as Generative Pre-trained Transformer 3, pioneers to scale up and pre-train a massive transformer that consists of 175 billion parameters on 45\,TB compressed plaintext data, which unlocks even stronger performance across almost all downstream NLP tasks, and more importantly, without requiring fine-tuning on specific NLP tasks. More recently, GPT-4 \cite{openai2023gpt4} has been proposed by OpenAI, which can significantly outperform GPT-3 across a wide range of language processing tasks and has also been widely integrated into various real-world language processing tasks, such as ChatGPT \cite{openai2020chatgpt}, to provide intelligent language processing services. These early transformer-based deep networks, thanks to their prohibitive computational complexity, have been pushing forward the boundaries of various language processing tasks and dominating recent advances in the field of NLP (see Fig.~\ref{fig:transformer-milestones}).

Nonetheless, it is quite challenging to deploy powerful transformers on embedded computing systems due to the computational gap between computation-intensive transformers and computation-limited embedded computing systems. For example, as pointed out in \cite{wang2020hat}, to translate a short sentence with only 30 words, a typical transformer model needs to execute 13\,G FLOPs, which takes 20 seconds on a Raspberry Pi device. This significantly hinders the user experience in real-world embedded scenarios. To tackle this issue, a series of computation-efficient transformers have emerged, among which TinyBERT \cite{jiao2020tinybert}, MobileBERT \cite{sun2020mobilebert}, DistilBERT \cite{sanh2019distilbert}, Linformer \cite{wang2020linformer}, and Reformer \cite{kitaev2020reformer} are some of the most representative ones. The main intuition behind these efficient transformers is to resolve the memory bottleneck and increase the parallelism, making it possible to deploy NLP workloads on resource-constrained embedded computing systems. Note that, compared with computer vision tasks like image classification and object detection, running NLP workloads on embedded computing systems is less common due to the high inference latency. For example, as demonstrated in \cite{wang2020hat}, running language translation workloads with hardware-tailored transformers on a Raspberry Pi device takes even seconds, whereas running image classification workloads typically takes milliseconds per image. More recently, inspired by the remarkable success of GPTs \cite{brown2020language, openai2023gpt4}, transformer-based large language models have become increasingly popular in the NLP community. To optimize the efficiency of transformer-based large language models, a plethora of efficient transformer-based large language models have been proposed, which typically focus on improving the training efficiency \cite{kaddour2024no, wang2023learning, ostendorff2023efficient}, the inference efficiency \cite{pope2023efficiently, zhou2023brainformers}, and the fine-tuning efficiency \cite{zhang2023towards, zhang2023llama} of transformers in the context of large language models. For example, to optimize the inference efficiency of transformer-based large language models, \cite{pope2023efficiently} partitions large language models over different hardware chips in order to fit weights and activation tensors into memory and run computation and memory workloads within the given latency constraint, which also features a simple yet effective strategy to alleviate the communication overheads among different hardware chips for cost-effective and latency-efficient inference.

\subsubsection{Transformer for Vision}
\begin{figure}[t]
    \begin{center}
    \includegraphics[width=0.85\columnwidth]{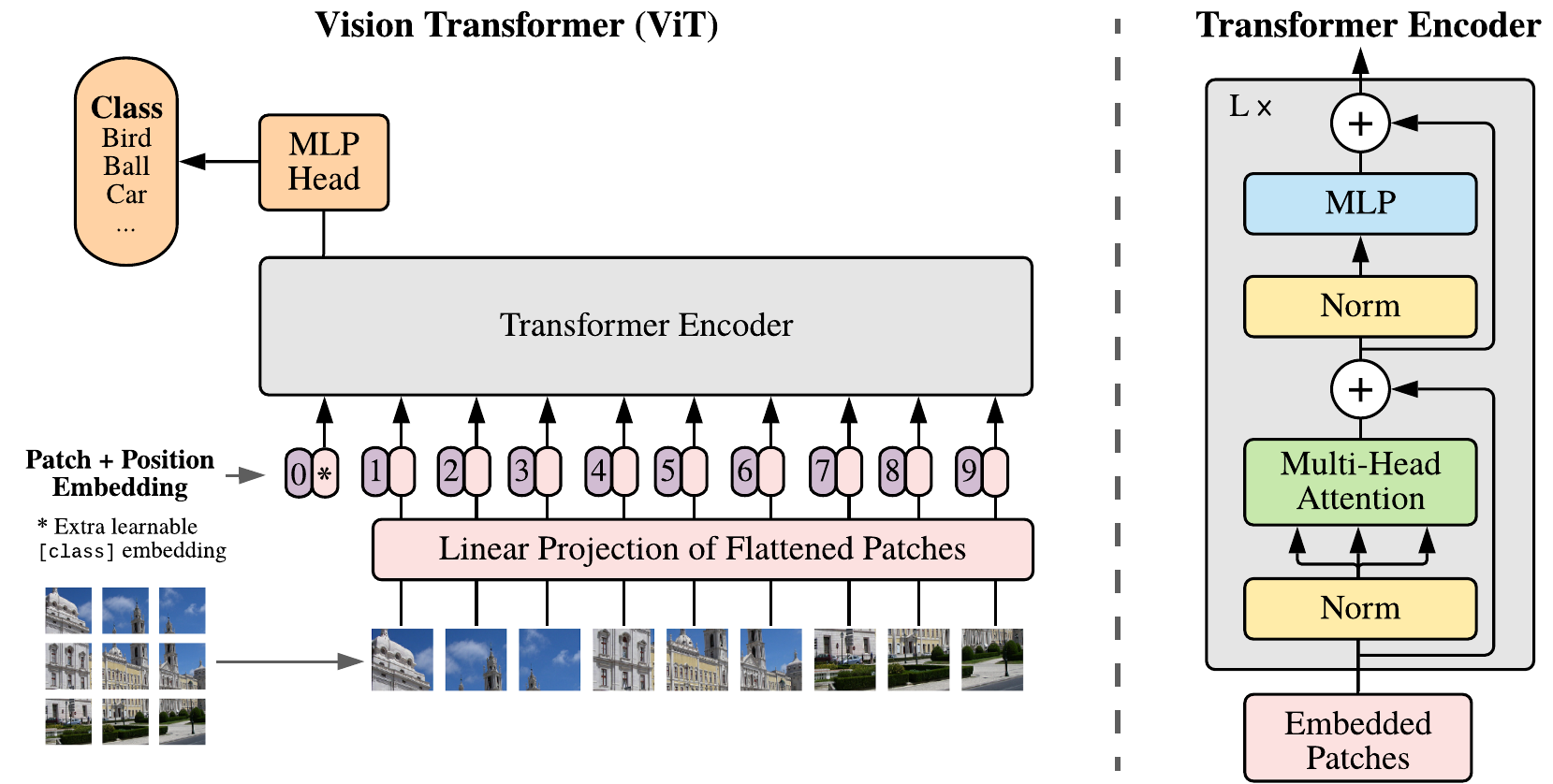}
    \end{center}
    \vspace{-5pt}
    \caption{Overview of Vision Transformer (ViT) \cite{dosovitskiy2020image}, which (1) splits the image into fixed-size patches, (2) linearly embeds each of them, and (3) feeds the sequence of vectors into the encoder. \textbf{(figure from \cite{dosovitskiy2020image})}}
    \vspace{-5pt}
    \label{fig:vit-overview}
\end{figure}

Inspired by the tremendous success of transformer in the field of NLP, researchers have recently applied transformer to vision tasks, which achieves surprisingly strong performance (see Fig.~\ref{fig:transformer-milestones}). This opens up a new direction and further challenges the dominant role of convolutional networks in vision tasks. Specifically, DETR \cite{carion2020end} and Vision Transformer (ViT) \cite{dosovitskiy2020image} are the very early transformers in vision tasks, among which ViT is the most representative one. These early pioneers have motivated a myriad of subsequent transformers in various vision tasks, such as image classification \cite{dosovitskiy2020image, liu2021swin, liu2022swin}, object detection \cite{li2022exploring, fang2021yolos, minderer2022simple}, semantic segmentation \cite{strudel2021segmenter, chen2021transunet, gu2022multi, kirillov2023segmentanything}, and video analysis \cite{liu2022video, arnab2021vivit, neimark2021video}. For example, ViT is first proposed in June 2020, which has since gained over 20,000 citations as shown in Google Scholar. In particular, the main intuition behind ViT is surprisingly simple and straightforward, which (1) splits the input image into a series of fixed-size patches, (2) linearly embeds each of them, and (3) feeds the resulting sequence of vectors into the standard transformer encoder as illustrated in Fig.~\ref{fig:vit-overview}. However, there is no free lunch. The surprisingly strong performance of ViT and its variants comes at the cost of prohibitive computational complexity, which significantly hinders the practical deployments of ViT and its variants on embedded computing systems with limited computational resources.

\begin{figure}[t]
    \begin{center}
    \includegraphics[width=0.9\columnwidth]{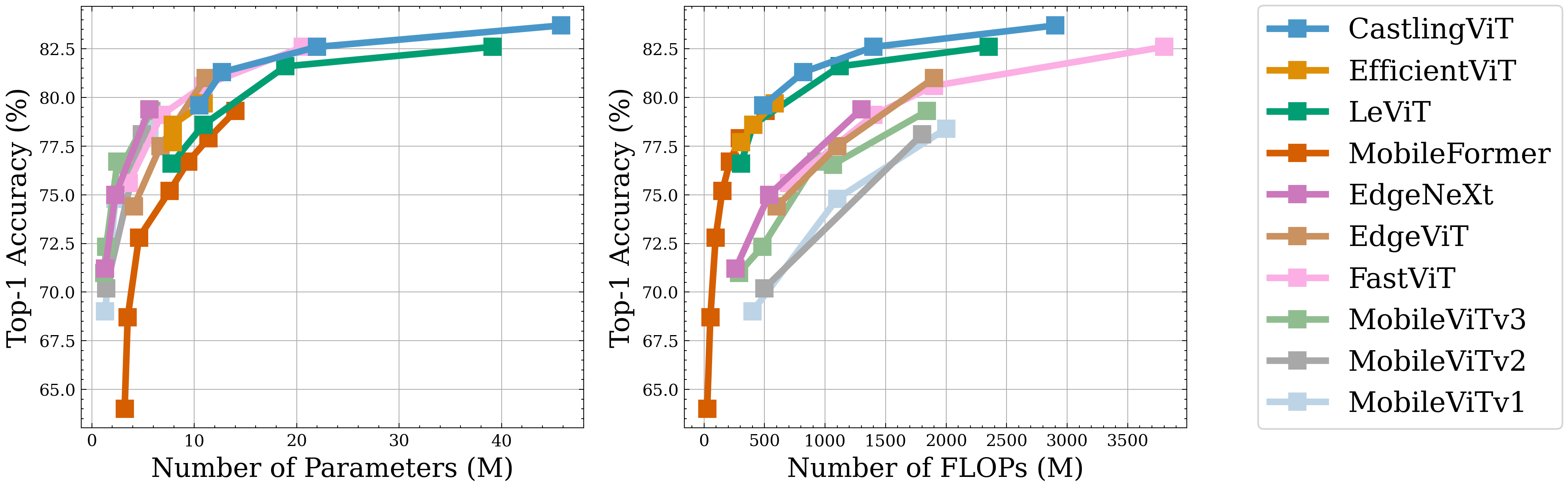}
    \end{center}
    \vspace{-5pt}
    \caption{Comparisons of efficient vision transformers that have been discussed in Section~\ref{sec:manual-transformers}, including LeViT \cite{graham2021levit}, MobileFormer \cite{chen2022mobile}, MobileViTs \cite{mehta2021mobilevit, mehta2022mobilevitv2, wadekar2022mobilevitv3}, EfficientViT \cite{cai2022efficientvit}, EdgeViT \cite{pan2022edgevits}, EdgeNeXt \cite{maaz2023edgenext}, CastlingViT \cite{you2022castling}, and FastViT \cite{vasu2023fastvit}, in which the accuracy is evaluated on ImageNet \cite{deng2009imagenet} and is taken from the respective paper. Note that the vision transformers here may be trained under different training recipes.}
    \vspace{-5pt}
    \label{fig:manual-vision-transformers}
\end{figure}

To resolve the complexity bottleneck, some recent works have pioneered to design computation-efficient transformers for vision tasks, with the aim of reducing the computational complexity while maintaining competitive accuracy. The representative computation-efficient transformers in vision tasks include LeViT \cite{graham2021levit}, MobileFormer \cite{chen2022mobile}, MobileViTs \cite{mehta2021mobilevit, mehta2022mobilevitv2, wadekar2022mobilevitv3}, EfficientViT \cite{cai2022efficientvit}, EdgeViT \cite{pan2022edgevits}, EdgeNeXt \cite{maaz2023edgenext}, CastlingViT \cite{you2022castling}, and FastViT \cite{vasu2023fastvit}. The above computation-efficient vision transformers are summarized and compared in Fig.~\ref{fig:manual-vision-transformers}.

\textbf{LeViT} \cite{graham2021levit} is a hybrid vision transformer built on top of convolutional networks, which aims to improve the trade-off between accuracy and efficiency. To this end, LeViT introduces several enhancements to shrink down the network size, including (1) a multi-stage transformer architecture that uses attention mechanisms as down-sampling, (2) a computation-efficient patch descriptor that shrinks down the number of features in the early layers, (3) a per-head translation-invariant attention bias that replaces ViT's positional embeddings, and (4) an efficient MLP-based attention block that improves the network capacity under given computational budgets. With 406\,M FLOPs and 9.2\,M parameters, LeViT achieves 78.6\% top-1 accuracy on ImageNet.

\textbf{MobileFormer} \cite{chen2022mobile} parallelizes MobileNetV2 \cite{sandler2018mobilenetv2} and transformer \cite{dosovitskiy2020image} with a two-way bridge, which shifts the network design paradigm from series to parallel. The network here is named MobileFormer, where \textit{Mobile} refers to MobileNetV2 and \textit{Former} stands for transformer. Specifically, \textit{Mobile} takes the image as input and stacks inverted residual blocks that consist of efficient pointwise and depthwise convolutional layers to extract local features. \textit{Former} takes learnable tokens as input and stacks multi-head attention and feed-forward networks, in which the learnable tokens encode global features of the image. As such, \textit{Mobile} and \textit{Former} can communicate through a two-way bridge to fuse local and global features for better expressiveness ability. With 294\,M FLOPs and 11.4\,M parameters, MobileFormer achieves 77.9\% top-1 accuracy on ImageNet.

\textbf{MobileViTs}, including MobileViTv1 \cite{mehta2021mobilevit}, MobileViTv2 \cite{mehta2022mobilevitv2}, and MobileViTv3 \cite{wadekar2022mobilevitv3}, are a family of efficient hybrid networks that combine the benefits of CNNs (e.g., spatial inductive bias and less sensitivity to data augmentations) and vision transformers (e.g., input-adaptive weighting and global processing). Different from mainstream vision transformers, both MobileViTv1 and MobileViTv2 are designed with the aim of low inference latency rather than low FLOPs since the number of FLOPs cannot accurately reflect the inference efficiency on target hardware. To this end, MobileViTv1 introduces a novel block that is able to efficiently and effectively encode both local and global features. In addition, MobileViTv1 also replaces local processing in convolutional layers with global processing using transformers, which can lead to better representation capability with fewer parameters and simpler training recipes. Finally, with 5.6\,M parameters, MobileViTv1 achieves 78.4\% top-1 accuracy on ImageNet. Furthermore, MobileViTv2 introduces a separable self-attention mechanism with linear complexity, which is integrated into MobileViTv1 to boost the accuracy and hardware efficiency. For example, MobileViTv2 achieves 75.6\% top-1 accuracy on ImageNet, which is $+$0.8\% higher than MobileViTv1 while maintaining $\times$3.2 speedup on iPhone 12. In addition, MobileViTv3 introduces two simple yet effective enhancements, including (1) replacing $3\times3$ convolutional layers with $1t\times1$ convolutional layers and (2) scaling up building blocks in terms of the network width. With 927\,M FLOPs, MobileViTv3 achieves 76.7\% top-1 accuracy on ImageNet, which is $+$1.9\% higher than MobileViTv1 under similar FLOPs.

\textbf{EfficientViT} \cite{cai2022efficientvit} investigates high-resolution low-computation visual recognition tasks using ViT and its variants, and identifies that the complexity bottleneck of ViT and its variants comes from the excessively used softmax attention mechanism. To resolve the complexity bottleneck, EfficientViT challenges the dominant role of softmax attention in vision transformers and further introduces a strong alternative, namely enhanced linear attention, to replace softmax attention, which demonstrates strong representation capability in local feature extraction while being able to maintain low computational complexity and high hardware efficiency. With 406\,M FLOPs and 7.9\,M parameters, EfficientViT achieves 78.6\% top-1 accuracy on ImageNet.

\textbf{EdgeViT} \cite{pan2022edgevits} investigates the design of efficient vision transformers from the perspective of on-device deployment, enabling vision transformers to compete with state-of-the-art CNNs in terms of the accuracy-efficiency trade-off. Specifically, EdgeViT is designed based on an optimal decomposition of self-attention using standard primitive operations, optimizing EdgeViT towards target hardware to achieve superior accuracy-efficiency trade-offs. With 600\,M FLOPs and 4.1\,M parameters, EdgeViT achieves 74.4\% top-1 accuracy on ImageNet, which is $+$2.4\% higher than MobileNetV2 under comparable latency constraints on Samsung Galaxy S21.

\textbf{EdgeNeXt} \cite{maaz2023edgenext} is an efficient hybrid network that marries both worlds of convolutional networks and vision transformers. To better encode the global information, EdgeNeXt introduces an efficient \textit{split depthwise transpose attention} (SDTA) encoder to address the issue of limited receptive fields in CNNs without increasing the number of FLOPs and parameters. In addition, EdgeNeXt also leverages adaptive kernel sizes to shrink down the network complexity. With 538\,M FLOPs and 2.3\,M parameters, EdgeNeXt achieves 75.0\% top-1 accuracy on ImageNet, which is comparable to MobileViTv1 \cite{mehta2021mobilevit} in terms of both accuracy and on-device latency.

\textbf{CastlingViT} \cite{you2022castling} proposes to (1) train ViT and its variants using both linear-angular attention and masked softmax-based quadratic attention and (2) switch to having only linear-angular attention during the inference in order to save computational resources. Specifically, the linear-angular attention leverages angular kernels to bridge the accuracy gap between linear attention and softmax-based attention. It expands angular kernels where linear terms are kept while complex high-order residuals are approximated. This aligns with the observation in EfficientViT \cite{cai2022efficientvit} that the complexity bottleneck of ViT and its variants comes from the excessively involved softmax attention mechanism. To address the complexity bottleneck, CastlingViT replaces softmax attention with linear-angular attention to further improve the efficiency of ViT and its variants. With 490\,M FLOPs and 10.5\,M parameters, CastlingViT achieves 79.6\% top-1 accuracy on ImageNet.

\textbf{FastViT} \cite{vasu2023fastvit} is an efficient hybrid network that combines both CNNs and vision transformer, which aims to marry the best of both and enable state-of-the-art accuracy-efficiency trade-offs. To this end, FastViT introduces a novel token mixing operator named RepMixer, which is the basic building block of FastViT that leverages structural reparameterization to reduce the memory access cost by removing the less important skip connections. In addition, FastViT also applies training-time over-parameterization and large kernel convolutions to further boost the accuracy with minimal effect on the inference latency. In practice, structural reparameterization enables FastViT to achieve strong accuracy on target task during the training process and maintain superior efficiency on target hardware during the on-device inference process. With 700\,M FLOPs and 3.6\,M parameters, FastViT achieves 75.6\% top-1 accuracy on ImageNet.

\subsection{Future Envision}
In this section, we envision the future trends and possible directions of manual network design, including convolutional networks and transformers, which are summarized as follows:
\begin{itemize}
    \item[(1)] {
    \textbf{Hardware-Aware Optimization.}
    The trend in the field of network design is to reduce the number of FLOPs. However, the number of FLOPs only represents the theoretical complexity and the reduction in the number of FLOPs does not necessarily lead to the inference speedup on target hardware \cite{mehta2021mobilevit, mehta2022mobilevitv2, pan2022edgevits, luo2020edgenas, luo2022you}. For example, PiT \cite{heo2021rethinking} has $\times$3 fewer FLOPs than DeiT \cite{touvron2021deit}, but both have similar inference latency on iPhone 12 (i.e., DeiT vs. PiT on iPhone 12: 10.99\,ms vs. 10.56\,ms) \cite{mehta2021mobilevit}. In parallel, the attention mechanisms are powerful plug-in enhancements in various real-world scenarios \cite{hu2020introductory, guo2022attention}, such as Squeeze-and-Excitation (SE) \cite{hu2018squeeze} in vision tasks and self-attention \cite{vaswani2017attention} in NLP tasks, which can further boost the attainable accuracy on target task while slightly increasing the number of FLOPs. However, DNNs with attention mechanisms, despite being able to push forward the accuracy on target task, introduce considerable extra parameters and are difficult to parallelize on target hardware, especially for transformers that are full of self-attention mechanisms. For example, EfficientViT \cite{cai2022efficientvit} demonstrates that the prohibitive computational complexity of ViT and its variants comes from the excessively used softmax attention. In light of the above, we should focus on optimizing more direct efficiency metrics, such as latency and energy, which may directly benefit real-world embedded computing systems.
    }
    \item[(2)] {
    \textbf{Interpretability and Explainability.}
    Recent manually designed DNNs, including efficient convolutional networks and transformers, have been empirically developed through trial and error. The intuition behind this is that DNNs suffer from limited interpretability and explainability \cite{angelov2020towards}. Therefore, to find one decent network solution with competitive accuracy, we have to repeat a plethora of training experiments to evaluate the accuracy of possible network configurations \cite{zoph2016neural, liu2019darts}, thereby necessitating non-trivial computational resources for repeated training workloads \cite{luo2020edgenas, luo2022you}. To avoid this, we, in the future, should focus on addressing the interpretability and explainability of DNNs so as to facilitate the network design process and minimize the required engineering efforts.
    }
    \item[(3)] {
    \textbf{Hybrid Multi-Modal Networks.}
    Compared with vision transformers, convolutional networks are able to maintain superior efficiency on target hardware, but may suffer from inferior accuracy on target task. However, self-attention mechanisms are excessively involved in vision transformers, which are difficult to parallelize on mainstream embedded computing systems \cite{cai2022efficientvit, you2022castling}. For example, as demonstrated in EdgeViT \cite{pan2022edgevits}, under similar FLOPs settings, MobileNetV2 \cite{sandler2018mobilenetv2} is about $\times$2 faster on Samsung Galaxy S21 than MobileViTv1 \cite{mehta2021mobilevit}. This further hinders the practical deployment of vision transformers in real-world embedded scenarios. In parallel, \cite{han2022vision} demonstrates that, similar to transformers, graph neural networks, when properly engineered, can also achieve competitive performance in vision tasks. More importantly, these hybrid networks have the potential to handle various modalities (i.e., different types of input), such as text, image, and audio \cite{jangra2021survey}. For example, convolutional networks are particularly effective at handling spatial data, such as image. In contrast, transformers are better suited for sequential data, such as text. Therefore, in order to achieve better accuracy-efficiency trade-offs and allow diverse input modalities, one natural and promising future direction is to continue exploring hybrid multi-modal networks that combine the strengths of existing representative networks, such as convolutional networks, vision transformers, and graph networks.
    }
    \item[(4)] {
    \textbf{Simpler Training Recipes.}
    As demonstrated in \cite{steiner2021train}, the competitive performance of ViT and its variants highly relies on more advanced training recipes, such as pre-trained on larger datasets, more training epochs, stronger data augmentations, and stronger regularization strategies. For example, ViT \cite{dosovitskiy2020image} is first pre-trained on ImageNet-21k and JFT and then fine-tuned on ImageNet. Note that ImageNet consists of 1,000 categories, whereas ImageNet-21k has 21,000 categories. This further makes it more difficult and challenging to train vision transformers under regular training settings and significantly increases the total training cost. Therefore, training vision transformers in a more computation-efficient manner and under simpler training recipes is a promising future direction.
    }
    \item[(5)] {
    \textbf{Adversarial Robustness.}
    In addition to the efficiency, the adversarial robustness is another desirable network property since efficient networks, especially vision transformers \cite{fu2022patch}, are more sensitive to input perturbations, and as a result, are more vulnerable to adversarial attacks than non-efficient ones \cite{ye2019adversarial}. Specifically, the adversarial robustness refers to the ability of the network to maintain its accuracy, even when encountering adversarial attacks that are intentionally designed to mislead the network. The adversarial robustness is critical in real-world scenarios, especially in those where the environments are complex and unpredictable, such as autonomous vehicles. Therefore, innovating efficient yet robust DNNs is a promising future direction in the field of network design.
    }
\end{itemize}

\section{Automated Network Design for Embedded Computing Systems}
\label{sec:automated-network-design-for-embedded-computing-systems}

In contrast to manual network design, automated network design, also known as neural architecture search (NAS) \cite{zoph2016neural}, has recently flourished, which strives to automate the design of efficient neural networks. In the past decade, NAS has achieved impressive performance in the field of network design, which delivers more advanced networks with both higher accuracy and efficiency than the conventional manual network design (see Section~\ref{sec:manual-network-design-for-embedded-computing-systems}). To this end, we, in this section, further discuss recent advances in the field of NAS, especially from the perspective of hardware-aware NAS that searches for hardware-efficient network solutions, including modular search space in Section~\ref{sec:modular-search-space}, search strategy in Section~\ref{sec:search-strategy}, and speedup techniques and extensions in Section~\ref{sec:speedup-techniques-and-extensions}.

\begin{figure}[t]
    \begin{center}
    \includegraphics[width=0.85\columnwidth]{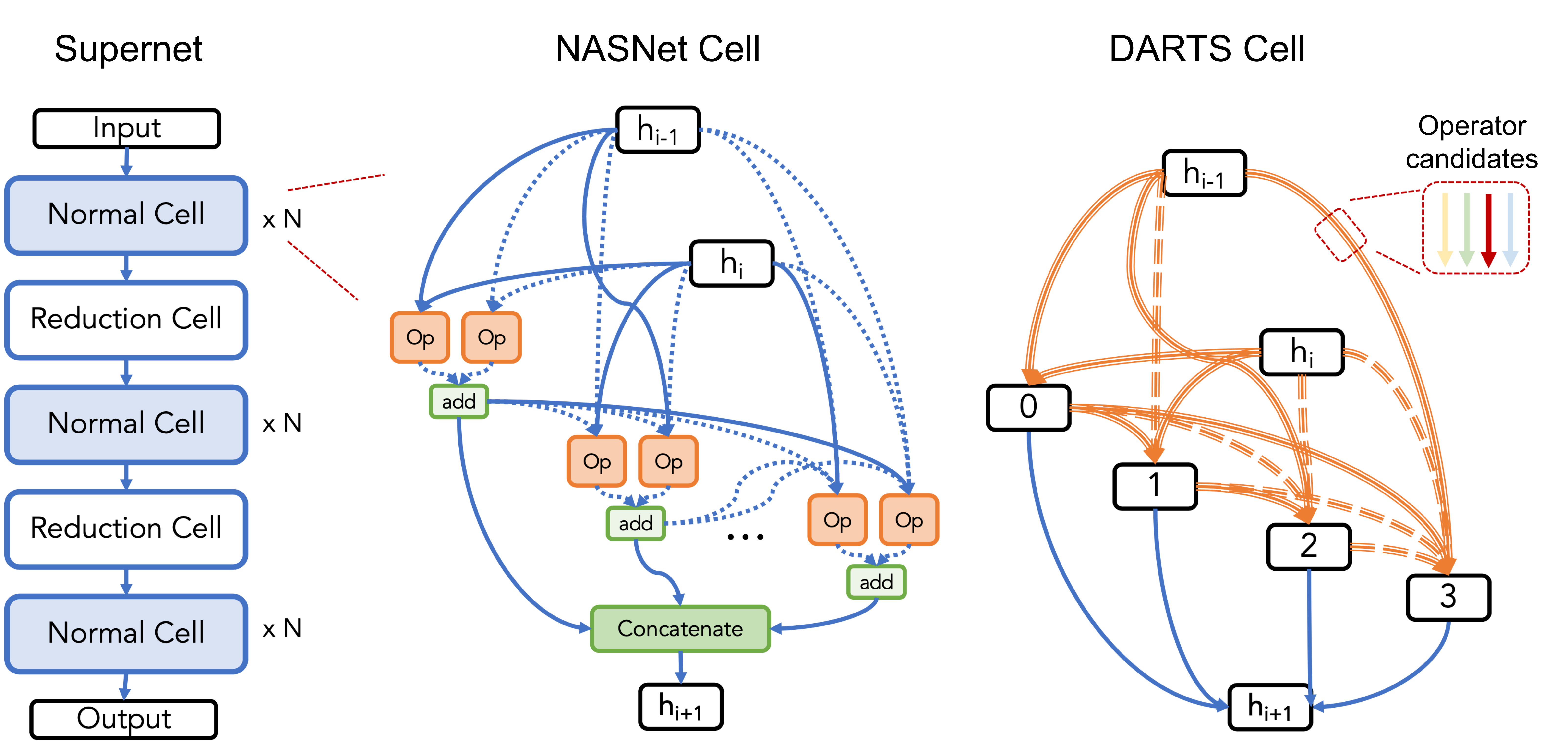}
    \end{center}
    \vspace{-5pt}
    \caption{Illustration of the cell-based search space $\mathcal{A}$ in NASNet \cite{zoph2018nasnet} and DARTS \cite{liu2019darts}, in which NASNet assigns operator candidates to nodes and DARTS assigns operator candidates to edges. \textbf{(figure from \cite{white2023neural})}}
    \vspace{-5pt}
    \label{fig:cell-search-space}
\end{figure}

\subsection{Modular Search Space}
\label{sec:modular-search-space}

The search space $\mathcal{A}$ plays a prominent role in the success of NAS since the search engine of NAS strives to search for top-performing architecture candidates within the pre-defined search space. This also indicates that the search space determines the upper-performance limit of modern NAS algorithms. However, designing efficient and effective search spaces is quite difficult and challenging since there are a myriad of possible operator candidates (e.g., $1\times1$, $3\times3$, $5\times5$, and $7\times7$ convolutional layers) and different network configurations (e.g., the combination strategies of different operator candidates and the network channel layouts) \cite{zoph2016neural, zoph2018nasnet, liu2019darts}. Therefore, to reduce the search space size and trim down the search complexity, previous state-of-the-art NAS methods \cite{zoph2016neural, zoph2018nasnet, liu2019darts} often restrict the search space to allow efficient search and leverage modular search spaces, which are coarse-grained in contrast to layer-wise fine-grained search spaces. In practice, previous state-of-the-art NAS methods are based on the following two representative types of modular search spaces, including cell-based search space and block-based search space.

\textbf{Cell-Based Search Space.}
The cell-based search space $\mathcal{A}$ has been dominating the early success in the field of NAS. Specifically, the cell-based search space is first introduced by NASNet \cite{zoph2018nasnet} and DARTS \cite{liu2019darts}. As defined in NASNet, the cell-based search space consists of two types of cell structures, which are denoted as the normal cell and the reduction cell. In practice, both types of cells are encoded into directed acyclic graphs (DAGs) and maintain the same cell structure as illustrated in Fig.~\ref{fig:cell-search-space}, except that the reduction cell starts with one convolutional layer with the stride of 2 to reduce the input spatial dimension. Once the cell structure is determined at the end of search, it is then repeatedly stacked to derive the final architecture candidate. In addition, DARTS introduces another type of cell-based search space, which has motivated a plethora of subsequent NAS methods that are also built on top of the same cell-based search space, such as RobustDARTS \cite{zela2020robustdarts}, EdgeNAS \cite{luo2020edgenas}, PC-DARTS \cite{xu2020pcdarts}, P-DARTS \cite{chen2019pdarts}, DARTS+ \cite{liang2019darts+}, DARTS- \cite{chu2021darts-}, FairDARTS \cite{chu2020fairdarts}, and $\beta$-DARTS \cite{ye2022betadarts}. Similar to NASNet, the cell-based search space in DARTS consists of two types of cells, including the normal cell and the reduction cell. As shown in Fig.~\ref{fig:cell-search-space}, each cell has an ordered sequence of nodes, where each node is a latent representation (e.g., a feature map in convolutional networks) and each directed edge has a set of possible operators $\{o^{(i, j)}\}$ that transform the input $x^{(i)}$. Different from NASNet, the cell in DARTS is assumed to have two different input nodes and one single output node. With the above in mind, we are able to mathematically calculate the intermediate node as follows, which is based on all of its predecessors:
\begin{equation}
    x^{(j)} = \sum_{i < j} o^{(i, j)}(x^{(i)})
\end{equation}
Finally, the search space of DARTS contains $6.3 \times 10^{29}$ possible architecture candidates \cite{luo2020edgenas}.

\begin{figure}[t]
    \begin{center}
    \includegraphics[width=1.0\columnwidth]{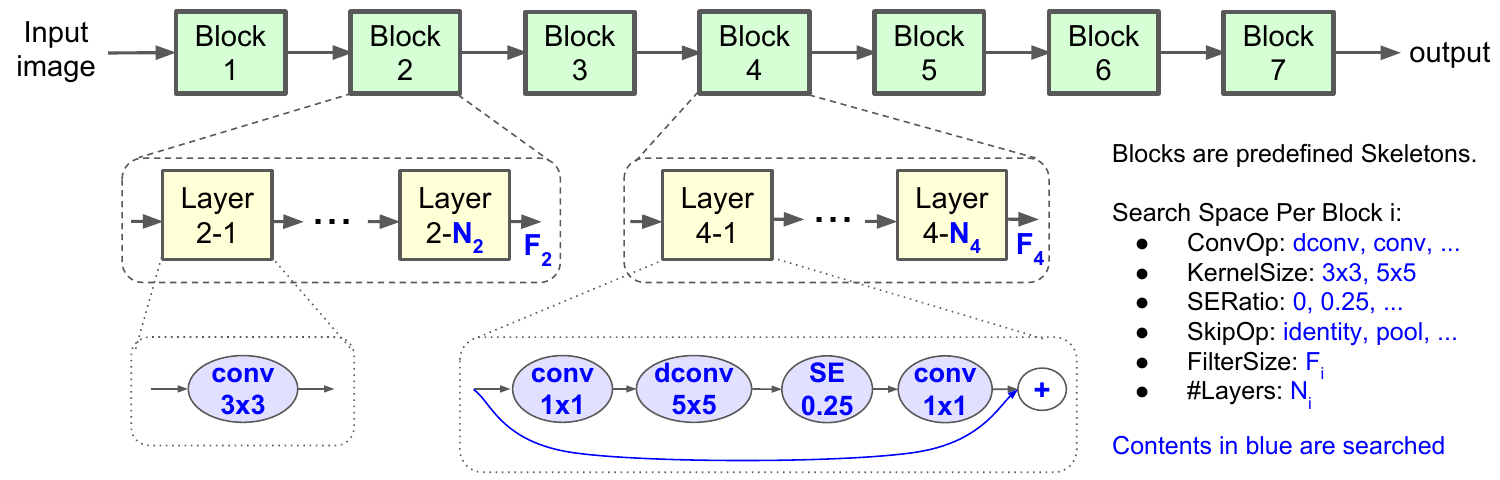}
    \end{center}
    \vspace{-5pt}
    \caption{Illustration of the block-based search space $\mathcal{A}$, which is based on MobileNetV2. \textbf{(figure from \cite{tan2019mnasnet})}}
    \vspace{-5pt}
    \label{fig:block-search-space}
\end{figure}

\textbf{Block-Based Search Space.}
The block-based search space $\mathcal{A}$ advocates for simple and diverse network topologies as illustrated in Fig.~\ref{fig:block-search-space}, in which each architecture candidate consists of multiple sequential operator candidates. As shown in previous NAS practices, the operator candidates in the block-based search space are usually taken from state-of-the-art manual DNNs, such as MobileNets \cite{howard2017mobilenets, sandler2018mobilenetv2} and ShuffleNets \cite{zhang2018shufflenet, ma2018shufflenet}. For example, the block-based search space in ProxylessNAS \cite{cai2019proxylessnas} is built on top of MobileNetV2, whereas the block-based search space in HSCoNAS \cite{luo2021hsconas} is built on top of ShuffleNetV2. In parallel, HURRICANE \cite{zhang2020fast} demonstrates that different hardware platforms favor different search spaces, based on which HURRICANE introduces a hybrid block-based search space that combines both MobileNetV2 and ShuffleNetV2 to deliver superior architecture solutions. Different from the cell-based search space, the block-based search space is hardware-friendly, due to which the block-based search space has been widely adopted in previous hardware-aware NAS methods, such as MnasNet \cite{tan2019mnasnet}, ProxylessNAS \cite{cai2019proxylessnas}, OFA \cite{cai2020once}, HSCoNAS \cite{luo2021hsconas}, SurgeNAS \cite{luo2022surgenas}, and LightNAS \cite{luo2022you}. The intuition behind this is that the architecture candidate in the cell-based search space consists of multiple parallel branches as shown in Fig.~\ref{fig:cell-search-space}, which introduce additional overheads in terms of the memory access, and as a result, deteriorate the inference efficiency on target hardware according to the roofline analysis \cite{williams2009roofline}. In addition, different from the cell-based search space that repeatedly stacks the same cell structure across the entire network, the block-based search space allows operator diversity within different blocks, encouraging to find architecture candidates with better accuracy-efficiency trade-offs \cite{luo2022lightnas}.

\subsection{Search Strategy} 
\label{sec:search-strategy}

In this section, we discuss recent state-of-the-art NAS algorithms and divide them into three main categories, including reinforcement learning-based search \cite{zoph2016neural}, evolutionary algorithm-based search \cite{real2019regularized}, and gradient-based search (also known as differentiable search) \cite{liu2019darts}.

\begin{figure}[t]
    \begin{center}
    \includegraphics[width=0.985\columnwidth]{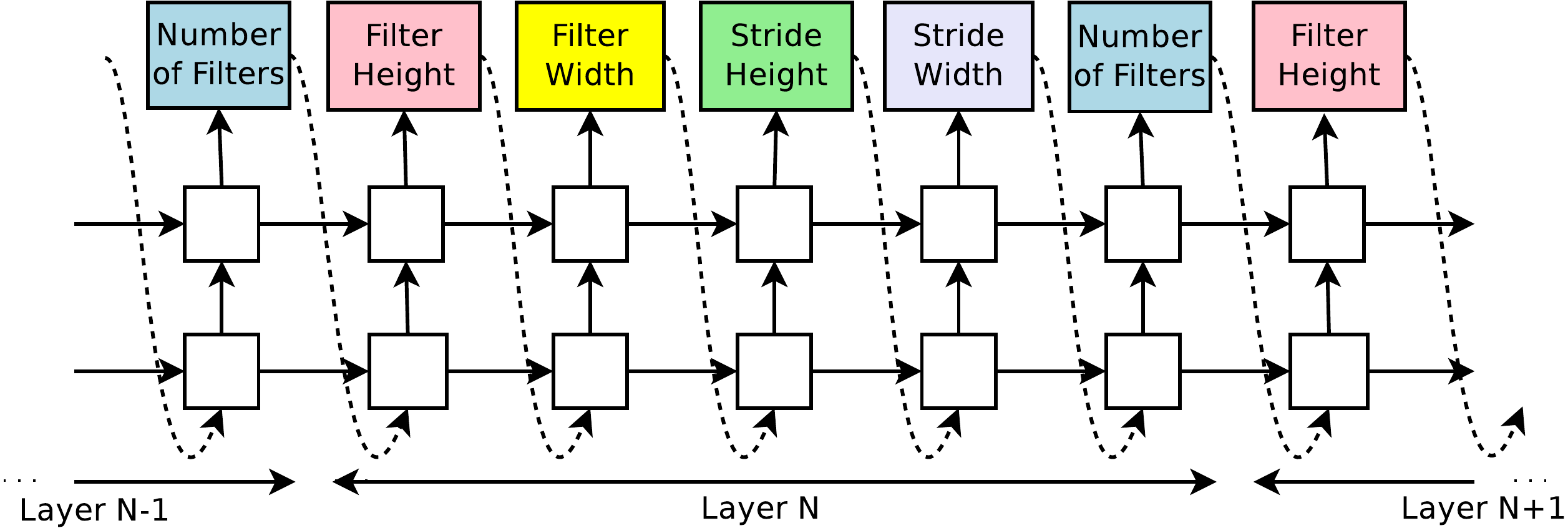}
    \end{center}
    \vspace{-5pt}
    \caption{Illustration of how the recurrent neural network (RNN) controller samples possible convolutional architecture candidates from the search space in reinforcement learning-based NAS. \textbf{(figure from \cite{zoph2016neural})}}
    \vspace{-5pt}
    \label{fig:overview-rl-nas}
\end{figure}

\textbf{Reinforcement Learning-Based Search.} 
In the field of NAS, \cite{zoph2016neural} is the first NAS work\footnote{MetaQNN \cite{baker2016designing} is another seminal NAS work in parallel to \cite{zoph2016neural}, both of which feature reinforcement learning as the search engine to automate the design of top-performing DNNs with competitive accuracy on target task.} that opens up the possibility to automate the design of top-performing DNNs, which features reinforcement learning (RL) \cite{williams1992simple} as the search engine. Specifically, \cite{zoph2016neural} leverages a simple yet effective recurrent neural network (RNN) as the RL controller to generate possible architecture candidates from the search space as shown in Fig.~\ref{fig:overview-rl-nas}. The generated architecture candidate is then trained from scratch on target task to evaluate the accuracy. And next, the accuracy of the generated architecture candidate is fed back into the aforementioned RNN controller, which optimizes the RNN controller to generate better architecture candidates in the next iteration. Once the search process terminates, the well-optimized RNN controller is able to provide DNNs with superior accuracy on target task. For example, the network generated by the RNN controller achieves 96.35\% top-1 accuracy on CIFAR-10, which is comparable to or even better than the family of manually designed DNNs, such as ResNet \cite{he2016deep}. The promising performance of \cite{zoph2016neural} marks an important milestone in the field of NAS, pioneering an effective alternative to automate the design of competitive DNNs.

Subsequently, based on \cite{zoph2016neural}, NASNet \cite{zoph2018nasnet} introduces the flexible cell-based search space as shown in Fig.~\ref{fig:cell-search-space}, which further boosts the attainable accuracy on target task. For example, NASNet achieves 97.6\% top-1 accuracy on CIFAR-10, which is $+$1.25\% higher than \cite{zoph2016neural} while involving fewer parameters (i.e., 37.4\,M in \cite{zoph2016neural} vs. 27.6\,M in NASNet). Despite the promising performance, \cite{zoph2016neural} and NASNet have to train a large number of possible architecture candidates from scratch, thus inevitably necessitating prohibitive computational resources. For example, to optimize the RNN controller, \cite{zoph2016neural} needs to train 12,800 stand-alone architecture candidates. To overcome such limitations, ENAS \cite{pham2018efficient} proposes an efficient NAS paradigm dubbed parameter sharing, which forces all the architecture candidates to share network weights to eschew training each architecture candidate from scratch. In practice, this leads to significant reduction in terms of search cost, while at the same time still maintaining strong accuracy on target task. For example, in \cite{zoph2016neural}, one single search experiment takes 3$\sim$4 days on 450 Nvidia GTX 1080 Ti GPUs \cite{zoph2018nasnet}. In contrast, benefiting from the paradigm of parameter sharing, ENAS is able to find one decent network solution with 97.11\% top-1 accuracy on CIFAR-10, and more importantly, in less than 16 hours on one single Nvidia GTX 1080 Ti GPU. Thanks to the significant search efficiency, the paradigm of parameter sharing has been dominating subsequent breakthroughs in the NAS community \cite{liu2019darts, guo2020single, cai2020once}.

\newlength{\columnsepsave} 
\setlength{\columnsepsave}{\columnsep} 
\setlength{\columnsep}{6pt} 
\begin{wrapfigure}{r}{0.525\columnwidth}
    \begin{center}
        \includegraphics[width=0.5\columnwidth]{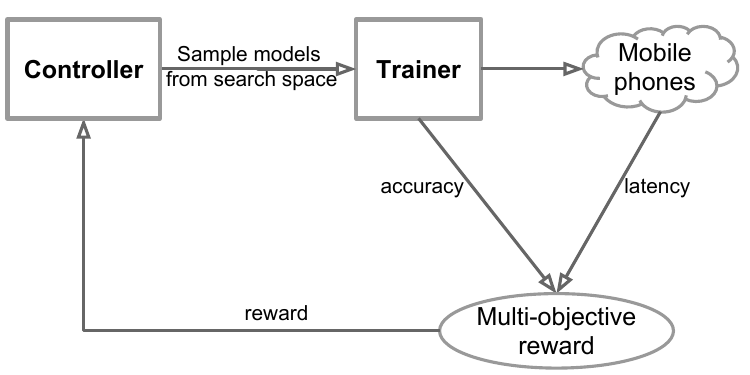}
    \end{center}
    \caption{Overview of MnasNet \cite{tan2019mnasnet}. \textbf{(figure from \cite{tan2019mnasnet})}}
    \label{fig:overview-mnasnet}
\end{wrapfigure}
Although early RL-based NAS methods \cite{zoph2016neural, zoph2018nasnet, pham2018efficient} have made tremendous success in automatic network design, they focus on accuracy-only optimization, which ignore other important performance metrics, such as latency and energy. To search for hardware-efficient network solutions, MnasNet \cite{tan2019mnasnet} formulates the search process as a multi-objective optimization problem that optimizes both accuracy and latency as shown in Fig.~\ref{fig:overview-mnasnet}. To achieve this, MnasNet introduces a flexible block-based search space (see Fig.~\ref{fig:block-search-space}) and designs an effective multi-objective RL reward function to optimize the RNN controller. Specifically, the goal of MnasNet is to find \textit{Pareto-optimal} architecture candidates $arch$ in the search space $\mathcal{A}$ that maximize the pre-defined multi-objective RL reward, which can be formulated as follows:
\begin{equation}
    \label{eq:mnasnet-reward}
    \mathop{\mathrm{maximize}}_{arch \in \mathcal{A}} \,\,\, Accuracy(arch) \times \left[\frac{Latency(arch)}{T}\right]^{w}
\end{equation}
where $Accuracy(\cdot)$ and $Latency(\cdot)$ denote the accuracy on target task and the latency on target hardware, respectively. Besides, $T$ is the specified latency constraint. It is worth noting that the latency $Latency(\cdot)$ in MnasNet is directly measured on target hardware, which suffers from non-trivial engineering efforts due to the prohibitive search space (e.g., $|\mathcal{A}| = \sim10^{39}$ in MnasNet) \cite{cai2019proxylessnas, cai2020once}. To avoid the tedious on-device latency measurements, we later discuss several efficient latency predictors in this section. Apart from these, $w$ is the trade-off coefficient to control the trade-off magnitude between accuracy and latency, which is defined as follows:
\begin{equation}
\begin{aligned}
    w = \begin{cases}
    \alpha,               & \text{if } Latency(arch) \leq T\\
    \beta,                & \text{otherwise}
    \end{cases}
\end{aligned}
\end{equation}
where $\alpha$ and $\beta$ are application-specific hyper-parameters to control the trade-off magnitude between accuracy and efficiency. According to the empirical observation that doubling the latency usually brings $\sim$5\% relative accuracy improvement, MnasNet assigns $\alpha = \beta = -0.07$. In practice, $\alpha$ and $\beta$ are both sensitive and difficult to tune. And even worse, given new hardware devices or new search spaces, $\alpha$ and $\beta$ involve additional engineering efforts for hyper-parameter tuning. For example, as observed in MobileNetV3 \cite{howard2019searching}, the accuracy changes much more dramatically with latency for small networks. Therefore, to obtain the required architecture candidate that satisfies the specified latency constraint $T$, we typically need to repeat 7 search experiments to tune $\alpha$ and $\beta$ through trial and error \cite{bender2020tunas}. This significantly increases the total search cost by $\times$7 times. To eliminate such additional hyper-parameter tuning, TuNAS \cite{bender2020tunas} investigates the multi-objective RL reward in Eq~(\ref{eq:mnasnet-reward}) and further introduces a similar RL reward function, which can be formulated as follows:
\begin{equation}
    \mathop{\mathrm{maximize}}_{arch \in \mathcal{A}} \,\,\, Accuracy(arch) + \gamma \times |\frac{Latency(arch)}{T} - 1|
\end{equation}
where $|\cdot|$ is the absolute function. Besides, $\gamma < 0$ is a finite negative value, which controls how strongly we enforce the architecture candidate to maintain the latency close to $T$.
\setlength{\columnsep}{\columnsepsave}

In addition, MONAS \cite{hsu2018monas} also introduces a simple yet effective RL reward function that considers to optimize both accuracy and energy, which can be formulated as follows:
\begin{equation}
\label{eq:monas-reward}
    Reward(arch) = \eta \times Accuracy(arch) - (1 - \eta) \times Energy(arch)
\end{equation} 
where $\eta \in [0, 1]$ is the coefficient to control the trade-off between accuracy and energy. We note that the RL reward function in Eq~(\ref{eq:monas-reward}) aims to find the architecture candidate with high accuracy and low energy, which can be generalized to other performance constraints like latency.

\textbf{Evolutionary Algorithm-Based Search.}
In addition to reinforcement learning-based search, evolutionary algorithm-based search is another popular branch in the NAS literature, thanks to its flexibility, conceptual simplicity, and competitive performance \cite{real2019regularized}. As seen in the very early evolutionary practices \cite{miller1989designing, angeline1994evolutionary, floreano2008neuroevolution, stanley2002evolving}, the evolutionary algorithm-based search typically consists of four key steps, including (1) sampling a set of possible architecture candidates from the search space as the child population, (2) evaluating the architecture candidates in the child population to interpret the performance, such as accuracy and efficiency, (3) reserving the top-k architecture candidates in the latest child population to form the parent population and discarding the architecture candidates with poor performance, and (4) manipulating the architecture candidates in the latest parent population to generate new architecture candidates to form the next-generation child population. The above four steps are repeated until the evolutionary process converges.

\begin{figure}[t]
    \begin{center}
    \includegraphics[width=1.0\columnwidth]{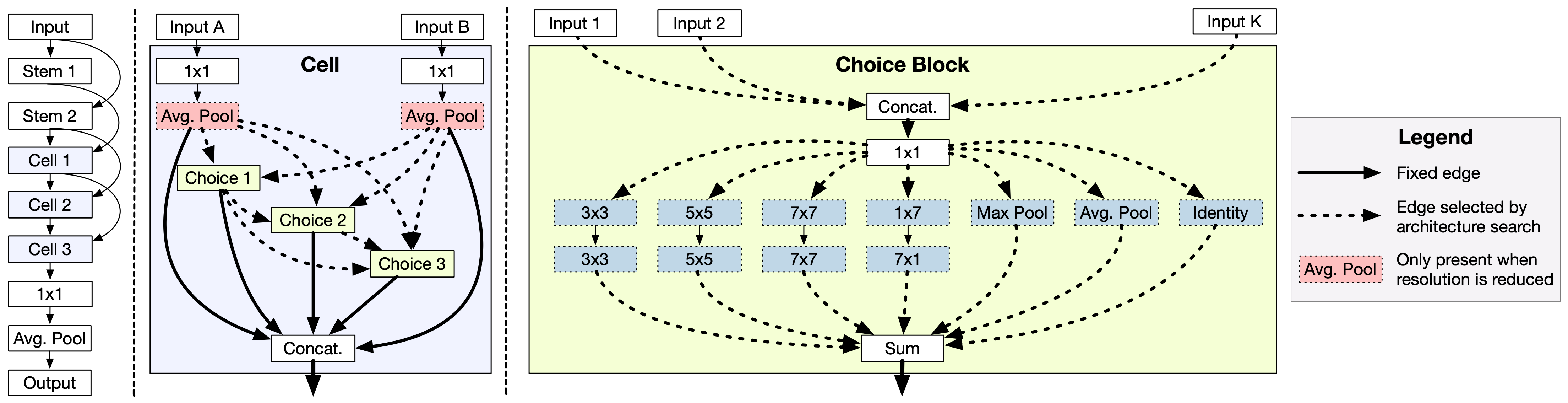}
    \end{center}
    \vspace{-5pt}
    \caption{Overview of the one-shot supernet, where solid lines mean that the operator candidates are enabled while dashed lines mean that the operator candidates are part of the search space but disabled. Here, the one-shot supernet contains all the possible architecture candidates in the search space. \textbf{(figure from \cite{bender2018understanding})}}
    \vspace{-5pt}
    \label{fig:overview-one-shot-nas}
\end{figure}

There are many other aspects in which the evolutionary algorithm may differ, including (1) how to sample the initial population, (2) how to select the parent population, and (3) how to generate the child population from the parent population. Among them, generating the child population from the parent population is of utmost importance in order to produce superior architecture candidates \cite{real2019regularized}. In practice, to allow efficient exploration and exploitation \cite{vcrepinvsek2013exploration}, crossover and mutation are two of the most popular strategies to generate the child population \cite{dominguez2011mutation, spears1995crossover}. Specifically, for crossover, two random architecture candidates from the parent population are crossed to produce one new child architecture candidate. For mutation, one randomly selected architecture candidate mutates its operators with a fixed probability. However, the early evolutionary NAS works have to train a large number of stand-alone architecture candidates from scratch to evaluate the accuracy \cite{real2019regularized}, and as a result, suffer from non-trivial computational resources \cite{bender2018understanding}.

\newlength{\columnsepsavesave} 
\setlength{\columnsepsavesave}{\columnsep} 
\setlength{\columnsep}{6pt} 
\begin{wrapfigure}{r}{0.525\columnwidth}
    \begin{center}
        \includegraphics[width=0.5\columnwidth]{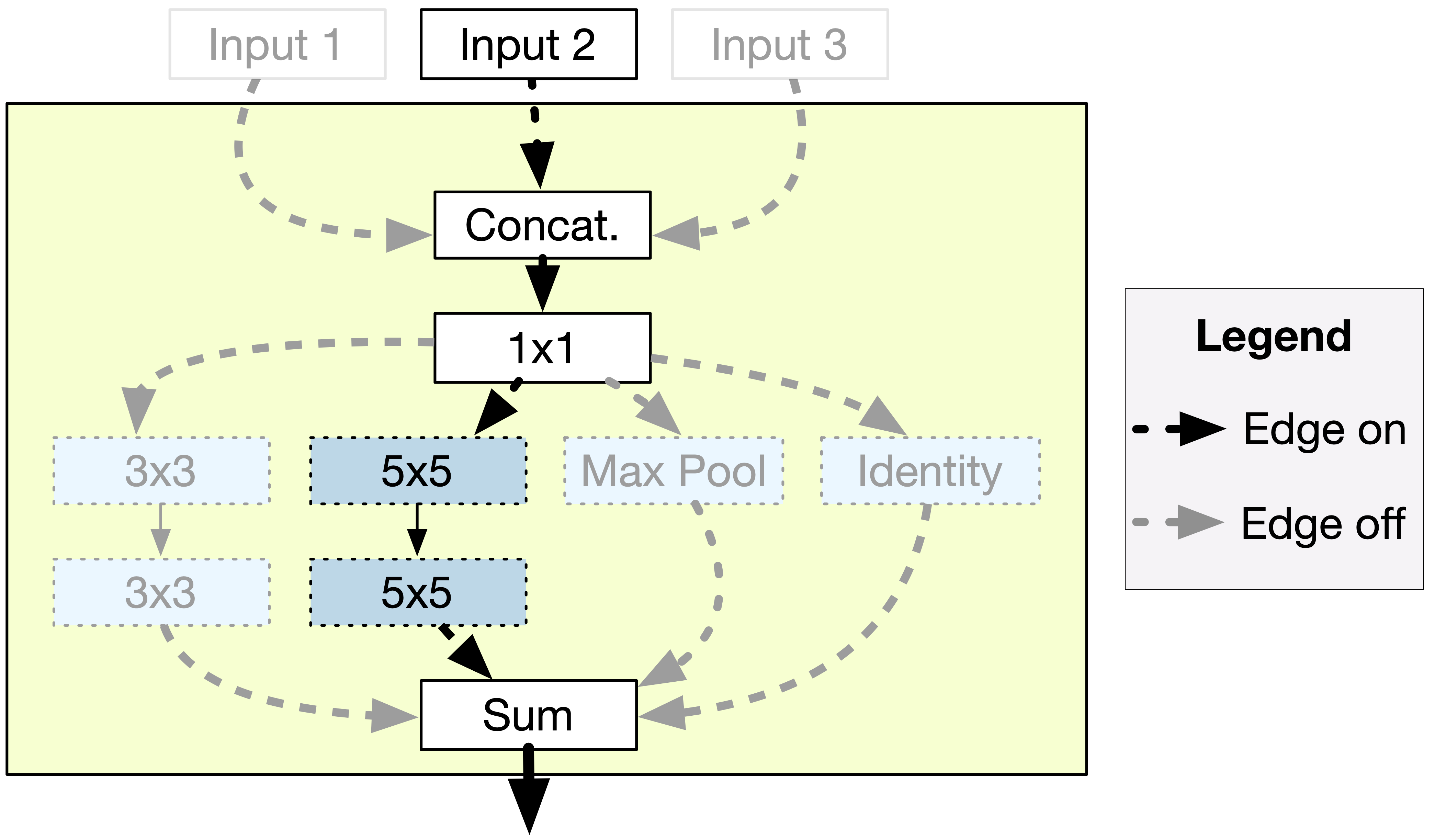}
    \end{center}
    \caption{Illustration of the architecture candidate evaluation in one-shot NAS \cite{bender2018understanding}. \textbf{(figure from \cite{bender2018understanding})}}
    \label{fig:overview-one-shot-nas-eval}
\end{wrapfigure}
To reduce the required computational resources for neural architecture search, \cite{bender2018understanding} introduces the paradigm of one-shot NAS, which has been widely applied in subsequent NAS methods \cite{cai2019proxylessnas, liu2019darts, cai2020once} thanks to its significant search efficiency. In parallel to \cite{bender2018understanding}, SMASH \cite{brock2018smash} also proposes a similar one-shot NAS paradigm, but \cite{bender2018understanding} is much more popular in the NAS community. Specifically, \cite{bender2018understanding} designs an effective one-shot supernet as visualized in Fig.~\ref{fig:overview-one-shot-nas}, which consists of all the possible architecture candidates in the search space. Therefore, we only need to train the one-shot supernet, after which we can evaluate different architecture candidates in the search space with inherited network weights from the pre-trained one-shot supernet as shown in Fig.~\ref{fig:overview-one-shot-nas-eval}. This effectively avoids to train a large number of stand-alone architecture candidates from scratch. In practice, the one-shot supernet is simply trained using the standard SGD optimizer with momentum. Once the one-shot supernet is well trained, it is able to quickly and reliably approximate the performance of different architecture candidates using the paradigm of weight sharing \cite{pham2018efficient}. With the well-trained one-shot supernet, it is straightforward and technically easy to leverage the standard evolutionary algorithm to search for top-performing architecture candidates with superior accuracy on target task \cite{bender2018understanding}. We note that the searched architecture candidates still need to be re-trained or fine-tuned on target task in order to recover the accuracy for further deployment on target hardware.
\setlength{\columnsep}{\columnsepsavesave}

Furthermore, SPOS \cite{guo2020single} investigates the one-shot NAS \cite{bender2018understanding} and identifies two critical issues. On the one hand, the network weights in the one-shot supernet are deeply coupled during the training process. On the other hand, the joint optimization introduces further coupling between architecture candidates and supernet weights. To address these, SPOS proposes the paradigm of single-path one-shot NAS, which uniformly samples one single-path sub-network from the supernet and train the sample single-path sub-network instead. This brings two main benefits, including (1) reducing the memory consumption to the single-path level and (2) improving the performance of the final searched architecture candidate. The success of SPOS has motivated a series of follow-up works \cite{zhang2020fast, chu2021fairnas, cai2020once, yu2020bignas, lu2021one, you2020greedynas, luo2021hsconas, luo2021designing, elsken2019lamarckian}. Note that all of the above follow-up works \cite{zhang2020fast, chu2021fairnas, cai2020once, yu2020bignas, lu2021one, you2020greedynas, luo2021hsconas, luo2021designing, elsken2019lamarckian} focus on training an effective and reliable supernet, which then serves as the evaluator to quickly query the performance of different architecture candidates. For example, FairNAS \cite{chu2021fairnas} demonstrates that the uniform sampling strategy only implies the soft fairness, and to imply the strict fairness, FairNAS samples multiple single-path sub-networks to enforce that all the operator candidates in the supernet are equally optimized during each training iteration. In parallel, OFA \cite{cai2020once} is another representative evolutionary NAS method that aims to train the supernet, after which we are allowed to detach single-path sub-networks from the supernet with inherited network weights for further deployment on target hardware. Note that the detached sub-network in OFA still requires to be fine-tuned on target task for several epochs (e.g., 25 epochs) in order to obtain competitive accuracy. To eliminate the fine-tuning process, BigNAS \cite{yu2020bignas} proposes several enhancements to train one single-stage supernet, where the single-path sub-network detached from the supernet with inherited network weights can achieve superior accuracy without being re-trained or fine-tuned on target task and can be directly deployed on target hardware. This significantly saves the computational resources required for training stand-alone architecture candidates, especially when targeting multiple different deployment scenarios like multiple different hardware platforms.

Thanks to its search flexibility, evolutionary algorithm-based NAS can be easily extended to search for hardware-efficient architecture candidates, which maximize the accuracy on target task while satisfying various real-world performance constraints \cite{zhang2020fast}, such as latency, energy, memory, etc. Without loss of generality, we consider the following multi-objective optimization:
\begin{equation}
    \label{eq:evolutionary-objective}
    \mathop{\mathrm{maximize}}_{arch \in \mathcal{A}} \,\,\, Accuracy(arch) \,\,\,
    s.t., \,\,\, Constraint_1(arch) \leq C_1, ..., Constraint_n(arch) \leq C_n
\end{equation}
where $\{Constraint_i(\cdot)\}_{i=1}^n$ and $\{C_i\}_{i=1}^n$ are a set of real-world performance constraints.

\begin{figure}[t]
    \begin{center}
    \includegraphics[width=1.0\columnwidth]{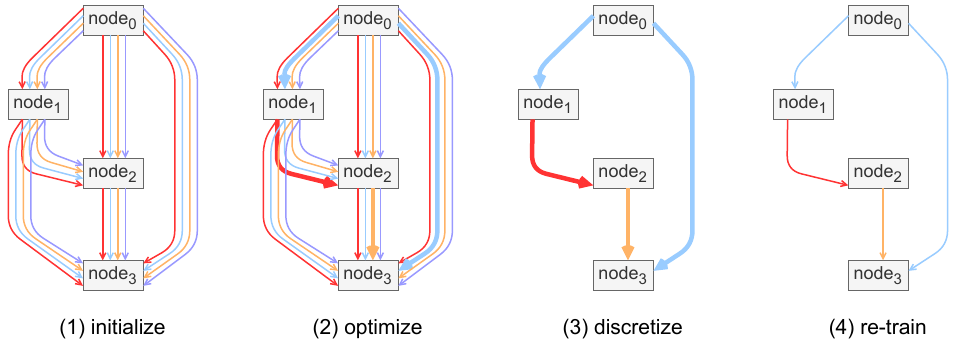}
    \end{center}
    \vspace{-5pt}
    \caption{Overview of DARTS \cite{liu2019darts}, which consists of four stages, including (1) initializing $w$ and $\alpha$ in the supernet, (2) optimizing $w$ and $\alpha$ via alternating gradient descent, (3) discretizing the optimal architecture candidate from the supernet, and (4) re-training the optimal architecture candidate to recover the accuracy.}
    \vspace{-5pt}
    \label{fig:overview-darts}
\end{figure}

\textbf{Gradient-Based Search.}
In addition to reinforcement learning-based search and evolutionary algorithm-based search, gradient-based search \cite{liu2019darts}, also known as differentiable search, is another representative branch of NAS, which has since gained increasing popularity in the NAS community and motivated a plethora of subsequent differentiable NAS works \cite{chen2019pdarts, xu2020pcdarts, liang2019darts+, chu2020fairdarts, chu2021darts-, li2020sgas, yang2021towards, zela2020robustdarts, wang2021rethinking, chen2020drnas, bi2020goldnas, hou2021singledarts, dong2019searching, xie2019snas, mei2020atomnas, ye2022betadarts}, thanks to its significant search efficiency \cite{dong2021automated}. For example, DARTS \cite{liu2019darts}, as the seminal differentiable NAS work, is able to deliver one superior architecture candidate in $\sim$1 day on one single Nvidia GTX 1080 Ti GPU. In contrast to previous non-differentiable NAS practices \cite{zoph2016neural, zoph2018nasnet, pham2018efficient, bender2018understanding} that highly rely on discrete search spaces, DARTS leverages a list of architecture parameters $\alpha$ to relax the discrete search space to become continuous. Benefiting from the continuous search space, both the network weights $w$ and the architecture parameters $\alpha$ can be optimized via alternating gradient descent. Once the differentiable search process terminates, we can interpret the optimal architecture candidate from the architecture parameters $\alpha$. Specifically, the supernet in DARTS is initialized by stacking multiple over-parameterized cells (see Fig.~\ref{fig:overview-darts} (1)), in which each cell consists of all the possible cell structures in the cell-based search space $\mathcal{A}$. As shown in Fig.~\ref{fig:overview-darts}, each cell is represented using the directed acyclic graph (DAG) that consists of $N$ nodes $\{x_i\}_{i=1}^N$. Note that the nodes here correspond to the intermediate feature maps. In addition, the directed edges between $x_i$ and $x_j$ correspond to a list of operator candidates $\{o | o \in \mathcal{O}\}$ in the operator space $\mathcal{O}$. Meanwhile, the directed edges between $x_i$ and $x_j$ are also assigned with a list of architecture parameters $\{\alpha_o^{(i, j)} | o \in \mathcal{O}\}$. Finally, following DARTS, we formulate $x_j$ as follows:
\begin{equation}
    \label{eq:darts-relaxation}
    x_j = \sum_{o \in \mathcal{O}} \frac{\exp \alpha_o^{(i, j)}}{\sum_{o' \in \mathcal{O}} \exp \alpha_{o'}^{(i, j)}} o(x_i)
\end{equation}
Note that the output $x_j$ is continuous with respect to $x_i$, $\alpha$, and $w$. In sight of this, DARTS proposes to optimize $\alpha$ and $w$ using the following bi-level optimization scheme: 
\begin{equation}
    \label{eq:darts-bilevel-optimization}
    \mathop{\mathrm{minimize}}_{\alpha} \,\,\, \mathcal{L}_{val}(w^*(\alpha), \alpha) \,\,\, s.t., \,\,\, w^*(\alpha) = \mathop{\mathrm{arg\,min}}_w \mathcal{L}_{train}(w, \alpha)
\end{equation}
where $\mathcal{L}_{train}(\cdot)$ and $\mathcal{L}_{val}(\cdot)$ are the loss functions on the training and validation datasets, respectively. Once the differentiable search process terminates, DARTS determines the optimal architecture candidate by reserving the strongest operator $\alpha_o^{(i, j)}$ and removing other operators between $x_i$ and $x_j$, in which the operator strength is defined as $\exp \alpha_o^{(i, j)} / \sum_{o' \in \mathcal{O}} \exp \alpha_{o'}^{(i, j)}$. It is worth noting that the searched optimal architecture candidate still needs to be re-trained on target task in order to recover its accuracy for further deployment on target hardware.

Inspired by the promising performance of DARTS, a plethora of follow-up works \cite{chen2019pdarts, xu2020pcdarts, liang2019darts+, chu2020fairdarts, chu2021darts-, li2020sgas, yang2021towards, zela2020robustdarts, wang2021rethinking, chen2020drnas, bi2020goldnas, hou2021singledarts, dong2019searching, xie2019snas, mei2020atomnas, ye2022betadarts} have recently emerged, which strive to unleash the power of differentiable NAS so as to deliver superior architecture candidates. For example, in contrast to DARTS that simultaneously optimizes all the operator candidates in the supernet, PC-DARTS \cite{xu2020pcdarts} introduces partial channel connections to alleviate the excessive memory consumption of DARTS. In addition, DARTS+ \cite{liang2019darts+} investigates the performance collapse issue of DARTS and finds that the performance collapse issue is caused by the over-selection of \textit{skip-connect}. To tackle this, DARTS+ proposes a simple yet effective early-stopping strategy to terminate the search process upon fulfilling a set of pre-defined criteria. In parallel, DARTS- \cite{chu2021darts-} also observes that the performance collapse issue of DARTS comes from the over-selection of \textit{skip-connect} and further leverages an auxiliary skip connection to mitigate the performance collapse issue and stabilize the search process. Apart from these, Single-DARTS \cite{hou2021singledarts} and Gold-NAS \cite{bi2020goldnas} investigate the bi-level optimization in Eq~(\ref{eq:darts-bilevel-optimization}) and point out that the bi-level optimization may end up with sub-optimal architecture candidates, based on which Single-DARTS and Gold-NAS turn back to the one-level optimization. Besides, to accelerate the search process, GDAS \cite{dong2019searching} introduces an efficient Gumbel-Softmax \cite{jang2017categorical} based differentiable sampling approach to reduce the optimization complexity to the single-path level. Similar to GDAS, SNAS \cite{xie2019snas} also leverages Gumbel-Softmax reparameterization to improve the search process, which can make use of gradient information from generic differentiable loss without sacrificing the completeness of NAS pipelines. Furthermore, PT-DARTS \cite{wang2021rethinking} revisits the architecture selection in differentiable NAS and demonstrates that the architecture parameters $\alpha$ cannot always imply the optimal architecture candidate, based on which PT-DARTS introduces the perturbation-based architecture selection to determine the optimal architecture candidate at the end of search.

The aforementioned differentiable NAS works \cite{chen2019pdarts, xu2020pcdarts, liang2019darts+, chu2020fairdarts, chu2021darts-, li2020sgas, yang2021towards, zela2020robustdarts, wang2021rethinking, chen2020drnas, bi2020goldnas, hou2021singledarts, dong2019searching, xie2019snas, mei2020atomnas, ye2022betadarts}, however, focus on accuracy-only neural architecture search, which indeed demonstrates promising performance in terms of finding the architecture candidate with competitive accuracy but fails to accommodate the limited available computational resources in real-world embedded scenarios. To overcome such limitations, the paradigm of hardware-aware differentiable NAS \cite{xu2020latency, luo2020edgenas, li2020lcnas, loni2022tas, kim2021mdarts} has recently emerged, which is based on DARTS and focuses on finding top-performing architecture candidates within the cell-based search space that can achieve both high accuracy on target task and high inference efficiency on target hardware. To achieve the above goal, one widely adopted approach is to integrate the latency-constrained loss term into the overall loss function to penalize the architecture candidate with high latency, which can be mathematically formulated as follows:
\begin{equation}
    \label{eq:hardware-aware-darts-bilevel-optimization}
    \mathop{\mathrm{minimize}}_{\alpha} \,\,\, \mathcal{L}_{val}(w^*(\alpha), \alpha) + \lambda \cdot Latency(\alpha) \,\,\, s.t., \,\,\, w^*(\alpha) = \mathop{\mathrm{arg\,min}}_w \mathcal{L}_{train}(w, \alpha)
\end{equation}
where $\lambda$ is the trade-off coefficient to control the trade-off magnitude between accuracy and latency. As demonstrated in \cite{luo2022you, luo2022lightnas}, a larger $\lambda$ ends up with the architecture candidate that maintains low accuracy and low latency, whereas a smaller $\lambda$ leads to the architecture candidate with high accuracy and high latency. Besides, $Latency(\alpha)$ corresponds to the latency of the architecture candidate encoded by $\alpha$. We note that the optimization objective in Eq~(\ref{eq:hardware-aware-darts-bilevel-optimization}) can be easily generalized to jointly optimize other types of hardware performance constraints, such as energy and memory consumption, in which we only need to incorporate $Energy(\alpha)$ and $Memory(\alpha)$ into the optimization objective in Eq~(\ref{eq:hardware-aware-darts-bilevel-optimization}). For example, we can re-formulate the optimization objective in Eq~(\ref{eq:hardware-aware-darts-bilevel-optimization}) as follows to jointly optimize the on-device latency, energy, and memory consumption:
\begin{equation}
    \label{eq:hardware-aware-darts-bilevel-optimization-pro}
    \mathop{\mathrm{minimize}}_{\alpha} \,\,\, \mathcal{L}_{val}(w^*(\alpha), \alpha) + \lambda_1 \cdot Latency(\alpha) + \lambda_2 \cdot Energy(\alpha) + \lambda_3 \cdot Memory(\alpha)
\end{equation}
where $\lambda_1$, $\lambda_2$, and $\lambda_3$ are trade-off coefficients to determine the trade-off magnitudes between accuracy and latency, energy, and memory, respectively.

\begin{figure}[t]
    \begin{center}
    \includegraphics[width=1.0\columnwidth]{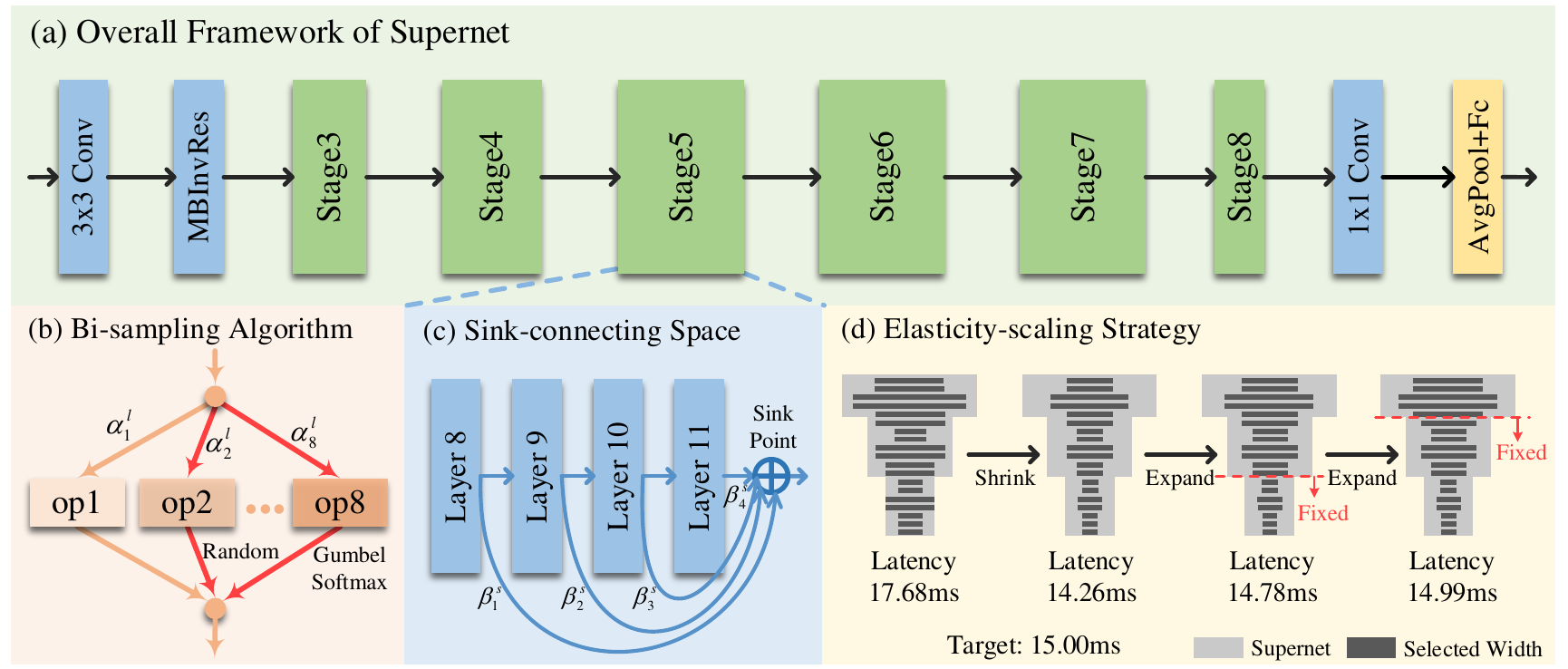}
    \end{center}
    \vspace{-5pt}
    \caption{Overview of TF-NAS \cite{hu2020tf}, which investigates the three search freedoms in conventional hardware-aware differentiable NAS, including (1) \textit{operator-level}, (2) \textit{depth-level}, and (3) \textit{width-level}. \textbf{(figure from \cite{hu2020tf})}}
    \vspace{-5pt}
    \label{fig:overview-tfnas}
\end{figure}

Despite the significant progress to date, the aforementioned hardware-aware differentiable NAS works \cite{xu2020latency, luo2020edgenas, li2020lcnas, loni2022tas, kim2021mdarts} highly rely on the cell-based search space, which first determine the optimal cell structure and then repeatedly stack the same cell structure across the entire network \cite{liu2019darts}. However, as demonstrated in MnasNet \cite{tan2019mnasnet}, such NAS practices suffer from inferior accuracy and efficiency due to the lack of operator diversity. And even worse, the architecture candidates in the cell-based search space are of multiple parallel branches as shown in Fig.~\ref{fig:cell-search-space}, which introduce considerable memory access overheads, and as a result, are difficult to benefit from the high computational parallelism on mainstream hardware platforms \cite{zhang2018shufflenet, ma2018shufflenet}. To overcome such limitations, recent hardware-aware differentiable NAS works \cite{wu2019fbnet, wan2020fbnetv2, stamoulis2020single, lee2021snas, cai2019proxylessnas, hu2020tf, luo2022surgenas, fang2020densenas, mei2020atomnas} have shifted their attention from the cell-based search space (see Fig.~\ref{fig:cell-search-space}) to the block-based search space (see Fig.~\ref{fig:block-search-space}). Among them, the most representative ones include FBNet \cite{wu2019fbnet}, ProxylessNAS \cite{cai2019proxylessnas}, SP-NAS \cite{stamoulis2020single}, and TF-NAS \cite{hu2020tf}. Specifically, similar to GDAS \cite{dong2019searching} and SNAS \cite{xie2019snas}, FBNet leverages Gumbel-Softmax reparameterization \cite{jang2017categorical} to relax the discrete search space to be continuous. Besides, FBNet collects a simple yet effective latency lookup table to quickly approximate the latency of different architecture candidates. The pre-collected latency lookup table is then integrated into the search process to derive hardware-efficient architecture candidates. However, similar to DARTS \cite{liu2019darts}, FBNet requires to simultaneously optimize all the operator candidates in the supernet during the search process, which is not scalable to large search spaces and suffers from the memory bottleneck \cite{cai2019proxylessnas, luo2022you}. In sight of this, ProxylessNAS introduces an effective path-level binarization approach to reduce the memory consumption to the single-path level, which significantly improves the search efficiency without compromising the search accuracy. In parallel, SP-NAS demonstrates that different operator candidates in the supernet can be viewed as subsets of an over-parameterized superkernel, based on which SP-NAS proposes to encode all the operator candidates into the superkernel. In practice, this explicitly reduces the memory consumption to the single-path level, which therefore alleviates the memory bottleneck during the search process. Furthermore, TF-NAS thoroughly investigates the three search freedoms in hardware-aware differentiable NAS, including (1) \textit{operator-level} search, (2) \textit{depth-level} search, and (3) \textit{width-level} search as shown in Fig.~\ref{fig:overview-tfnas}, which is able to perform fine-grained architecture search. Besides, to obtain hardware-efficient architecture candidates, TF-NAS integrates the pre-collected latency lookup table into the search process. In the meantime, TF-NAS introduces a simple yet effective bi-sampling search algorithm to accelerate the search process towards enhanced search efficiency.

But even so, we should consider not only the \textit{explicit} search cost - the time required for one single search experiment, but also the \textit{implicit} search cost - the time required for manual hyper-parameter tuning in order to find the desired architecture candidate. This is because, in real-world embedded scenarios like autonomous vehicles, DNNs must be executed under strict latency constraints (e.g., 24\,ms), in which any violation may lead to catastrophic consequences \cite{bender2020tunas, liu2022bringing}. However, to find the architecture candidate with the latency of 24\,ms, the aforementioned hardware-aware differentiable NAS works \cite{xu2020latency, luo2020edgenas, li2020lcnas, loni2022tas, kim2021mdarts, wu2019fbnet, wan2020fbnetv2, stamoulis2020single, lee2021snas, cai2019proxylessnas, hu2020tf, luo2022surgenas, fang2020densenas, mei2020atomnas} have to repeat a plethora of search experiments to tune the trade-off coefficient $\lambda$ (see Eq~(\ref{eq:hardware-aware-darts-bilevel-optimization})) through trial and error \cite{luo2022you, luo2022lightnas}, which significantly increases the total search cost. The intuition behind this is that $\lambda$, despite being able to trade off between accuracy and latency, is quite sensitive and difficult to control \cite{luo2022you, luo2022lightnas}. To overcome such limitations, HardCoRe-NAS \cite{nayman2021hardcore} leverages an elegant Block Coordinate Stochastic Frank-Wolfe (BCSFW) algorithm \cite{lacoste2013block-frankwolfe} to restrict the search direction around the specified latency requirement. In addition, LightNAS \cite{luo2022you, luo2022lightnas} introduces a simple yet effective hardware-aware differentiable NAS approach, which investigates the optimization objective in Eq~(\ref{eq:hardware-aware-darts-bilevel-optimization}) and proposes to optimize the trade-off coefficient $\lambda$ during the search process in order to satisfy the specified latency requirement. In other words, LightNAS focuses on automatically learning $\lambda$ that strictly complies with the specified latency requirement, which is able to find the required architecture candidate in one single search (i.e., \textit{you only search once}) and avoids performing manual hyper-parameter tuning over $\lambda$. Specifically, the optimization objective of LightNAS is formulated as follows:
\begin{equation}
    \label{eq:optimization-objective-lightnas}
    \mathop{\mathrm{minimize}}_{\alpha} \,\,\, \mathcal{L}_{val}(w^*(\alpha), \alpha) + \lambda \cdot \left( \frac{Latency(\alpha)}{T} - 1\right)\,\,\,s.t., \,\,\, w^*(\alpha) = \mathop{\mathrm{arg\,min}}_w \mathcal{L}_{train}(w, \alpha)
\end{equation}
where $T$ is the specified latency requirement. Different from previous hardware-aware differentiable NAS works \cite{xu2020latency, luo2020edgenas, li2020lcnas, loni2022tas, kim2021mdarts, wu2019fbnet, wan2020fbnetv2, stamoulis2020single, lee2021snas, cai2019proxylessnas, hu2020tf, luo2022surgenas, fang2020densenas, mei2020atomnas}, $\lambda$ in Eq~(\ref{eq:optimization-objective-lightnas}) is not a constant but a learnable hyper-parameter that can be automatically optimized during the search process. For the sake of simplicity, below we use $\mathcal{L}(w, \alpha, \lambda)$ to denote the optimization objective in Eq~(\ref{eq:optimization-objective-lightnas}). Finally, to satisfy the specified latency requirement (i.e., $Latency(\alpha) = T$), $w$ and $\alpha$ are updated using gradient descent \cite{liu2019darts}, whereas $\lambda$ is updated using gradient ascent as follows:
\begin{equation}
    \label{eq:parameter-update-lightnas}
    \begin{cases}
        w^* = w - lr_w \cdot \frac{\partial\mathcal{L}(w, \alpha, \lambda)}{\partial w}, \,\, \alpha^* = \alpha - lr_{\alpha} \cdot \frac{\partial \mathcal{L}(w, \alpha, \lambda)}{\partial \alpha} \\ 
        \lambda^* = \lambda + lr_{\lambda} \cdot \frac{\partial \mathcal{L}(w, \alpha, \lambda)}{\partial \lambda} = \lambda + lr_{\lambda} \cdot \left(\frac{LAT(\alpha)}{T} - 1\right)
    \end{cases}
\end{equation}
where $lr_{w}$, $lr_{\alpha}$, and $lr_{\lambda}$ are the learning rates of $w$, $\alpha$, and $\lambda$, respectively. Below we further demonstrate why LightNAS guarantees $Latency(\alpha) = T$. As shown in LightNAS, a larger $\lambda$ leads to the architecture candidate with low latency, whereas a smaller $\lambda$ results in the architecture candidate with high latency. Therefore, if $Latency(\alpha) > T$, the gradient ascent scheme increases $\lambda$ to reinforce the latency regularization magnitude. As a result, $Latency(\alpha)$ decreases towards $T$ in the next search iteration. Likewise, if $Latency(\alpha) < T$, the gradient ascent scheme decreases $\lambda$ to diminish the latency regularization magnitude, after which $Latency(\alpha)$ increases towards $T$ in the next search iteration. Finally, the search engine ends up with the architecture candidate that strictly satisfies the specified latency requirement (i.e., $Latency(\alpha) = T$). More recently, Double-Win NAS \cite{luo2024double, luo2025exploring} proposes deep-to-shallow transformable search to further marry the best of both deep and shallow networks towards an aggressive accuracy-efficiency win-win. Similar to LightNAS \cite{luo2022you, luo2022lightnas}, the resulting shallow network can also satisfy the specified latency constraint. Finally, we compare previous representative hardware-aware NAS works, which are summarized in Table~\ref{tab:nas-comparisons}.

\begin{table}[t]
\resizebox{\textwidth}{!}{%
\begin{tabular}{l|c|c|ccc|c|c|cc}
\toprule[0.175em]
\multirow{2}{*}{Method} &
  \multirow{2}{*}{\makecell[c]{Search \\ Space}} &
  \multirow{2}{*}{\makecell[c]{Search \\ Strategy}} &
  \multicolumn{3}{c|}{Search Cost} &
  \multirow{2}{*}{\makecell[c]{Target \\ Hardware}} &
  \multirow{2}{*}{\makecell[c]{Hardware \\ Modeling}} &
  \multicolumn{2}{c}{ImageNet} \\ \cline{4-6} \cline{9-10} 
 &
   &
   &
  \multicolumn{1}{c|}{Dataset} &
  \multicolumn{1}{c|}{GPU-Hours} &
  GPU &
   &
   &
  \multicolumn{1}{c|}{FLOPs (M)} &
  Top-1 Acc (\%) \\ \hline
 MnasNet \cite{tan2019mnasnet} & Block & Reinforce & \multicolumn{1}{c|}{ImageNet} & \multicolumn{1}{c|}{40,000} & V100 & Mobile Phones & N/A & \multicolumn{1}{c|}{312} & 75.2 \\ \hline
 ProxylessNAS \cite{cai2019proxylessnas} & Block & Gradient & \multicolumn{1}{c|}{ImageNet} & \multicolumn{1}{c|}{200} & V100 & \makecell[c]{GPUs, CPUs, and \\ Mobile Phones} & LUT & \multicolumn{1}{c|}{N/A} & 75.1 \\ \hline
 MobileNetV3 \cite{howard2019searching} & Block & Evolution & \multicolumn{1}{c|}{ImageNet} & \multicolumn{1}{c|}{N/A} & N/A & Mobile Phones & N/A & \multicolumn{1}{c|}{219} & 75.2 \\ \hline
 FBNet \cite{wu2019fbnet} & Block & Gradient & \multicolumn{1}{c|}{ImageNet} & \multicolumn{1}{c|}{216} & N/A & Mobile Phones & LUT & \multicolumn{1}{c|}{375} & 74.9 \\ \hline
 TuNAS \cite{bender2020tunas} & Block & Reinforce & \multicolumn{1}{c|}{ImageNet} & \multicolumn{1}{c|}{N/A} & N/A & Mobile Phones & LUT & \multicolumn{1}{c|}{-} & 75.4 \\ \hline
 OFA \cite{cai2020once} & Block & Evolution & \multicolumn{1}{c|}{ImageNet} & \multicolumn{1}{c|}{1,200} & V100 & \makecell[c]{GPUs, CPUs, \\ Edge GPUs, and \\ Mobile Phones} & LUT & \multicolumn{1}{c|}{230} & 76.0 \\ \hline
 SP-NAS \cite{stamoulis2020single} & Block & Gradient & \multicolumn{1}{c|}{ImageNet} & \multicolumn{1}{c|}{30} & TPU & Mobile Phones & LUT & \multicolumn{1}{c|}{N/A} & 75.0 \\ \hline
 LA-DARTS \cite{xu2020latency} & Cell & Gradient & \multicolumn{1}{c|}{CIFAR-10} & \multicolumn{1}{c|}{17} & P100 & GPUs and CPUs & Predictor & \multicolumn{1}{c|}{575} & 74.8 \\ \hline
 MDARTS \cite{kim2021mdarts} & Cell & Gradient & \multicolumn{1}{c|}{CIFAR-10} & \multicolumn{1}{c|}{$\sim$6.5} & Titan XP & Eyeriss & Predictor & \multicolumn{1}{c|}{N/A} & N/A \\ \hline
 EH-DNAS \cite{jiang2021ehdnas} & Cell & Gradient & \multicolumn{1}{c|}{CIFAR-10} & \multicolumn{1}{c|}{24} & 1080 Ti & \makecell[c]{Customized \\ Accelerators} & Predictor & \multicolumn{1}{c|}{840} & 69.6 \\ \hline
 E-DNAS \cite{lopez2021ednas} & Block & Gradient & \multicolumn{1}{c|}{ImageNet} & \multicolumn{1}{c|}{N/A} & V100 & CPUs and DSPs & Predictor & \multicolumn{1}{c|}{365} & 76.9 \\ \hline
 SNAS \cite{lee2021snas} & Block & Gradient & \multicolumn{1}{c|}{ImageNet} & \multicolumn{1}{c|}{30} & N/A & TPUs & Predictor & \multicolumn{1}{c|}{1290} & 79.4 \\ \hline
 HSCoNAS \cite{luo2021hsconas} & Block & Evolution & \multicolumn{1}{c|}{ImageNet} & \multicolumn{1}{c|}{N/A} & N/A & \makecell[c]{GPU, CPUs, and \\ Edge GPUs} & LUT & \multicolumn{1}{c|}{N/A} & 74.9  \\ \hline
 DenseNAS \cite{fang2020densenas} & Block & Gradient & \multicolumn{1}{c|}{ImageNet} & \multicolumn{1}{c|}{64} & Titan XP & GPUs & LUT & \multicolumn{1}{c|}{361} & 75.3 \\ \hline
 TF-NAS \cite{hu2020tf} & Block & Gradient & \multicolumn{1}{c|}{ImageNet} & \multicolumn{1}{c|}{43} & Titan RTX & GPUs & LUT & \multicolumn{1}{c|}{284} & 75.2 \\ \hline
 HardCoRe-NAS \cite{nayman2021hardcore} & Block & Gradient & \multicolumn{1}{c|}{ImageNet} & \multicolumn{1}{c|}{400} & P100 & GPUs and CPUs & LUT & \multicolumn{1}{c|}{N/A} & 75.7 \\ \hline
 LightNAS \cite{luo2022you} & Block & Gradient & \multicolumn{1}{c|}{ImageNet} & \multicolumn{1}{c|}{10} & RTX 3090 & Edge GPUs & Predictor & \multicolumn{1}{c|}{N/A} & 75.2 \\ \hline
 SurgeNAS \cite{luo2022surgenas} & Block & Gradient & \multicolumn{1}{c|}{ImageNet} & \multicolumn{1}{c|}{30} & V100 & \makecell[c]{GPUs, CPUs, \\ and Edge GPUs} & Predictor & \multicolumn{1}{c|}{N/A} & 75.5 \\ \hline
 SPOS \cite{guo2020single} & Block & Evolution & \multicolumn{1}{c|}{ImageNet} & \multicolumn{1}{c|}{288} & 1080 Ti & GPUs & LUT & \multicolumn{1}{c|}{328} & 74.7 \\ \hline
 HURRICANE \cite{zhang2020fast} & Block & Evolution & \multicolumn{1}{c|}{ImageNet} & \multicolumn{1}{c|}{N/A} & N/A & \makecell[c]{CPUs, DSPs, \\ and VPUs} & LUT & \multicolumn{1}{c|}{409} & 75.1 \\ \hline
 ProxyNAS \cite{lu2021one} & Block & Evolution & \multicolumn{1}{c|}{ImageNet} & \multicolumn{1}{c|}{N/A} & N/A & \makecell[c]{GPUs, CPUs, \\ TPUs, and FPGAs} & Predictor & \multicolumn{1}{c|}{N/A} & N/A \\ 
 \toprule[0.175em]
\end{tabular}%
}
\caption{Comparisons of representative hardware-aware NAS works. This table is to \textit{roughly} compare different hardware-aware NAS works, in which N/A means that the related data is not reported in the respective paper. Note that the accuracy in this table may be trained under different training recipes.}
\vspace{-5pt}
\label{tab:nas-comparisons}
\end{table}

\subsection{Speedup Techniques and Extensions}
\label{sec:speedup-techniques-and-extensions}

In this section, we further discuss recent state-of-the-art advances in general speedup techniques and extensions for NAS algorithms, including one-shot NAS enhancements, efficient latency prediction, efficient accuracy prediction, low-cost proxies, zero-cost proxies, efficient transformer search, efficient domain-specific search, and mainstream NAS benchmarks, which have the potential to significantly benefit NAS algorithms and largely facilitate the search process.

\textbf{Beyond One-Shot NAS.}
Despite the high search efficiency, one-shot NAS often suffers from poor ranking correlation between one-shot search and stand-alone training. As pointed out in \cite{yu2019evaluating}, one-shot search results do not necessarily correlate with stand-alone training results across various search experiments. To overcome such limitations, a plethora of one-shot NAS enhancements have been recently proposed \cite{zhao2021few, hu2022generalizing, xu2022few, lyunderstanding, su2021kshotnas, zhou2022close}. Specifically, \cite{zhao2021few, hu2022generalizing, xu2022few, lyunderstanding, su2021kshotnas} turn back to few-shot NAS. In contrast to one-shot NAS \cite{pham2018efficient} that only features one supernet, few-shot NAS further introduces multiple supernets to explore different regions of the pre-defined search space, which slightly increases the search cost over one-shot NAS but can deliver much more reliable search results. For example, as shown in \cite{zhao2021few}, with only up to 7 supernets, few-shot NAS can establish new state-of-the-art search results on ImageNet. Among them, \cite{lyunderstanding} demonstrates that zero-cost proxies can be integrated into few-shot NAS, which can further enhance the search process of one-shot NAS and thus end up with better search results. More recently, \cite{xu2022few} generalizes few-shot NAS to distill large language models, which focuses on automatically distilling multiple compressed student models under various computational budgets from a large teacher model. In contrast to few-shot NAS that leverages multiple supernets to improve the ranking correlation performance of one-shot NAS, CLOSE \cite{zhou2022close} instead features an effective curriculum learning-like schedule to control the parameter sharing extent within the proposed supernet dubbed CLOSENet, in which the parameter sharing extent can be flexibly adjusted during the search process and the parameter sharing scheme is built upon an efficient graph-based encoding scheme.

\textbf{Efficient Latency Prediction\footnote{We mainly discuss latency prediction since latency is the most dominant performance constraint in hardware-aware NAS \cite{laube2022expect, lu2021one}, which can be generalized to predict other performance constraints, such as energy and memory consumption.}.}
As seen in MnasNet \cite{tan2019mnasnet}, the latency is directly measured on target hardware, which is then integrated into the RL reward (see Eq~(\ref{eq:mnasnet-reward})) to penalize the architecture candidate with high latency. The direct on-device latency measurement is indeed accurate, which, however, is time-consuming and unscalable to large search spaces \cite{cai2019proxylessnas}. To overcome such limitations, several latency prediction strategies have been recently proposed. For example, ProxylessNAS \cite{cai2019proxylessnas}, FBNet \cite{wu2019fbnet}, and OFA \cite{cai2020once} leverage the latency lookup table to approximate the on-device latency, which sums up the latency of all the operator candidates. In addition, HSCoNAS \cite{luo2021hsconas, luo2021designing} demonstrates that the data movements and communications among different operator candidates introduce additional latency overheads, making the pre-collected latency lookup table inaccurate. To mitigate this issue, HSCoNAS quantifies the latency that corresponds to the intermediate data movements and communications, which is then fed into the pre-collected latency lookup table to achieve more accurate latency prediction performance. However, the latency lookup table is only applicable to the block-based search space, which leads to unreliable latency prediction performance in terms of the cell-based search space \cite{dudziak2020brp}. To this end, EdgeNAS \cite{luo2020edgenas}, LA-DARTS \cite{xu2020latency}, and LC-NAS \cite{li2020lcnas} propose to use learning-based approaches for the latency prediction purpose. For example, EdgeNAS trains an efficient multi-layer perceptron (MLP) to predict the latency of different architecture candidates in the cell-based search space, which can also be generalized to predict the latency of different architecture candidates in the block-based search space as shown in \cite{luo2022you, luo2022lightnas, laube2022expect, lu2021one, li2021hwnasbench}. Furthermore, BRP-NAS \cite{dudziak2020brp} and SurgeNAS \cite{luo2022surgenas} introduce graph neural networks (GNNs) based latency predictors to achieve more reliable latency prediction performance. In practice, the above latency predictors (1) rely on a large number of training samples to achieve decent latency prediction performance (e.g., 100,000 training samples in EdgeNAS) and (2) need to be reconstructed for either new hardware or new search spaces. To avoid these, HELP \cite{lee2021hardware} and MAPLE-Edge \cite{nair2022maple} focus on building an efficient latency predictor using only few training samples (e.g., as few as 10 training samples in HELP), which can be generalized to new hardware or new search spaces with only minimal re-engineering efforts. More recently, EvoLP \cite{huai2023evolp} considers an effective self-evolving scheme to construct efficient yet accurate latency predictors, which can adapt to unseen hardware with only minimal re-engineering efforts.

\textbf{Efficient Accuracy Prediction.}
In parallel to latency prediction, accuracy prediction has also received increasing attention from the NAS community \cite{wen2020neural, luo2020accuracy, white2021powerful, moons2021donna, dudziak2020brp, ning2020generic}, which strives to directly predict the accuracy of different architecture candidates in the search space. Specifically, \cite{wen2020neural} introduces a simple yet effective graph convolutional networks (GCNs) based accuracy predictor, which can achieve reliable accuracy prediction performance, thanks to GCNs' strong capability to learn graph-structured data. Similar to \cite{wen2020neural}, BRP-NAS \cite{dudziak2020brp} also considers GCNs for reliable accuracy prediction, which introduces transfer learning to further improve the accuracy prediction performance from the pre-trained latency predictor. In parallel, \cite{luo2020accuracy} leverages the non-neural network (i.e., GBDT) as the accuracy predictor, which is of stronger capability to learn representations than neural networks based accuracy predictors. In addition, NASLib \cite{white2021powerful} investigates a wide range of accuracy predictors from learning curve extrapolation, weight-sharing, supervised learning, and zero-cost proxies on three popular NAS benchmarks (i.e., NAS-Bench-101 \cite{ying2019nasbench101}, NAS-Bench-201 \cite{dong2020nasbench201}, and NAS-Bench-NLP \cite{klyuchnikov2022nasbenchnlp}). In particular, NASLib reveals that different accuracy predictors can be combined to achieve substantially better accuracy prediction performance than any single accuracy predictor. Furthermore, DONNA \cite{moons2021donna} proposes to build an efficient accuracy predictor, which only involves minimal computational resources, and more importantly, can scale to diverse search spaces. To achieve this, DONNA uses blockwise knowledge distillation to construct an architecture candidate pool, in which each architecture candidate only needs to be fine-tuned for several epochs to derive the accuracy rather than being trained from scratch. Different from the aforementioned accuracy predictors that feature graph-based encoding schemes, GATES \cite{ning2020generic, ning2022generic} instead models the operations as the transformation of the propagating information, which can effectively mimic the actual data processing of different neural architecture candidates. More importantly, the encoding scheme of GATES can be integrated into the above accuracy predictors to further boost their accuracy prediction performance. Similar to GATES, TA-GATES \cite{ning2022tagates} also introduces an effective encoding scheme with analogous modeling of the training process of different neural architecture candidates, which can further achieve better accuracy prediction performance than GATES on various representative NAS benchmarks.

\begin{figure}[t]
    \begin{center}
    \includegraphics[width=1.0\columnwidth]{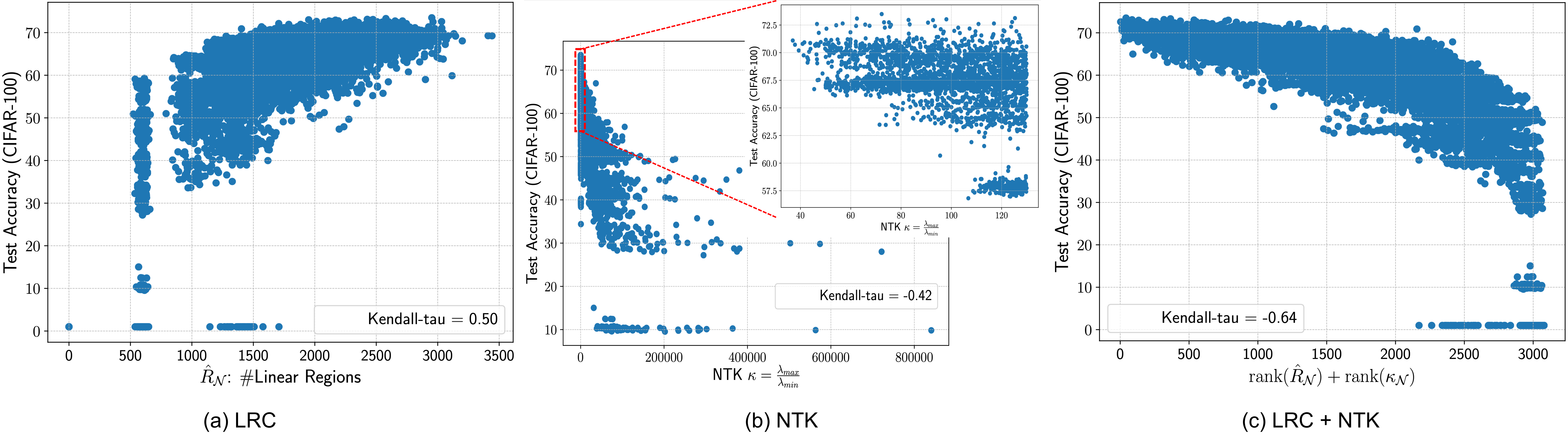}
    \end{center}
    \vspace{-5pt}
    \caption{Comparisons of different zero-cost proxies on CIFAR-100, including (a) LRC \cite{xiong2020linearregions}, (b) NTK \cite{xiao2020ntk}, and (c) LRC + NTK \cite{chen2021neural}, in which the accuracy is queried from NAS-Bench-201 \cite{dong2020nasbench201}. \textbf{(figure from \cite{chen2021neural})}}
    \vspace{-5pt}
    \label{fig:lrc-ntk-correlation-cifar100-nasbench201}
\end{figure}

\textbf{Low-Cost Proxies (Learning Curve Extrapolations).}
Low-cost proxies, also referred to as Learning curve extrapolations \cite{domhan2015speeding}, aim to interpret the accuracy of the given architecture candidate only using its early training statistics, such as the training loss in the first few training epochs, which has motivated a plethora of subsequent works to continue exploring learning curve extrapolation \cite{ru2021speedy, yan2021nasbenchx11, zhao2022loss, klein2017learning, baker2017accelerating, luo2022work, luo2022lightnas}. For example, different from the conventional accuracy predictor that only uses the network configuration as input features, \cite{baker2017accelerating} proposes to combine the network configuration and a series of validation accuracy in the first few training epochs as input features to train a simple regression model, which can be generalized to predict the accuracy of unseen architecture candidates. In addition, \cite{ru2021speedy} introduces \textit{Training Speed Estimation} (TSE), which simply accumulates the early training statistics to achieve reliable yet computationally cheap ranking among different architecture candidates. Besides, \cite{luo2022lightnas, luo2022work} introduce \textit{Batchwise Training Estimation} (BTE) and \textit{Trained Batchwise Estimation} (TBE), which both consider the fine-grained batchwise training statistics to provide more reliable prediction performance using minimal computational resources. In parallel, \cite{zhao2022loss} introduces \textit{Loss Curve Gradient Approximation} (LCGA) to rank the accuracy of different architecture candidates with minimal training. Furthermore, \cite{yan2021nasbenchx11} introduces NAS-Bench-x11 to unleash the power of learning curve extrapolation by predicting the training trajectories, which can be easily integrated into the aforementioned learning curve extrapolation works to quickly estimate the performance of the given architecture candidate.

\textbf{Zero-Cost Proxies\footnote{Note that most of the covered zero-cost proxies are available at \url{https://github.com/automl/naslib/tree/zerocost}.}.}
In addition to the above low-cost proxies (i.e., learning curve extrapolation), zero-cost proxies have recently flourished \cite{abdelfattah2021zero, krishnakumar2022bench, lopes2021epenas, turner2019fisher, wang2020grasp, mellor2021neural, lee2018snip, tanaka2020synflow, lin2021zenscore, akhauri2022eznas, chen2021neural, xiong2020linearregions, xiao2020ntk}, which focus on interpreting the performance of the given architecture candidate in training-free manners. Specifically, zero-cost proxies, such as EPE \cite{lopes2021epenas}, Fisher \cite{turner2019fisher}, GradNorm \cite{abdelfattah2021zero}, Grasp \cite{wang2020grasp}, Jacov \cite{mellor2021neural}, Snip \cite{lee2018snip}, Synflow \cite{tanaka2020synflow}, ZenScore \cite{lin2021zenscore}, LRC \cite{xiong2020linearregions}, and NTK \cite{xiao2020ntk}, can provide reliable performance estimation using only one single mini-batch of data and one single forward/backward propagation pass, which necessitate near-zero computational cost \cite{abdelfattah2021zero, krishnakumar2022bench, chen2021neural}. Thanks to their reliable performance estimation and low cost, these zero-cost proxies have been widely adopted in recent NAS works to accelerate the search process \cite{chen2021neural, akhauri2022eznas, mellor2021neural}. In particular, as demonstrated in \cite{abdelfattah2021zero, chen2021neural}, combining different zero-cost proxies may lead to more reliable ranking performance estimation than any single zero-cost proxy. For example, as shown in Fig.~\ref{fig:lrc-ntk-correlation-cifar100-nasbench201}, combining LRC and NTK is able to provide more reliable ranking performance estimation than LRC or NTK itself. In sight of this, TE-NAS \cite{chen2021neural} further leverages LRC and NTK to jointly estimate the ranking performance among different architecture candidates in the search space, which quickly ends up with the optimal architecture candidate on ImageNet in less than 4 hours on one single Nvidia GTX 1080 Ti GPU.

\newlength{\columnsepsavesavesavesave} 
\setlength{\columnsepsavesavesavesave}{\columnsep} 
\setlength{\columnsep}{6pt} 
\begin{wrapfigure}{r}{0.45\columnwidth}
    \begin{center}
        \includegraphics[width=0.45\columnwidth]{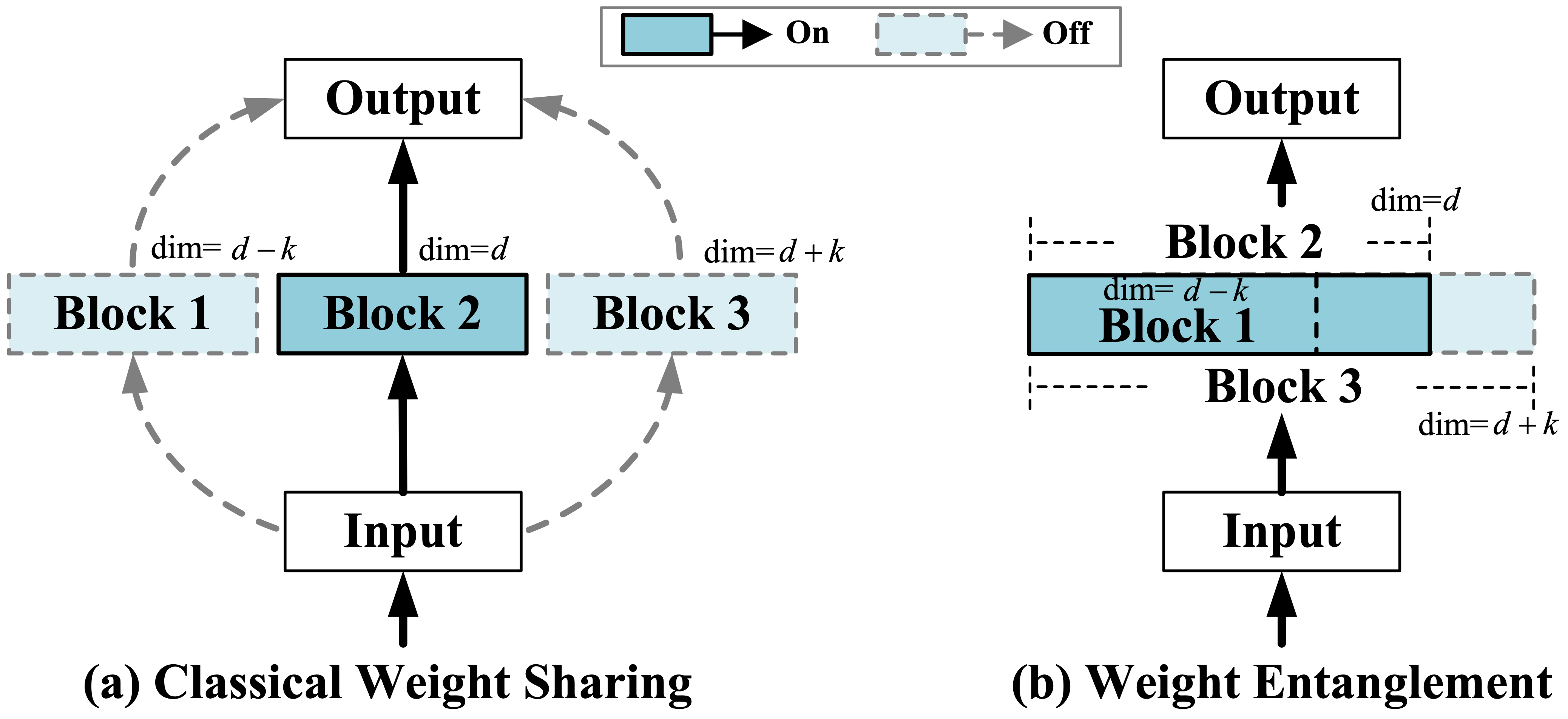}
    \end{center}
    \caption{Illustration of weight sharing \cite{pham2018efficient} and weight entanglement \cite{chen2021autoformer}. \textbf{(figure from \cite{chen2021autoformer})}}
    \label{fig:weight-sharing-vs-weight-entanglement}
\end{wrapfigure}
\textbf{Efficient Transformer Search.}
In addition to CNNs, transformers are another important branch of DNNs. Inspired by the tremendous success of NAS in searching for superior CNNs, automated transformer search has gained increasing popularity, which applies NAS techniques to automatically search for superior transformers, including transformers for NLP tasks \cite{wang2020hat, gao2022autobert, so2021primer, yin2021autotinybert, xu2021bert, luo2021lightspeech, kim2020evolved} and vision transformers for vision tasks \cite{jin2022alpha, chen2021glit, gong2021nasvit, chen2021autoformer, guo2019nat, ding2021hr, su2022vitas}. In practice, automated transformer search is technically the same as automated convolutional network search, in which both feature the same search pipeline. For example, HAT \cite{wang2020hat}, as one of the state-of-the-art NAS works in the field of NLP, focuses on searching for hardware-efficient transformers for NLP tasks. To achieve this, HAT first initializes an over-parameterized \textit{superformer} that consists of all the possible transformer candidates in the search space, which is technically the same as the supernet in automated convolutional network search. After that, HAT trains the \textit{superformer} using the standard weight-sharing technique \cite{pham2018efficient}, which then serves as the accuracy predictor to quickly interpret the accuracy of different transformer candidates. In the meantime, HAT builds an efficient latency predictor to avoid the tedious on-device latency measurement. Finally, HAT applies the standard evolutionary algorithm to find hardware-efficient transformer candidates with both high accuracy and high efficiency, which is technically the same as OFA \cite{cai2020once} that searches for hardware-efficient convolutional networks. Furthermore, due to the tremendous success of vision transformers in vision tasks as discussed in Section~\ref{sec:manual-transformers}, a plethora of NAS works \cite{jin2022alpha, chen2021glit, gong2021nasvit, chen2021autoformer, guo2019nat, ding2021hr, su2022vitas} have been subsequently proposed to automate the design of superior vision transformers. Among them, \cite{chen2021autoformer}, as the first one, introduces an evolutionary algorithm-based NAS framework dubbed AutoFormer. Similar to HAT, AutoFormer first constructs an over-parameterized \textit{superformer} that consists of all the possible vision transformer candidates in the search space, which is then trained using the weight entanglement scheme. The difference between weight sharing and weight entanglement is visualized in Fig.~\ref{fig:weight-sharing-vs-weight-entanglement}, in which weight entanglement is technically similar to the superkernel in SP-NAS \cite{stamoulis2020single}. Finally, AutoFormer applies the standard evolutionary algorithm to explore the optimal vision transformer candidate. These clearly demonstrate that we can easily leverage recent state-of-the-art NAS techniques that focus on searching for competitive CNNs to automate the design of top-performing transformers for both NLP and vision tasks.
\setlength{\columnsep}{\columnsepsavesavesavesave}

\begin{table}[t]
\resizebox{\textwidth}{!}{%
\begin{tabular}{l|cc|cc|c|c|c}
\toprule[0.175em]
\multirow{2}{*}{Benchmark} & \multicolumn{2}{c|}{Search Space} & \multicolumn{2}{c|}{Queryable} & \multirow{2}{*}{Tasks} & \multirow{2}{*}{Datasets} & \multirow{2}{*}{Metrics} \\ \cline{2-5}
 & \multicolumn{1}{c|}{Size} & Type & \multicolumn{1}{c|}{Tabular} & Surrogate &  &  &  \\ \hline
NAS-Bench-101 \cite{ying2019nasbench101} & \multicolumn{1}{c|}{423k} & Cell-Based & \multicolumn{1}{c|}{\ding{51}} & \ding{55}  & \makecell[c]{Image \\ Classification} & CIFAR-10 & \makecell[c]{Training Accuracy, \\ Validation Accuracy, \\ Testing Accuracy, \\ Training Time, and \\ Number of Parameters} \\ \hline
NAS-Bench-201 \cite{dong2020nasbench201} & \multicolumn{1}{c|}{15.6k} & Cell-Based & \multicolumn{1}{c|}{\ding{51}} & \ding{55} & \makecell[c]{Image \\ Classification} & \makecell[c]{CIFAR-10, \\ CIFAR-100, and \\ ImageNet-16-120} & \makecell[c]{Training Accuracy, \\ Validation Accuracy, \\ Testing Accuracy, \\ Training Loss, \\ Validation Loss, \\ Testing Loss, \\ Training Time, \\ Number of FLOPs, and \\ Number of Parameters} \\ \hline
NATS-Bench \cite{dong2021nats-bench} & \multicolumn{1}{c|}{39.3k} & Cell-Based & \multicolumn{1}{c|}{\ding{51}} & \ding{55} & \makecell[c]{Image \\ Classification} & \makecell[c]{CIFAR-10, \\ CIFAR-100, and \\ ImageNet-16-120} & \makecell[c]{Training Accuracy, \\ Validation Accuracy, \\ Testing Accuracy, \\ Training Loss, \\ Validation Loss, \\ Testing Loss, \\ Training Time, \\ Number of FLOPs, and \\ Number of Parameters} \\ \hline
NAS-Bench-301 \cite{siems2020nasbench301} & \multicolumn{1}{c|}{$10^{18}$} & Cell-Based & \multicolumn{1}{c|}{\ding{55}} & \ding{51} & \makecell[c]{Image \\ Classification} & CIFAR-10 & Validation Accuracy \\ \hline
NAS-Bench-360 \cite{tu2022nasbench360} & \multicolumn{1}{c|}{N/A} & \makecell[c]{Cell- and \\ Block-Based} & \multicolumn{1}{c|}{\ding{51}} & \ding{55} & 10 Diverse Tasks & 10 Diverse Datasets & N/A \\ \hline
NAS-Bench-1Shot1 \cite{zela2020nasbench1shot1} & \multicolumn{1}{c|}{399k} & Cell-Based & \multicolumn{1}{c|}{\ding{51}} & \ding{55} & \makecell[c]{Image \\ Classification} & CIFAR-10 & Validation Accuracy \\ \hline
NAS-Bench-ASR \cite{mehrotra2021nasbenchasr} & \multicolumn{1}{c|}{8.2k} & Cell-Based & \multicolumn{1}{c|}{\ding{51}} & \ding{55} & \makecell[c]{Automatic \\ Speech \\ Recognition} & TIMIT & \makecell[c]{CTC Loss, \\ Phoneme Error Rate (PER), \\ On-Device Latency, \\ Number of FLOPs, and \\ Number of Parameters}  \\ \hline
NAS-Bench-Graph \cite{qin2022nasbenchgraph} & \multicolumn{1}{c|}{26.2k} & Cell-Based & \multicolumn{1}{c|}{\ding{51}} & \ding{55} & 9 Graph Tasks  & 9 Graph Datasets & \makecell[c]{Training Loss, \\ Validation Loss, \\ Testing Loss, \\ Validation Accuracy, \\ On-Device Latency, and \\ Number of Parameters} \\ \hline
NAS-Bench-NLP \cite{klyuchnikov2022nasbenchnlp} & \multicolumn{1}{c|}{14k} & Cell-Based & \multicolumn{1}{c|}{\ding{51}} & \ding{55} & \makecell[c]{Language \\ Understanding} & \makecell[c]{PTB and \\  WikiText-2} & \makecell[c]{Testing Perplexity, \\ Training Time, and \\ Number of Parameters} \\ \hline
NAS-Bench-111 \cite{yan2021nasbenchx11} & \multicolumn{1}{c|}{423k} & Cell-Based & \multicolumn{1}{c|}{\ding{55}} & \ding{51} & \makecell[c]{Image \\ Classification} & CIFAR-10 & \makecell[c]{Training Accuracy, \\ Validation Accuracy, \\ Testing Accuracy, \\ Training Loss, \\ Validation Loss, \\ and Testing Loss} \\ \hline
NAS-Bench-311 \cite{yan2021nasbenchx11} & \multicolumn{1}{c|}{$10^{18}$} & Cell-Based & \multicolumn{1}{c|}{\ding{55}} & \ding{51} & \makecell[c]{Image \\ Classification} & CIFAR-10 & Same as NAS-Bench-111 \\ \hline
NAS-Bench-NLP11 \cite{yan2021nasbenchx11} & \multicolumn{1}{c|}{$10^{53}$} & Cell-Based & \multicolumn{1}{c|}{\ding{55}} & \ding{51} & \makecell[c]{Language \\ Understanding} & PTB & Same as NAS-Bench-111 \\ \hline
NAS-Bench-Suite \cite{mehta2022nasbenchsuite} & \multicolumn{1}{c|}{N/A} & Cell-Based & \multicolumn{1}{c|}{\ding{51}} & \ding{51} & \multicolumn{3}{c}{A suite of 11 tabular and surrogate NAS benchmarks} \\ \hline
HW-NAS-Bench \cite{li2021hwnasbench} & \multicolumn{1}{c|}{15.6k} & Cell-Based & \multicolumn{1}{c|}{\ding{51}} & \ding{55} & \makecell[c]{Image \\ Classification} & \makecell[c]{CIFAR-10, \\ CIFAR-100, and \\ ImageNet-16-120} & On-Device Latency  \\ \hline
HW-NAS-Bench \cite{li2021hwnasbench} & \multicolumn{1}{c|}{$10^{21}$} & Block-Based & \multicolumn{1}{c|}{\ding{55}} & \ding{51} & \makecell[c]{Image \\ Classification} & \makecell[c]{CIFAR-100 and \\ ImageNet} & On-Device Latency \\ 
\toprule[0.175em]
\end{tabular}%
}
\caption{Comparisons of different NAS benchmarks. Note that ImageNet-16-120 is a subset of ImageNet that consists of 120 object categories, in which the input image resolution is fixed to 16$\times$16 \cite{dong2020nasbench201}.}
\vspace{-14pt}
\label{tab:nas-benchmarks}
\end{table}

\textbf{Efficient Domain-Specific Search.}
In addition to image classification, NAS can also be applied to a wide range of real-world scenarios, such as object detection \cite{xiong2021mobiledets, wang2020fcos, ghiasi2019fpn}, semantic segmentation \cite{liu2019auto, shaw2019squeezenas, zhang2021dcnas}, point cloud processing \cite{liu2022lidarnas, liu2021pvnas, tang2020searching, li2020lcnas}, image super-resolution \cite{liu2021evsrnet, wu2022compiler}, etc. For example, MobileDets \cite{xiong2021mobiledets} are a family of hardware-efficient object detection networks, which can deliver promising detection accuracy while maintaining superior detection efficiency on multiple embedded computing systems, including mobile CPUs, edge TPUs, and edge GPUs. Specifically, MobileDets first construct an enlarged search space that contains a large number of possible object detection networks and then leverage MnasNet-like reinforcement learning-based search algorithm \cite{tan2019mnasnet} to find top-performing object detection networks, which also feature the same reward function as TuNAS \cite{bender2020tunas} to trade off between the detection accuracy and efficiency. Besides, \cite{li2020lcnas} introduces an efficient hardware-aware differentiable NAS framework dubbed LC-NAS, aiming to automate the design of competitive network solutions for point cloud processing. Here, similar to EdgeNAS \cite{luo2020edgenas} and LA-DARTS \cite{xu2020latency} that focus on finding top-performing architecture candidates for image classification, LC-NAS exploits the same cell-based search space and integrates the latency constraint into the optimization objective to penalize the architecture candidate with high latency. These demonstrate that we can easily include domain-specific knowledge (e.g., domain-specific search spaces) into mainstream NAS techniques (e.g., differentiable, evolutionary algorithm-based, and reinforcement learning-based NAS) to search for domain-specific network solutions.

\textbf{Mainstream NAS Benchmarks.}
Although NAS has achieved substantial performance improvement across various NLP and vision tasks, fair comparisons between different NAS works are frustratingly hard and still an open issue as demonstrated in \cite{yang2020evaluation}. This is because different NAS works may feature quite different training recipes, such as different training epochs and training enhancements. For example, DARTS+ \cite{liang2019darts+} trains the searched architecture candidate on CIFAR-10 for 2,000 epochs, whereas DARTS \cite{liu2019darts} only applies 600 training epochs. Meanwhile, DARTS+ trains the searched architecture candidate on ImageNet for 800 epochs with the batch size of 2,048, where AutoAugment \cite{cubuk2019autoaugment} is also integrated in order to achieve stronger data augmentations. In contrast, DARTS only applies 250 training epochs with the batch size of 128 by default. We note that, for the same architecture candidate, longer training epochs and stronger data augmentations typically achieve better training accuracy on target task as shown in \cite{yang2020evaluation}. Furthermore, RandomNAS \cite{li2020random} challenges the effectiveness of early state-of-the-art NAS works and demonstrates that random search, as one strong search baseline to explore random networks, can achieve even better performance on target task than early state-of-the-art NAS works. In parallel, RandWire \cite{xie2019exploring} shows that randomly wired networks can also exhibit strong accuracy on ImageNet. Therefore, it remains unknown whether the performance improvement of NAS is due to the more advanced training recipe or the search algorithm itself, making it difficult to evaluate and compare the technical contributions of different NAS works \cite{yang2020evaluation, ying2019nasbench101, dong2020nasbench201}.

To overcome such limitations, a plethora of tabular and surrogate NAS benchmarks have been subsequently proposed, including NAS-Bench-101 \cite{ying2019nasbench101}, NAS-Bench-201 \cite{dong2020nasbench201}, NATS-Bench \cite{dong2021nats-bench}, NAS-Bench-301 \cite{siems2020nasbench301}, NAS-Bench-360 \cite{tu2022nasbench360}, NAS-Bench-1Shot1 \cite{zela2020nasbench1shot1}, NAS-Bench-ASR \cite{mehrotra2021nasbenchasr}, NAS-Bench-Graph \cite{qin2022nasbenchgraph}, NAS-Bench-NLP \cite{klyuchnikov2022nasbenchnlp}, HW-NAS-Bench \cite{li2021hwnasbench}, NAS-Bench-x11 \cite{yan2021nasbenchx11}, and NAS-Bench-Suite \cite{mehta2022nasbenchsuite}. We note that NAS benchmarks typically have two important parts, including the pre-defined search space and the related performance metrics for all the possible architecture candidates that can be easily queried. Specifically, in tabular NAS benchmarks \cite{ying2019nasbench101, dong2020nasbench201, dong2021nats-bench, tu2022nasbench360, zela2020nasbench1shot1, mehrotra2021nasbenchasr, qin2022nasbenchgraph, klyuchnikov2022nasbenchnlp, li2021hwnasbench}, all the possible architecture candidates are enumerated and trained from scratch on target task, respectively, to obtain the performance metrics, such as the training and validation accuracy. In contrast, surrogate NAS benchmarks \cite{siems2020nasbench301, mehta2022nasbenchsuite, yan2021nasbenchx11, li2021hwnasbench} leverage learning-based methods to predict the performance metrics of different architecture candidates rather than directly enumerating and training all the possible architecture candidate candidates on target task, thus leading to significantly reduced computational resources. In sight of this, surrogate NAS benchmarks can be easily extended to deal with larger search spaces than tabular NAS benchmarks ($10^{18}$ in NAS-Bench-301 \cite{siems2020nasbench301} vs. 15,625 in NAS-Bench-201 \cite{dong2020nasbench201}). Finally, we compare and summarize the aforementioned state-of-the-art NAS benchmarks in Table~\ref{tab:nas-benchmarks}.

\subsection{Future Envision}
In this section, we further envision several promising future trends and possible directions in the field of automated network design, which are summarized as follows:
\begin{itemize}
    \item[(1)] {
    \textbf{General Search Spaces.}
    The success of NAS highly relies on the well-engineered search space, such as the cell-based search space \cite{zoph2016neural, zoph2018nasnet, pham2018efficient} and the block-based search space \cite{tan2019mnasnet, wu2019fbnet, cai2019proxylessnas}. In the past, researchers manually design the search spaces using heuristic-based strategies, which are typically based on existing state-of-the-art networks, such as MobileNets \cite{xiong2021mobiledets, mehta2022mobilevitv2} and ShuffleNets \cite{zhang2018shufflenet, ma2018shufflenet}. This effectively restricts the search space to improve the search efficiency and delivers competitive architecture candidates with promising accuracy and efficiency. In the meantime, this, however, may significantly limit the search performance, which may reject more competitive architecture candidates outside the well-engineered search space. To overcome such limitations, \cite{radosavovic2020designing, guo2020hit} pioneer to design more general search spaces than the cell-based and block-based search spaces, which, unfortunately, are under-explored since \cite{radosavovic2020designing, guo2020hit} still suffer from human biases. Therefore, one promising future direction in the field of NAS is to innovate and explore more general search spaces to unleash the power of automated network design. 
    }
    \item[(2)] {
    \textbf{Fully Automated Architecture Search.}
    The early NAS practices either focus on searching for the optimal architecture candidate \cite{zoph2016neural, zoph2018nasnet, pham2018efficient}, the optimal data augmentation \cite{cubuk2019autoaugment, hataya2020fasterautoaugment}, the optimal activation function \cite{ramachandran2017searching, zhou2021learning}, or the optimal training recipe \cite{dai2021fbnetv3, dong2020autohas}. As demonstrated in FBNetV3 \cite{dai2021fbnetv3} and AutoHAS \cite{dong2020autohas}, different architecture candidates may prefer different training recipes, in which jointly searching for the optimal architecture candidate and its tailored training recipe has the potential to push forward the attainable accuracy. This observation can be easily generalized. For example, different architecture candidates may prefer different data augmentations. Therefore, one promising future direction in the field of NAS is fully automated search, which jointly searches for the optimal architecture candidate and its tailored data augmentation, activation function, and training recipe in one single search experiment to maximize the attainable accuracy.
    }
    \item[(3)] {
    \textbf{Multi-Task Architecture Search.}
    Previous NAS works typically focus on searching for task-specific architecture candidates that can achieve promising performance in the specified task, such as image classification \cite{liu2019darts, xu2020pcdarts}, object detection \cite{xiong2021mobiledets, wang2020fcos, ghiasi2019fpn}, and semantic segmentation \cite{liu2019auto, shaw2019squeezenas, zhang2021dcnas}. This search paradigm, however, significantly increases the total search cost when the number of tasks exponentially evolves since we have to conduct search experiments for each task, respectively. To alleviate this issue, FBNetV5 \cite{wu2021fbnetv5} takes the first step to search for multi-task architecture candidates which can achieve competitive performance across multiple tasks, including image classification on ImageNet \cite{deng2009imagenet}, object detection on COCO \cite{lin2014coco}, and semantic segmentation on ADE20K \cite{zhou2017ade20k}. Nonetheless, this is far from enough since we have a large number of tasks in real-world scenarios. Therefore, one promising future direction in the field of NAS is multi-task search, which automates the design of top-performing architecture candidates that can be generalized to multiple different tasks without being re-engineered (i.e., \textit{once for all}).    
    }
    \item[(4)] {
    \textbf{Dynamic Architecture Search.}
    Previous NAS works \cite{zoph2016neural, zoph2018nasnet, pham2018efficient, liu2019darts} typically focus on searching for static neural networks that can only run at fixed computational budgets, which cannot adapt to lower or higher computational complexity. In addition, dynamic neural networks, such as slimmable neural networks \cite{yu2018slimmable, yu2019universally, li2021dynamic}, are another important branch of DNNs, which can be executed to accommodate different computational resources in real-world environments. This is because, even on the same hardware device, the available computational resources may vary with respect to time. For example, mobile phones may be in low-power or power-saving modes to reduce the power consumption. To overcome such limitations, deploying multiple static neural networks on the same hardware device seems to be the first-in-mind solution, which, unfortunately, demands high on-device storage requirements. Therefore, one promising future direction in the field of NAS is to search for top-performing dynamic neural networks, which can instantly, adaptively, and efficiently trade off between accuracy and inference efficiency to accommodate the rapidly-changing computational budgets in real-world embedded computing scenarios.
    }
    \item[(5)] {
    \textbf{Hybrid Architecture Search.}
    As discussed in Section~\ref{sec:manual-network-design-for-embedded-computing-systems}, both convolutional networks and vision transformers have their own technical merits when applied to vision tasks. Specifically, convolutional networks demonstrate superior efficiency on target hardware, whereas vision transformers achieve better accuracy on target task. In sight of this, designing hybrid networks on top of both convolutional networks and vision transformers has the potential to push forward accuracy-efficiency trade-offs. Nonetheless, previous NAS works typically focus on searching for either convolutional networks \cite{zoph2016neural, zoph2018nasnet, pham2018efficient, liu2019darts} or vision transformers \cite{chen2021glit, gong2021nasvit, chen2021autoformer, su2022vitas}. To alleviate this, \cite{li2021bossnas, mecharbat2023hyt} have taken the very first steps to investigate hybrid architecture search, which, however, still remains under-explored. Therefore, one promising future direction in the field of NAS is to search for competitive hybrid networks that combine the technical merits of both convolutional networks and vision transformers to achieve better accuracy-efficiency trade-offs. 
    }
    \item[(6)] {
    \textbf{Explainable Architecture Search.}
    Previous representative NAS works \cite{zoph2018nasnet, liu2019darts, cai2019proxylessnas, cai2020once} highly rely on the weight-sharing paradigm \cite{pham2018efficient}, also known as one-shot NAS, which initializes an over-parameterized supernet that consists of all the possible architectures in the search space and then searches for the optimal architecture candidate within the supernet through weight-sharing. Despite the promising search efficiency, one-shot NAS has been widely criticized due to its limited explainability, which implies that weight-sharing may lead to sub-optimal architectures due to weight interference. And even worse, the intuition behind one-shot NAS still remains unknown in the NAS community. To alleviate this, a plethora of zero-shot NAS works have been recently proposed \cite{abdelfattah2021zero, krishnakumar2022bench, lopes2021epenas, turner2019fisher, wang2020grasp, mellor2021neural, lee2018snip, tanaka2020synflow, lin2021zenscore, akhauri2022eznas, chen2021neural, xiong2020linearregions, xiao2020ntk}, which leverage zero-cost proxies to quickly interpret the accuracy of different architectures. However, existing zero-cost proxies still cannot achieve reliable performance estimation as shown in Fig.~\ref{fig:lrc-ntk-correlation-cifar100-nasbench201}. Therefore, one promising future direction in the field of NAS is to develop more explainable NAS techniques and innovate more reliable zero-cost proxies.
    }
    \item[(7)] {
    \textbf{Meta Architecture Search.}
    Meta-learning \cite{finn2017model}, also referred to as \textit{learning-to-learn}, aims to facilitate and accelerate common learning-based practices such that the learned model can quickly adapt to unseen tasks/environments using minimal engineering efforts. For example, HELP \cite{lee2021hardware} introduces an efficient meta-learning based latency predictor, which can be generalized to new hardware platforms using as few as 10 latency measurements. In particular, the widely used weight-sharing paradigm in the field of NAS can be considered to be a special case of meta-learning, which takes the over-parameterized supernet as the meta-model. As the weight-sharing paradigm has been dominating recent advances in the field of NAS, one promising future direction is to explore meta-learning to accelerate the search process and enhance the few-shot learning capability \cite{shaw2019meta, wang2020m, lee2021rapid}. 
    }
\end{itemize}

\section{Network Compression for Embedded Computing Systems}
\label{sec:network-compression-for-embedded-computing-systems}

\begin{figure}[t]
    \begin{center}
    \includegraphics[width=0.95\columnwidth]{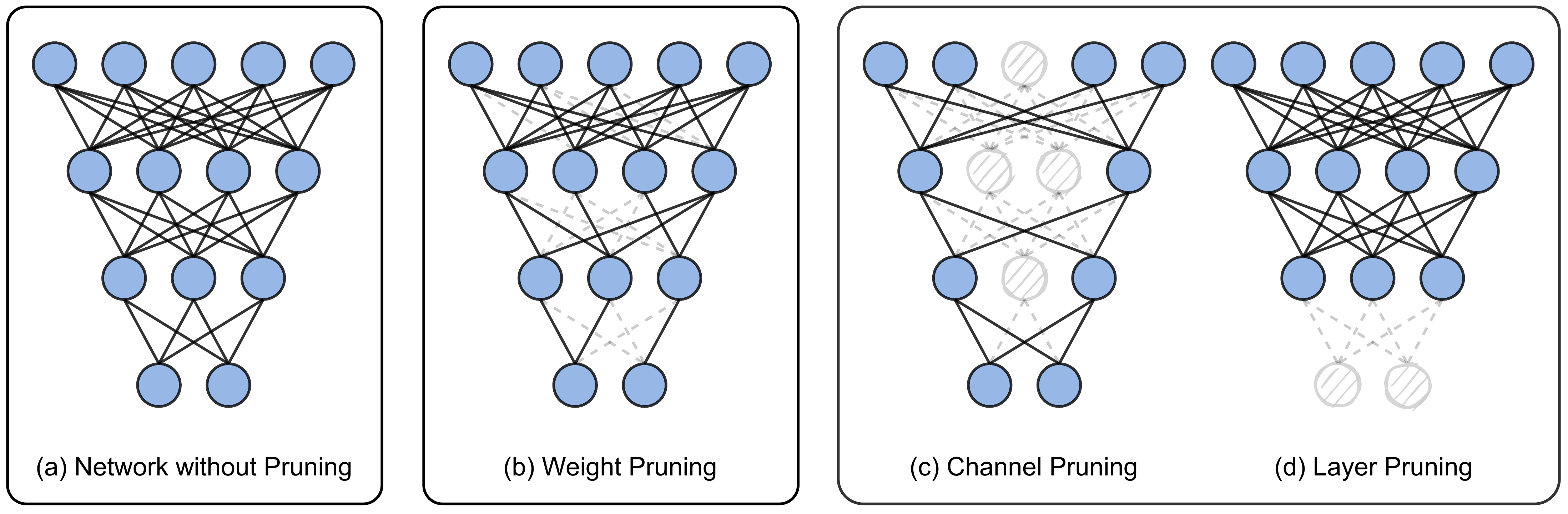}
    \end{center}
    \vspace{-5pt}
    \caption{Illustration of different structured and non-structured pruning strategies. Among them, weight pruning is non-structured, whereas channel pruning and layer pruning are structured.}
    \vspace{-5pt}
    \label{fig:structured-pruning-vs-non-structured-pruning}
\end{figure}

In addition to designing novel networks, another alternative is to compress existing networks at hand, either manually designed networks or automatically searched networks, to reduce the network redundancy, which therefore leads to network variants with better accuracy-efficiency trade-offs. As illustrated in previous relevant literature \cite{deng2020model, cai2022enable}, there are three popular branches of network compression techniques, including network pruning, network quantization, and network distillation. Note that these three branches are parallel with each other as shown in \cite{wang2020apq}, which indicates that they can be combined to further enable better accuracy-efficiency trade-offs. Among them, network pruning and network quantization focus on improving the accuracy-efficiency trade-off from the efficiency perspective, whereas network distillation enhances the accuracy-efficiency trade-off from the accuracy perspective. To this end, we, in this section, systematically discuss recent state-of-the-art network compression techniques. For better understanding, we divide these network compression techniques into three main categories and sub-sections, including network pruning in Section~\ref{sec:network-pruning}, network quantization in Section~\ref{sec:network-quantization}, and network distillation in Section~\ref{sec:network-distillation}, since these network compression techniques feature different algorithms to improve the accuracy-efficiency trade-off from different perspectives. Note that these network compression techniques can typically generalize across different networks (e.g., convolutional networks and transformers). For example, we can leverage knowledge distillation to enhance the training process of both convolutional networks and transformers towards better training accuracy.

\subsection{Network Pruning}
\label{sec:network-pruning}

The rationale behind network pruning is that DNNs are usually over-parameterized and redundant in terms of network weights and channels \cite{du2019gradient, liu2018rethinking}. As such, eliminating redundant network weights and channels can largely benefit the network efficiency at the cost of minimal accuracy loss, thus being able to accommodate limited available computational resources and rigorous storage requirements in real-world embedded scenarios. Following previous well-established pruning conventions, we divide recent state-of-the-art pruning methods into two main categories according to their pruning granularity, in which non-structured pruning (i.e., weight pruning) is fine-grained whereas structured pruning (i.e., channel pruning and layer pruning) is coarse-grained. As illustrated in Fig.~\ref{fig:structured-pruning-vs-non-structured-pruning}, weight pruning focuses on removing the redundant weight connections, whereas channel pruning and layer pruning focus on removing the redundant channels and layers. In practice, both non-structured pruning and structured pruning can explore simplified network structures with optimized computational efficiency. Nonetheless, non-structured weight pruning highly relies on specialized hardware accelerators \cite{han2015learning} and cannot provide realistic runtime speedups on modern embedded computing systems due to irregular network sparsity \cite{han2016eie, liu2018rethinking}. In contrast, structured channel pruning and layer pruning are coarse-grained and do not introduce irregular network sparsity, which can deliver realistic runtime speedups on modern embedded computing systems. For better coverage, below we further discuss recent representative works in the field of both non-structured pruning and structured pruning, which are also summarized in Fig.~\ref{fig:comparisons-pruning}.

\subsubsection{Non-Structured Pruning.}
\label{sec:non-structured-pruning}

Non-structured pruning, also referred to as weight pruning, removes the less important network weights, which is typically more fine-grained than structured pruning as illustrated in Fig.~\ref{fig:structured-pruning-vs-non-structured-pruning}. In particular, applying weight pruning for network compression can be traced back to the early 1990s. For example, Optimal Brain Damage \cite{lecun1989optimal} and Optimal Brain Surgeon \cite{hassibi1993optimal}, as the very early weight pruning approaches, pioneer to investigate the efficacy of weight pruning on vanilla fully-connected networks, in which the less important network weights are removed based on Hessian of the loss function. More recently, \cite{han2015learning} proposes a simple yet effective weight pruning technique to compress deep convolutional networks, such as AlexNet \cite{krizhevsky2012alexnet} and VGGNet \cite{simonyan2014vggnet}, instead of vanilla fully-connected networks. Specifically, \cite{han2015learning} observes that the network weights with smaller magnitudes typically contribute less to the network accuracy, based on which \cite{han2015learning} removes the less important network weights with smaller magnitudes. Subsequently, this weight pruning technique is further integrated into Deep Compression \cite{han2015deep} to obtain highly compressed networks, making it possible to aggressively reduce the network size without sacrificing the network accuracy. For example, Deep Compression is able to significantly reduce the network size of VGGNet by $\times$49 times, from 552\,MB to 11.3\,MB, while maintaining comparable accuracy on ImageNet. Nonetheless, the reduction in terms of the network size cannot directly translate into the speedup on target hardware since the resulting compressed networks are of high irregular network sparsity. To overcome such limitations, EIE \cite{han2016eie} designs an efficient specialized inference engine to maximize the inference efficiency of compressed networks. In parallel, \cite{srinivas2015data} proposes an efficient data-free weight pruning approach to iteratively remove redundant network weights. Besides, \cite{molchanov2017variational} and \cite{louizos2017learning} leverage Variational Dropout and $L_0$-norm regularization-based stochastic gates, respectively, to remove the less important network weights.

\newlength{\columnsepsavesavesavesavesave} 
\setlength{\columnsepsavesavesavesavesave}{\columnsep} 
\setlength{\columnsep}{6pt} 
\begin{wrapfigure}{r}{0.4\columnwidth}
    \begin{center}
        \includegraphics[width=0.375\columnwidth]{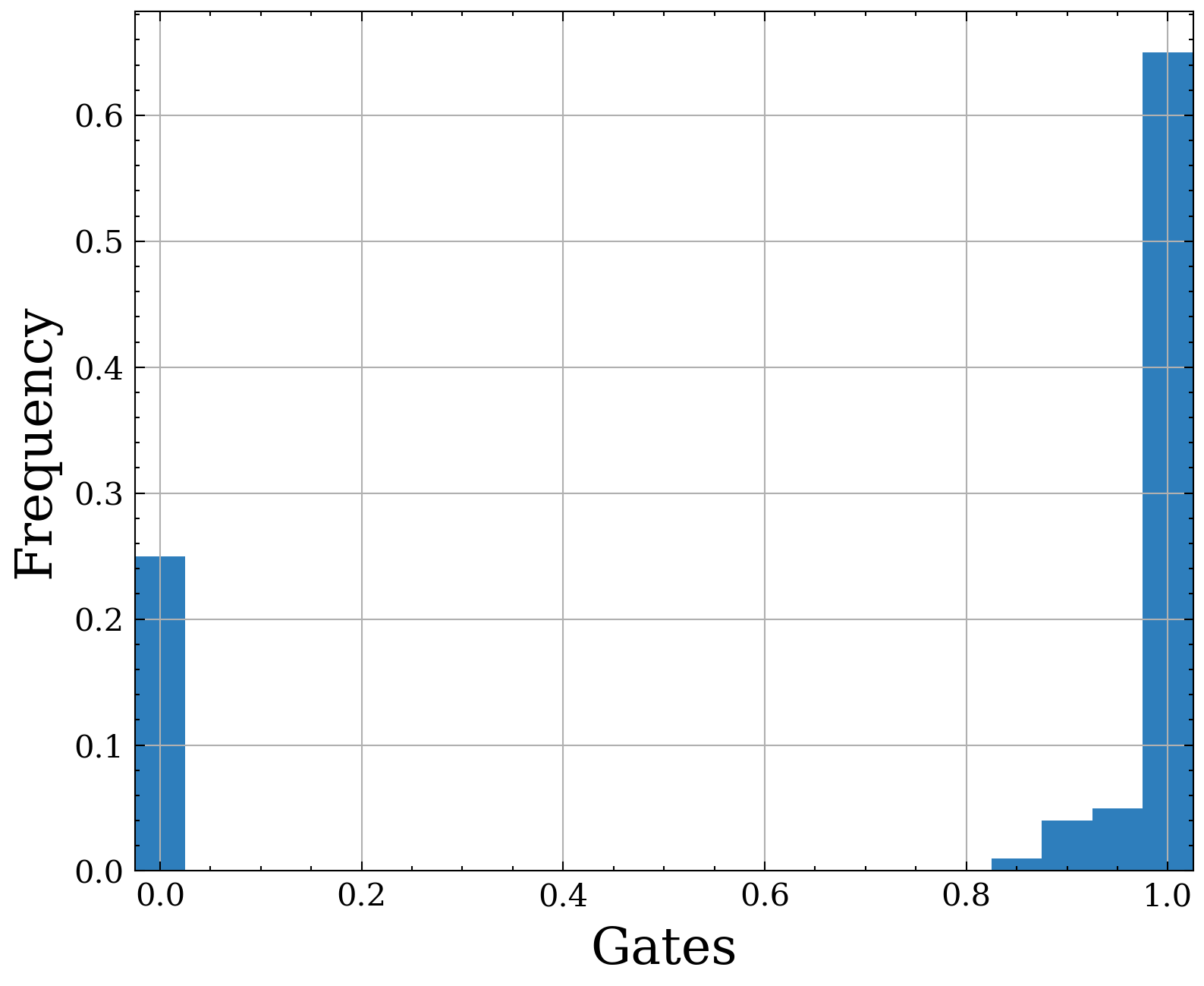}
    \end{center}
    \caption{Distribution of weight gates \cite{guo2021gdp}.}
    \label{fig:gdp-pruning-gates}
\end{wrapfigure}
\textbf{Weight Importance Criteria.}
The core of weight pruning is to determine the importance of different network weights, based on which we can easily rank different network weights and remove the less important network weights at the cost of minimal accuracy loss. There have been several representative importance criteria to measure the importance of different network weights after the network is trained. Among them, the most straight-forward criterion is based on the weight magnitude thanks to its conceptual simplicity and surprisingly strong performance, which leverages the absolute weight $|w|$ to interpret the importance of different network weights (i.e., the larger, the more important) \cite{gale2019state, han2015learning, renda2020comparing}. The rationale behind magnitude-based weight pruning is that smaller network weights typically contribute less to the output of the network. Apart from this, other importance criteria include second-order derivative-based \cite{lecun1989optimal, hassibi1992second}, Taylor expansion-based \cite{molchanov2016pruning}, output sensitivity-based \cite{engelbrecht2001new} strategies, etc. More recently, GDP \cite{guo2021gdp} proposes an effective strategy, namely gates with differentiable polarization, which introduces learnable gates to interpret the importance of different network weights. In particular, GDP encourages a large margin between exact zero gates and non-zero gates as shown in Fig.~\ref{fig:gdp-pruning-gates}, while still allowing gradient optimization. Finally, GDP removes the network weights with exact zero gates and further merges the remaining non-zero gates into the resulting pruned network without hurting the network accuracy once the optimization process terminates. Despite the impressive progress to date, the design of efficient and effective importance criterion is quite under-explored and still remains an open challenge in the community.
\setlength{\columnsep}{\columnsepsavesavesavesavesave}

\textbf{Sparse Network Acceleration.}
Different from structured pruning that is hardware-friendly, non-structured pruning, despite being able to maintain competitive accuracy under high compression ratios, introduces considerable irregular network sparsity, making it difficult to parallelize the resulting sparse networks on mainstream hardware systems like GPUs and CPUs \cite{li2023efficient}. This indicates that non-structured pruning highly relies on specialized hardware to achieve superior on-device speedups. To this end, a plethora of specialized hardware accelerators \cite{chen2016eyeriss, chen2019eyeriss, han2016eie, han2017ese, zhang2016cambricon, parashar2017scnn, deng2021gospa, zhang2020snap, gudaparthi2022candles} and compiler-based optimization techniques \cite{ma2020pconv, niu2020patdnn} have been recently developed to accelerate the on-device inference of sparse networks, which typically focus on improving the irregular memory access on target hardware. For example, Cambricon-X \cite{zhang2016cambricon} features an access-efficient indexing module to select and transfer irregular network weights to different processing elements (PEs) with reduced bandwidth requirements. This indexing module allows each PE to store irregular network weights for local computation in an asynchronous manner, and as a result, significantly reduces the irregular memory access overheads across different PEs.

\textbf{Sparse Training Techniques.}
As shown in \cite{han2015learning}, weight pruning can effectively lead to efficient sparse network variants that are as smaller as 90\% than the unpruned network, significantly alleviating the storage requirements and reducing the computational complexity. Despite the promising efficiency improvement, training the resulting sparse networks is quite challenging, in which using conventional training strategies may lead to non-negligible accuracy loss as demonstrated in \cite{frankle2018lottery}. To recover the accuracy, a plethora of sparse training techniques have been developed to train the resulting sparse networks \cite{han2015learning, han2015deep, srinivas2017training, evci2019difficulty, jaiswal2022training, sung2021training}. For example, \cite{han2015learning} proposes to fine-tune the pruned sparse network with inherited network weights from the unpruned network. Furthermore, \cite{han2015deep} generalizes the above fine-tuning strategy to become iterative, where multiple iterations of pruning and fine-tuning are iteratively repeated to recover the attainable accuracy. In addition, \cite{evci2019difficulty} investigates the performance collapse issue of training sparse networks, which may simply be trapped into sub-optimal local minima. To escape the sub-optimal local minima, \cite{evci2019difficulty} proposes to traverse extra dimensions between dense and sparse sub-spaces during training sparse networks. More recently, \cite{jaiswal2022training} introduces an alternative sparse training strategy, which customizes the sparse training techniques to deviate from the default vanilla training protocols, consisting of introducing \textit{ghost} neurons and skip connections at the early training stage, and strategically modifying the initialization as well as the labels. Below we further introduce another representative branch of weight pruning and sparse training techniques, namely lottery ticket hypothesis \cite{frankle2018lottery}, which demonstrates that the pruned sparse networks, when properly initialized, can be trained from scratch to recover the accuracy comparable to the unpruned network.

\textbf{Lottery Ticket Hypothesis.}
The lottery ticket hypothesis \cite{frankle2018lottery} is a special case of non-structured pruning and has since gained increasing popularity in the pruning community \cite{malach2020proving, zhang2021validating, frankle2019stabilizing, chen2021unified, kim2022exploring, banerjee2022pruning}. Specifically, the lottery ticket hypothesis reveals that: \textit{A randomly-initialized unpruned network contains sparse sub-networks (i.e., winning tickets) that are initialized such that, when trained in isolation, they can match the accuracy of the unpruned network after training for at most the same number of iterations.} In particular, the winning tickets can be as smaller as 90\% than the unpruned network, while at the same time maintaining comparable accuracy. To identify the winning ticket, the lottery ticket hypothesis \cite{frankle2018lottery} proposes to leverage the following steps:
\begin{itemize}
    \item[(1)] {
    Randomly initialize the unpruned network $f(x;\theta_0)$, in which $\theta_0 \sim \mathcal{D}_{\theta}$;
    }
    \item[(2)] {
    Train the unpruned network for a number of $j$ iterations, arriving at weights $\theta_j$;
    }
    \item[(3)] {
    Prune $p$\% of the weights in $\theta_j$, creating the sparse mask $m \in \{0, 1\}^{|\theta_0|}$;
    }
    \item[(4)] {
    Reset the remaining weights to their values in $\theta_0$, creating the winning ticket $f(x;m \odot \theta_0)$.
    }
\end{itemize}
As demonstrated in \cite{frankle2018lottery}, directly pruning $p$\% of the network weights may lead to significant accuracy loss, which also makes the training process unstable. To overcome such limitations, the lottery ticket hypothesis proposes to iteratively repeat the above steps, also referred to as iterative pruning, which repeatedly trains, prunes, and resets the network weights over $n$ rounds where each round prunes $p^{\frac{1}{n}}$\% of the network weights, respectively. The lottery ticket hypothesis opens up the possibility and provides empirical guidelines to train sparse networks from scratch to match the accuracy of the unpruned network. Subsequently, \cite{malach2020proving, zhang2021validating} prove the lottery ticket hypothesis from insightful theoretical perspectives. In parallel, \cite{frankle2019stabilizing} investigates the performance collapse issue of the lottery ticket hypothesis, especially when dealing with deeper networks, such as ResNets \cite{he2016deep} and DenseNets \cite{huang2017densely}, based on which \cite{frankle2019stabilizing} proposes an effective modified iterative pruning scheme called rewinding iteration to stabilize the lottery ticket hypothesis. Furthermore, \cite{chen2021unified, kim2022exploring, banerjee2022pruning} generalize the lottery ticket hypothesis to other types of networks beyond convolutional networks, such as graph networks \cite{chen2021unified}, spiking networks \cite{li2022exploring}, and photonic networks \cite{banerjee2022pruning}.

\textbf{Semi-Structured Pruning.}
In contrast to the above mainstream non-structured pruning methods that introduce considerable irregular network sparsity, semi-structured pruning focuses on removing the less important \textit{consecutive} weight connections \cite{zhang2022learning}. The resulting semi-structured sparse networks can exhibit less irregular network sparsity than non-structured pruning, which are well supported by some existing deep learning libraries (e.g., cuSPARSElt \cite{mishra2021accelerating} and TVM \cite{chen2018tvm}) and thus can maintain much higher parallelism and speedups on modern embedded computing systems than non-structured pruning. For example, popular BERT models can achieve about 1.3x to 1.6 runtime inference speedups on Nvidia A100 GPUs using the optimized sparse tensor cores \cite{mishra2021accelerating, holmes2021nxmtransformer}. Thanks to its superior accuracy-efficiency trade-offs over non-structured pruning, semi-structured pruning has been widely employed to optimize the computational complexity of convolutional networks \cite{zhou2021learningsemi, pool2021channel}, transformers \cite{holmes2021nxmtransformer, bambhaniya2024progressive}, and large language models \cite{li2023sparse, jaiswal2024emergence}. For example, \cite{pool2021channel} introduces an effective channel permutation scheme to optimize the attainable accuracy of the resulting semi-structured sparse convolutional network. \cite{zhou2021learningsemi} introduces sparse-refined straight-through estimator (SR-STE) to explore the optimal semi-structured sparse convolutional network. In addition to convolutional networks, \cite{holmes2021nxmtransformer} and \cite{bambhaniya2024progressive} introduce alternating direction method of multipliers (ADMM) and progressive gradient flow to explore the optimal semi-structured sparse transformer for real-world language processing tasks. More recently, \cite{li2023sparse, jaiswal2024emergence} investigate semi-structured sparsity to enhance the inference efficiency of large language models. In parallel to the above semi-structured pruning methods that optimize the inference efficiency, \cite{zhang2023bi, lasby2023dynamic} instead focus on optimizing the training efficiency of semi-structured sparse networks. Furthermore, \cite{fang2022algorithm, fang2022efficient, luo2022codg} also focus on exploring dedicated hardware accelerators to further enhance the runtime inference efficiency of semi-structured sparse networks.

\tikzset{
basic/.style = {draw, font=\footnotesize, rectangle},
root/.style = {basic, font=\footnotesize,  text width=3cm, rounded corners=2pt, thin, align=left, fill=color1!18},
parent/.style = {basic, text width=3.25cm, rounded corners=1.5pt, thin, align=left, fill=color1!7},
child/.style = {basic, text width=5.75cm, rounded corners=1.5pt, thin, align=left, fill=color1!1.5},
}
\forestset{
  main'/.style={
    l sep=5mm,
    anchor=west,
  },
  root'/.style={root,
    anchor=west,
    edge path={
     \noexpand\path[\forestoption{edge}]
     ($(!u.east)!.0!(!.west)$) ++(0.5,0) |- (!.west);
    },
  },
  parent'/.style={parent,
    anchor=west,
    calign=child edge,
    l sep=0.4cm
  },
}
\begin{figure*}[t]
    \hspace{-0.85cm} 
    \centering
\begin{forest}
    for tree={
        forked edges,
        text centered,
        grow'=east,
        reversed=true,
        font=\footnotesize,
        rectangle, /tikz/align=left, anchor=base west, tier/.pgfmath=level(),
        rounded corners,
    }
    [, main'
    [Non-Structured Pruning, root'
        [Weight Importance, parent'
                [
                    \cite{gale2019state, han2015learning, renda2020comparing, lecun1989optimal, hassibi1992second, molchanov2016pruning, engelbrecht2001new, guo2021gdp}
                    , child
                ]
        ]
        [Sparse Acceleration, parent'
                [
                    \cite{chen2016eyeriss, chen2019eyeriss, han2016eie, han2017ese, zhang2016cambricon, parashar2017scnn, deng2021gospa, zhang2020snap, gudaparthi2022candles, ma2020pconv, niu2020patdnn}
                    , child
                ]
        ]
        [Sparse Training, parent'
                [
                    \cite{han2015learning, han2015deep, srinivas2017training, evci2019difficulty, jaiswal2022training, sung2021training}
                    , child
                ]
        ]
        [Lottery Ticket Hypothesis, parent'
                [
                    \cite{frankle2018lottery, malach2020proving, zhang2021validating, frankle2019stabilizing, chen2021unified, kim2022exploring, banerjee2022pruning, chen2021unified, kim2022exploring, banerjee2022pruning}
                    , child
                ]
        ]
        [Semi-Structured Pruning, parent'
                [
                    \cite{zhang2022learning, zhou2021learningsemi, pool2021channel, holmes2021nxmtransformer, bambhaniya2024progressive, li2023sparse, jaiswal2024emergence, zhang2023bi, lasby2023dynamic, fang2022algorithm, fang2022efficient, luo2022codg}
                    , child
                ]
        ]
    ]
    [Structured Pruning, root'
        [Weight-Based, parent'
                [
                    \cite{li2016pruning, he2018soft, he2019filter, yvinec2021red, yvinec2022red++, wang2019cop, wang2021convolutional}
                    , child
                ]
        ]
        [Activation-Based, parent'
                [
                    \cite{he2017channel, lin2020hrank, sui2021chip, tan2020dropnet, luo2017thinet, ding2019approximated, lin2017runtime, li2021dynamic, lee2018snip}
                    , child
                ]
        ]
        [Statistics-Based, parent'
                [
                    \cite{liu2017learning, you2019gate, zhuang2020neuron, ye2018rethinking, kang2020operation}
                    , child
                ]
        ]
        [Search-Based, parent'
                [
                    \cite{he2018amc, yu2021auto, alwani2022decore, liu2019metapruning, lin2021channel, li2022sub, li2020dhp, guo2020dmcp, ning2020dsa}
                    , child
                ]
        ]
        [Layer-Based, parent'
                [
                    \cite{jordao2020discriminative, chen2018shallowing, elkerdawy2020filter, tang2023sr, zhang2022layer, jordao2023layers}
                    , child
                ]
        ]
    ]
    ]
\end{forest}
\caption{Illustration of non-structured and structured pruning works that have been discussed in Section~\ref{sec:network-pruning}.}
\vspace{-5pt}
\label{fig:comparisons-pruning}
\end{figure*}
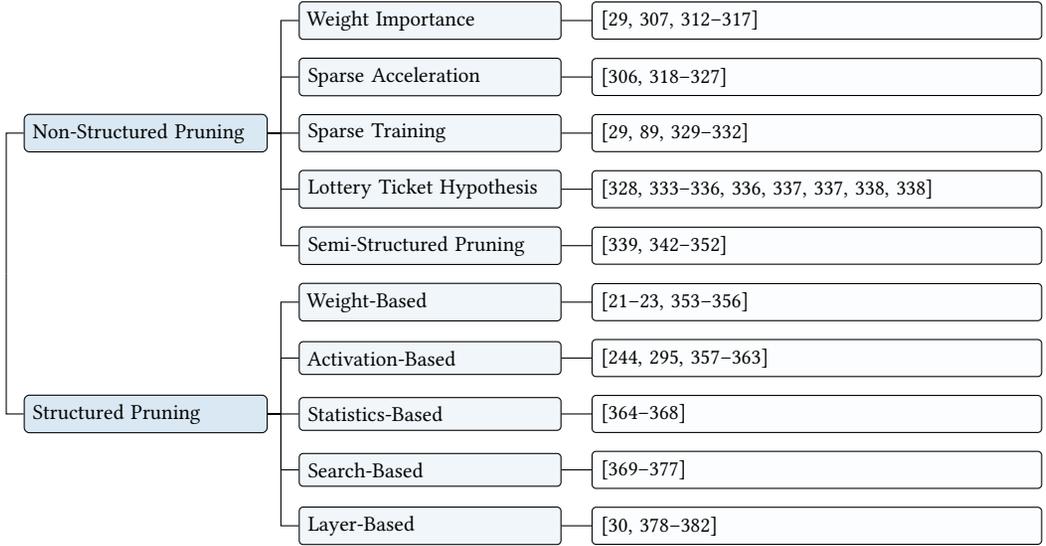

\subsubsection{Structured Pruning.}
\label{sec:structured-pruning}

In parallel to non-structured pruning, structured pruning, including channel pruning\footnote{Filter pruning is another name for channel pruning since removing filters is technically equivalent to removing channels \cite{he2023structured}. In this work, we use channel pruning by default and may interchangeably use filter pruning and channel pruning.} and layer pruning, is another popular branch, which removes the less important channels or layers to reduce the network complexity as shown in Fig.~\ref{fig:structured-pruning-vs-non-structured-pruning}. In practice, layer pruning is a special case of channel pruning and channel pruning is equivalent to layer pruning when all the channels in the same layer are removed. We emphasize that non-structured pruning, despite being able to achieve significant compression ratios, introduces considerable irregular network sparsity and the resulting pruned network also features irregular computational patterns, which highly relies on specialized hardware accelerators to achieve realistic speedups as demonstrated in \cite{han2016eie}. In contrast, structured pruning can easily achieve realistic speedups on mainstream hardware such as GPUs and CPUs, thanks to its high on-device parallelisms \cite{liu2018rethinking}. This unique technical merit has been making structured pruning more and more popular, especially in the context of designing hardware-friendly network solutions \cite{he2023structured}. With the above in mind, below we further elaborate on recent representative structured pruning works, which can be roughly divided into the following four categories, including weight-based pruning, activation-based pruning, batch normalization statistics-based pruning, and search-based pruning.

\textbf{Weight-Based Pruning.}
Weight-based pruning, also referred to as weight-dependent pruning, determines the importance of different channels based on the corresponding weights, which is technically similar to magnitude-based pruning as discussed in Section~\ref{sec:non-structured-pruning}. There have been two popular weight-based pruning criteria, including weight norm and weight correlation. Without loss of generality, we can easily calculate the $L_n$-norm as $||w||_n$, where $w$ is the corresponding network weight. For example, \cite{li2016pruning} proposes to remove the less important channels based on their $L_1$-norm values, which indicates that the channel with smaller $L_1$ is considered less important and contributes less to the network output. Besides, \cite{he2018soft} observes that $L_2$-norm can achieve better pruning performance than $L_1$-norm. Furthermore, \cite{he2019filter} challenges the empirical assumption in \cite{li2016pruning, he2018soft} and demonstrates that the channels with smaller $L_1$- and $L_2$-norm magnitudes are not necessarily less important. To avoid this, \cite{he2019filter} instead turns back to the channel correlation, which reveals that the channels close to the geometric median are typically redundant since they represent similar feature maps in the same layer. As a result, removing the channels close to the geometric median only leads to minimal accuracy loss. Inspired by the promising performance of \cite{he2019filter}, \cite{yvinec2021red, yvinec2022red++} propose to first apply scalar hashing on the weights of each layer and then remove redundant channels based on the corresponding weight similarity. This is because similar channels are of high redundancy in terms of the contributions to the network representation capability. In addition, unlike \cite{li2016pruning, he2018soft, he2019filter, yvinec2021red, yvinec2022red++} that measure the channel redundancy in the same layer, \cite{wang2019cop} investigates the channel redundancy across multiple different layers in order to minimize the accuracy loss. Furthermore, \cite{wang2021convolutional} prioritizes to remove the channels in more redundant layers rather than globally ranking different channels across all the network layers.

\textbf{Activation-Based Pruning.}
Activation-based pruning typically leverages the intermediate activation maps to interpret the importance of different channels, in which activation maps, also known as feature maps, correspond to the output features from one specific network layer. As demonstrated in \cite{he2023structured}, there have been three representative techniques to determine the importance of different channels in the $L$-th layer, including (1) using the activation maps of the $L$-th layer, (2) using the activation maps of adjacent layers (e.g., $(L+1)$-th and $(L+2)$-th layers), and (3) using the activation maps of the last layer (i.e., the network output):
\begin{itemize}
    \item[(1)] {
    \textit{Current Layer.} 
    To determine the importance of different channels in the $L$-th layer, \cite{he2017channel} proposes a simple yet effective two-step scheme, which first removes different channels and then measures the reconstruction error based on the output activation maps of the $L$-th layer. And next, the channels that lead to smaller reconstruction error are removed to reduce the network complexity. Similarly, \cite{lin2020hrank} measures the channel importance according to the decomposition error. Furthermore, subsequent works utilize the channel independence \cite{sui2021chip} and post-activation maps \cite{tan2020dropnet} to measure the channel importance.     
    }
    \item[(2)] {
    \textit{Adjacent Layers.}
    Recent state-of-the-art DNNs are naturally coupled and of sequential layer structures, which indicates that there has significant layer dependency between different adjacent layers. In sight of this convention, \cite{luo2017thinet, ding2019approximated} investigate the dependency between the current layer and the subsequent layer in order to measure the channel importance in the current layer. In parallel, \cite{lin2017runtime, li2021dynamic} demonstrate that the activation maps of previous layers can also reflect the channel importance in the subsequent layers.
    }
    \item[(3)] {
    \textit{Last Layer.}
    The channel importance can also be evaluated using the activation maps of the last layer, which correspond to the network output. The rationale behind this is that we are allowed to use the network output to interpret the accuracy of the pruned network. For example, we can simply determine the channel importance based on the reconstruction error \cite{lee2018snip} and the discrimination of the entire network \cite{zhuang2018discrimination}.
    }
\end{itemize}

\textbf{Statistics-Based Pruning.}
Statistics-based pruning refers to those that exploit batch normalization statistics \cite{ioffe2015batch} to interpret the channel importance, which has since gained increasing popularity thanks to its conceptual simplicity and surprisingly strong pruning performance. As shown in previous representative networks \cite{he2016deep, huang2017densely, howard2017mobilenets, mehta2022mobilevitv2, zhang2018shufflenet, ma2018shufflenet}, batch normalization is a widely used plug-and-play technique to accelerate and stabilize the network training process towards better training convergence, while also reducing \textit{internal covariate shift} to benefit the network accuracy. Specifically, batch normalization $BN(\cdot)$ transforms the input $x \in \mathbb{R}^{B \times C \times W \times H}$ as follows:
\begin{equation}
    \label{eq:batch-normalization}
    BN(x) = \gamma \cdot \frac{x - \mu}{\sqrt{\delta^2 + \epsilon}} + \beta
\end{equation}
where $\mu$ and $\delta$ are the mean and standard deviation of the input $x$, respectively. And $\epsilon$ is a small constant (e.g., $1\times10^{-9}$) to avoid zero-division. Besides, $\gamma \in \mathbb{R}^{B}$ and $\beta \in \mathbb{R}^{B}$ are learnable parameters to scale and shift $\frac{x - \mu}{\sqrt{\delta^2 + \epsilon}}$, which are optimized during the training process to recover the input $x$. Note that $\gamma$ and $\beta$ have the same dimension as the number of input channels. As seen in \cite{he2016deep, huang2017densely, howard2017mobilenets, mehta2022mobilevitv2, zhang2018shufflenet, ma2018shufflenet}, it is common practice to insert one batch normalization layer after one convolutional layer. In sight of these, \cite{liu2017learning} pioneers to leverage batch normalization statistics $\gamma$ to enable and disable different input channels, among which those disabled channels are pruned at the end of the training process. To this end, \cite{liu2017learning} applies $L_1$-norm regularization on $\gamma$ to achieve sparsity. Similar to \cite{liu2017learning}, Gate Decorator \cite{you2019gate} introduces gated batch normalization that leverages batch normalization statistics $\gamma$ as channel gates to enable and disable different input channels from the previous convolutional layer. To achieve sparsity, Gate Decorator also exploits $L_1$-norm regularization to penalize $\gamma$ during the training process. In addition, Gate Decorator introduces an iterative pruning scheme, which progressively prunes redundant channels during the training process and fine-tunes the resulting pruned network to recover the accuracy. However, as demonstrated in \cite{zhuang2020neuron}, $L_1$-norm regularization suffers from inferior discrimination between different channels since $L_1$-norm regularization pushes all the scaling factors $\gamma$ towards zero. To tackle this, different from \cite{liu2017learning, you2019gate} that regularize $\gamma$ with $L_1$-norm penalty, \cite{zhuang2020neuron} instead polarizes $\gamma$ to enforce a large margin between zero and non-zero $\gamma$. Furthermore, \cite{ye2018rethinking} challenges \cite{liu2017learning, you2019gate, zhuang2020neuron}, which observes that smaller non-zero $\gamma$ does not imply that the corresponding channel is less important. Based on this observation, \cite{ye2018rethinking} introduces a simple yet effective iterative pruning approach, which (1) prunes the less important channels with exact zero $\gamma$, (2) rescales the magnitude of $\gamma$, and (3) fine-tunes the resulting pruned network to recover the accuracy. In addition to the scaling factors $\gamma$, \cite{kang2020operation} demonstrates that the shifting factors $\beta$ can also be leveraged to interpret the channel importance and jointly considering $\gamma$ and $\beta$ has the potential to achieve more reliable channel pruning. The aforementioned statistics-based channel pruning works \cite{liu2017learning, you2019gate, zhuang2020neuron, ye2018rethinking, kang2020operation} explicitly demonstrate that batch normalization statistics (i.e., $\gamma$ and $\beta$), when properly engineered, can reflect the importance of input channels from the previous convolutional layer.

\begin{figure}[t]
    \begin{center}
    \includegraphics[width=0.85\columnwidth]{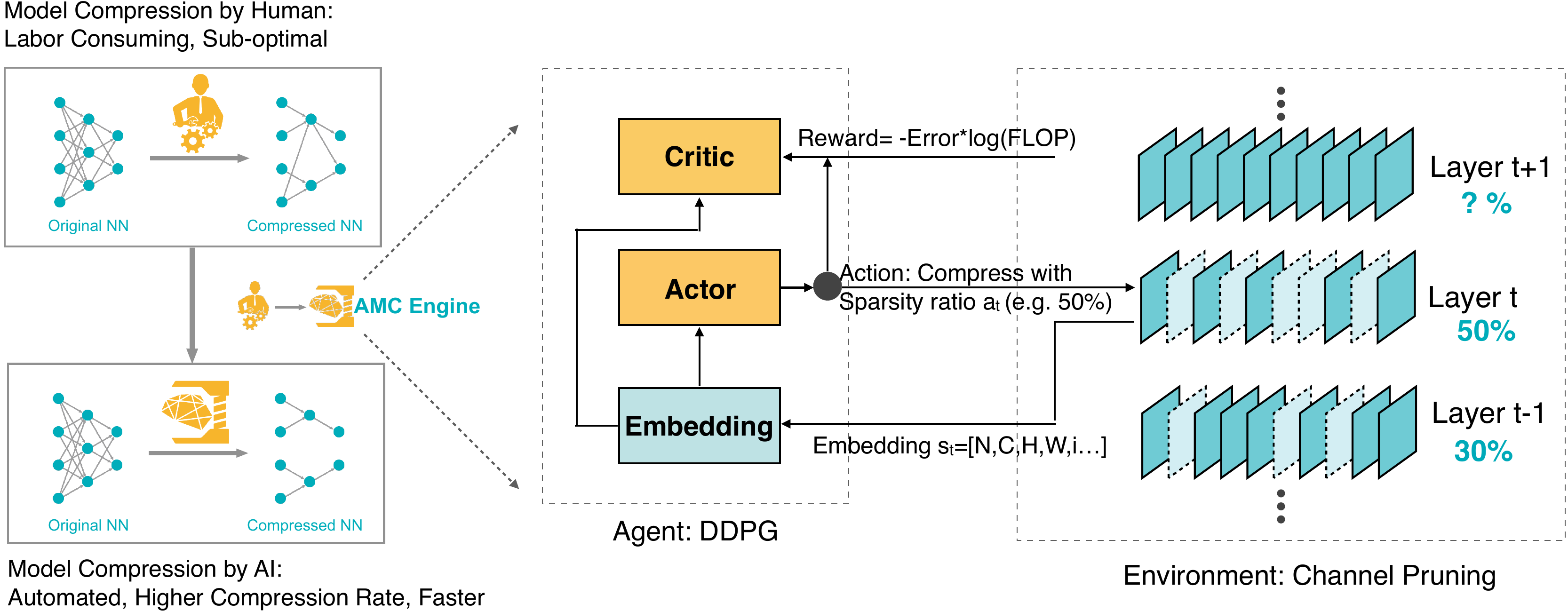}
    \end{center}
    \vspace{-5pt}
    \caption{Overview of AMC \cite{he2018amc}, which formulates channel pruning as reinforcement learning-based search and instead automatically searches for the less important channels to be pruned. \textbf{(figure from \cite{he2018amc})}}
    \vspace{-5pt}
    \label{fig:overview-amc}
\end{figure}

\textbf{Search-Based Pruning.} 
Inspired by the tremendous success of neural architecture search (NAS) as discussed in Section~\ref{sec:automated-network-design-for-embedded-computing-systems}, a plethora of search-based pruning works have recently emerged, which typically leverage search-based techniques, including reinforcement learning-based \cite{he2018amc, yu2021auto, alwani2022decore}, evolutionary algorithm-based \cite{liu2019metapruning, lin2021channel, li2022sub}, and gradient-based search \cite{li2020dhp, guo2020dmcp, ning2020dsa}, to automatically search for the optimal pruning policy, instead of using manually designed pruning heuristics. The rationale behind this is that different channels can be alternatively viewed as a list of possible operator candidates, making it possible to generalize the novel findings and advances in the field of NAS to address research challenges in the field of pruning. Specifically, previous search-based channel pruning works can be divided into the following three categories:
\begin{itemize}
    \item [(1)] {
    \textit{Reinforcement Learning-Based Search.}
    Reinforcement learning is a well-established technique for solving search problems as discussed in Section~\ref{sec:search-strategy}. Several pruning works \cite{he2018amc, yu2021auto, alwani2022decore} have pioneered to exploit reinforcement learning to search for the optimal channel pruning policy. For example, AMC \cite{he2018amc} proposes to train an efficient deep deterministic policy gradient (DDPG) \cite{lillicrap2015ddpgrl} agent such that the well-trained DDPG agent can output the optimal layer-wise channel pruning policy to maximize the pre-defined reward function as shown in Fig.~\ref{fig:overview-amc}. In addition, AGMC \cite{yu2021auto} demonstrates that AMC may yield sub-optimal pruning policies due to the fixed number of environment states. To tackle this, AGMC instead leverages graph convolutional networks (GCNs) to encode the pruned network and exploits the graph-based encoder-decoder to automatically learn the optimal environment state. Furthermore, DECORE \cite{alwani2022decore} turns back to multi-agent search, which assigns each agent to one specific network layer to learn better pruning policies.
    }
    \item [(2)] {
    \textit{Evolutionary Algorithm-Based Search.}
    Evolutionary algorithm is another well-established search technique, thanks to its conceptual simplicity, flexibility, and surprisingly strong performance. There have been several pruning works \cite{liu2019metapruning, lin2021channel, li2022sub} that employ evolutionary algorithm to automatically search for the optimal channel pruning policy. For example, MetaPruning \cite{liu2019metapruning} introduces an efficient two-stage pruning pipeline. In the first stage, MetaPruning trains an over-parameterized PruningNet that consists of all the possible pruned network configurations. Note that PruningNet here is technically the same as the supernet in the field of NAS as discussed in Section~\ref{sec:automated-network-design-for-embedded-computing-systems}. Next, in the second stage, MetaPruning leverages the well-trained PruningNet to quickly evaluate the accuracy of different pruned networks with inherited weights from the well-trained PruningNet \cite{pham2018efficient}. In the meantime, an evolutionary engine is integrated to explore the optimal pruned network.
    }
    \item [(3)] {
    \textit{Gradient-Based Search.}
    Unlike the aforementioned reinforcement learning-based and evolutionary algorithm-based pruning works that explore the optimal pruning policy within the discrete space, gradient-based search instead allows to learn the optimal pruning policy within the continuous space \cite{li2020dhp, guo2020dmcp, ning2020dsa}. As a result, gradient-based search is able to maintain much better computational efficiency than reinforcement-learning-based and evolutionary algorithm-based counterparts. For example, DSA \cite{ning2020dsa} proposes an efficient differentiable sparsity allocation approach to learn optimal layer-wise pruning ratios with gradient-based optimization, in which each pruning experiment only requires about 5 GPU-hours. To relax the discrete search space to become continuous, DSA introduces learnable pruning ratios, which are conceptually the same as the architecture parameters in differentiable NAS \cite{liu2019darts}. During the training process, the above learnable pruning ratios can be jointly optimized together with the network weights using standard gradient descent. 
    }
\end{itemize}
In general, search-based pruning is similar to NAS, in which search-based pruning searches for pruned network structures and NAS searches for stand-alone network structures. This indicates that we can generalize more advanced NAS algorithms to search for better pruned networks.

\textbf{Layer-Based Pruning.}
Layer pruning is a special case of channel pruning, which aggressively removes all the channels in the same layer as shown in Fig.~\ref{fig:structured-pruning-vs-non-structured-pruning}. In practice, under similar compression ratios, layer pruning can achieve better performance in terms of latency reduction than channel pruning as demonstrated in \cite{elkerdawy2020filter}. However, there is no free lunch, which indicates that layer pruning may suffer from greater accuracy loss than channel pruning. Note that layer pruning is conceptually and technically similar to channel pruning. Specifically, channel pruning aims to remove the less important channels, whereas layer pruning focuses on removing the less important layers as seen in previous representative layer pruning works \cite{jordao2020discriminative, chen2018shallowing, elkerdawy2020filter, tang2023sr, zhang2022layer, jordao2023layers}. In sight of this, the aforementioned channel pruning techniques can be easily generalized to prune redundant layers. For example, \cite{elkerdawy2020filter} introduces several importance criteria from the lens of channel pruning, such as weight magnitudes, activation maps, and batch normalization statistics, which are further combined to reliably determine the less important layers. Besides, \cite{jordao2023layers} investigates the lottery ticket hypothesis \cite{frankle2018lottery} from the perspective of layer pruning, which confirms that there also exist winning tickets at initialization in terms of layer pruning. More importantly, the winning tickets here are more environment-friendly with less carbon emission, while at the same time achieving better training efficiency and adversarial robustness \cite{jordao2023layers}. In addition, several recent methods \cite{luo2024pearls, luo2024domino} observe that the intermediate non-linear activation layers can also be grafted with negligible accuracy loss. Based on this observation, \cite{luo2024pearls, luo2024domino} propose to first graft the less important intermediate non-linear activation layers with their linear counterparts and then reparameterize multiple consecutive linear layers into one single linear layer to explore shallow network solutions with fewer layers. Furthermore, several recent pruning methods \cite{kong2022smart, kong2023towards, kong2023emnape, kong2023edgecompress} focus on multi-dimensional pruning, which strive to actively prune less important channels, layers, and input resolutions to aggressively trim down the model's complexity towards enhanced inference efficiency on target hardware. These multi-dimensional pruning methods can achieve much better accuracy-efficiency trade-offs than traditional channel-based and layer-based pruning methods. Similarly, HACScale \cite{kong2022hacscale} proposes an effective scaling paradigm to re-scale different channels and layers towards more efficient inference on target hardware.

\tikzset{
basic/.style = {draw, font=\footnotesize, rectangle},
root/.style = {basic, font=\footnotesize,  text width=3cm, rounded corners=2pt, thin, align=left, fill=color1!18},
parent/.style = {basic, text width=3.6cm, rounded corners=1.5pt, thin, align=left, fill=color1!7},
child/.style = {basic, text width=5.25cm, rounded corners=1.5pt, thin, align=left, fill=color1!1.5},
}
\forestset{
  main'/.style={
    l sep=5mm,
    anchor=west,
  },
  root'/.style={root,
    anchor=west,
    edge path={
     \noexpand\path[\forestoption{edge}]
     ($(!u.east)!.0!(!.west)$) ++(0.5,0) |- (!.west);
    },
  },
  parent'/.style={parent,
    anchor=west,
    calign=child edge,
    l sep=0.4cm
  },
}
\begin{figure*}[t]
    \hspace{-0.85cm} 
    \centering
\begin{forest}
    for tree={
        forked edges,
        text centered,
        grow'=east,
        reversed=true,
        font=\footnotesize,
        rectangle, /tikz/align=left, anchor=base west, tier/.pgfmath=level(),
        rounded corners,
    }
    [, main'
    [Quantized Networks, root'
        [Binarized Networks, parent'
                [
                    \cite{courbariaux2015binaryconnect, hubara2016binarized, rastegari2016xnor, bulat2019xnor, liu2018bi, qin2020forward, lin2020rotated, lin2022siman, falkena2023lab, tu2022adabin}
                    , child
                ]
        ]
        [Ternarized Networks, parent'
                [
                    \cite{li2016ternary, zhu2016trained, alemdar2017ternary, mellempudi2017ternary, li2021trq, li2020rtn, chen2021fatnn, xu2022soft}
                    , child
                ]
        ]
        [INT8 Quantized Networks, parent'
                [
                    \cite{vincent2011improving, vanholder2016efficient, kim2021performance, zhang2023spaceevo, bhandare2019efficient, jacob2018quantization}
                    , child
                ]
        ]
        [Mixed-Precision Networks, parent'
                [
                    \cite{wan2018tbn, faraone2018syq, choi2018pact, micikevicius2017mixed, das2018mixed, jia2018highly, kuchaiev2018openseq2seq}
                    , child
                ]
        ]
    ]
    [Quantization Extensions, root'
        [Quantization-Aware Training, parent'
                [
                    \cite{jacob2018quantization, zhu2020towards, zhao2021distribution, tailor2020degree, nagel2022overcoming, sakr2022optimal, youn2022bitwidth}
                    , child
                ]
        ]
        [Automated Mixed-Precision, parent'
                [
                    \cite{wu2018mixed, wang2019haq, dong2019hawq, yu2020search, chen2021towards, cai2020rethinking, wang2021generalizable, habi2020hmq, yang2020searching}
                    , child
                ]
        ]
        [Quantization Accelerators, parent'
                [
                    \cite{andri2016yodann, guo2018fbna, conti2018xnor, jain2020tim, scherer2021cutie, zhu2022fat, kim2020exploiting, sun2022film, lee2019energy, huang2021mixed}
                    , child
                ]
        ]
    ]
    ]
\end{forest}
\caption{Illustration of different network quantization techniques that have been discussed in Section~\ref{sec:network-quantization}.}
\vspace{-5pt}
\label{fig:comparisons-quantization}
\end{figure*}
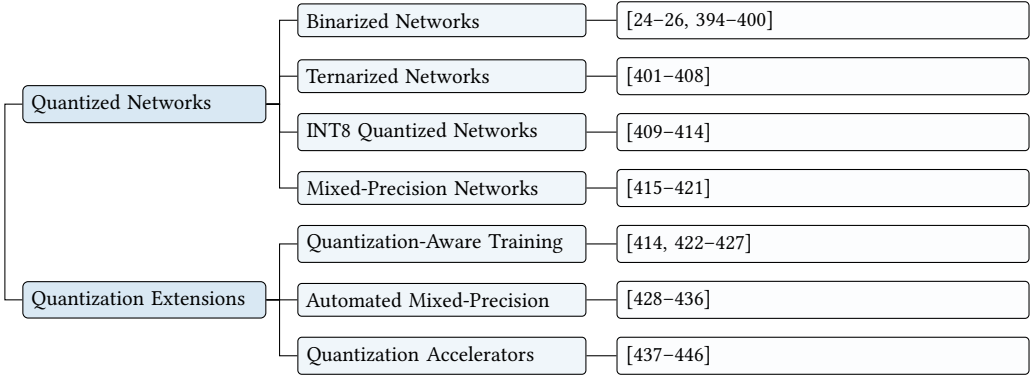

\subsection{Network Quantization}
\label{sec:network-quantization}

Different from network pruning that aims to reduce the network complexity at the structure level, network quantization instead focuses on representing the network weights and activations with lower bits, which is able to significantly reduce the network complexity at the precision level. Therefore, the resulting quantized network maintains the same network structure (i.e., the same number of layers and channels), but with lower-bit network weights and activations. In practice, network quantization can be traced back to the 1990s, when early quantization works pioneer to quantize the network weights for Boltzmann machines \cite{balzer1991weight}, optical networks \cite{fiesler1990weight}, and multi-layer perceptrons (MLPs) \cite{dundar1995effects}. Note that quantization has the potential to significantly trim down the network size to accommodate the limited storage in real-world embedded scenarios. For example, Deep Compression \cite{han2015deep} is able to reduce the network size of VGGNet by $\times$49 times, from 552\,MB to 11.3\,MB, while delivering comparable accuracy on ImageNet \cite{deng2009imagenet}. Thanks to its surprisingly strong performance in reducing the computational complexity and alleviating the storage requirements, renewed research interest in network quantization has emerged since the 2010s \cite{vincent2011improving}, which demonstrates that, compared with full-precision weights (i.e., 32 bits), 8-bit quantized weights can effectively accelerate the network inference on mainstream CPUs without significant accuracy degradation. To this end, we, in this section, discuss recent advances in the field of network quantization, including representative quantized networks and popular quantization-related extensions/implementations, which are also summarized in Fig.~\ref{fig:comparisons-quantization}.

\subsubsection{Quantized Networks}
Below we introduce several representative quantized networks, including binarized networks, ternarized networks, INT8 networks, and mixed-precision networks.

\textbf{Binarized Networks.}
Binarized networks are built upon only 1-bit weights, which are constrained to be either $+$1 or $-$1 during forward and backward propagations \cite{hubara2016binarized, courbariaux2015binaryconnect}. This can effectively eliminate computation-intensive multiply-accumulate operations and allows to replace multiply-accumulate operations with cheap additions and subtractions, and as a result, can lead to significant performance improvement in terms of latency and energy consumption as demonstrated in \cite{courbariaux2015binaryconnect}. In the relevant literature, BinaryConnect \cite{courbariaux2015binaryconnect} and BinaryNet \cite{hubara2016binarized} are the very early seminal binarized networks, which pioneer to investigate the efficacy of 1-bit weights in order to reduce the computational complexity and alleviate the storage bottleneck. Specifically, BinaryConnect \cite{courbariaux2015binaryconnect} introduces the first binarized network, which explores both deterministic binarization:
\begin{equation}
\label{eq:deterministic-binarization}
\begin{aligned}
    w_b = \mathrm{sign}(w) = 
    \begin{cases}
    +1, & \text{if} \,\,\, w \geq T \\
    -1, & \text{otherwise}
    \end{cases}
\end{aligned}
\end{equation}
and stochastic binarization to stochastically binarize the network weights:
\begin{equation}
\label{eq:stochastic-binarization}
\begin{aligned}
    w_b = 
    \begin{cases}
    +1, & \text{with probability} \,\,\, p=\sigma(w) \\
    -1, & \text{with probability} \,\,\, p=1-\sigma(w)
    \end{cases} 
\end{aligned}
\end{equation}
where $\sigma(\cdot)$ is the hard sigmoid function and can be mathematically formulated as follows:
\begin{equation}
\label{eq:hard-sigmoid}
    \sigma(x) = \mathrm{clip}(\frac{x+1}{2}, 0, 1) = \mathrm{max}(x, \mathrm{min}(1, \frac{x+1}{2}))
\end{equation}
Note that BinaryConnect exploits the above hard sigmoid function rather than the soft version because it is far less computationally expensive and can still yield competitive results. As shown in BinaryConnect, the stochastic binarization is more advanced and can achieve much better quantization accuracy than the deterministic counterpart. So far, BinaryConnect only enables weight-level binarization, whereas the network inputs are still required to be full-precision. In sight of this, BinaryNet \cite{hubara2016binarized} extends BinaryConnect to support both binarized weights and binarized inputs in order to maximize the inference efficiency of binarized networks. Furthermore, XNOR-Net \cite{rastegari2016xnor} demonstrates that \cite{courbariaux2015binaryconnect, hubara2016binarized} cannot be generalized to large-scale datasets like ImageNet. To address this, XNOR-Net introduces an effective approach to estimate binarized weights to maintain $w \approx \alpha \cdot w_b$, after which the estimated $\alpha$ can be detached from the binarized weight to rescale the input. To further enhance the binarization accuracy, a plethora of binarized networks \cite{bulat2019xnor, liu2018bi, qin2020forward, lin2020rotated, lin2022siman, falkena2023lab, tu2022adabin} have been subsequently proposed. For example, \cite{falkena2023lab, tu2022adabin} propose learnable activation binarizer \cite{falkena2023lab} and adaptive binary sets \cite{tu2022adabin} to explore more accurate binarized networks.

\begin{figure}[t]
    \begin{center}
    \includegraphics[width=0.85\columnwidth]{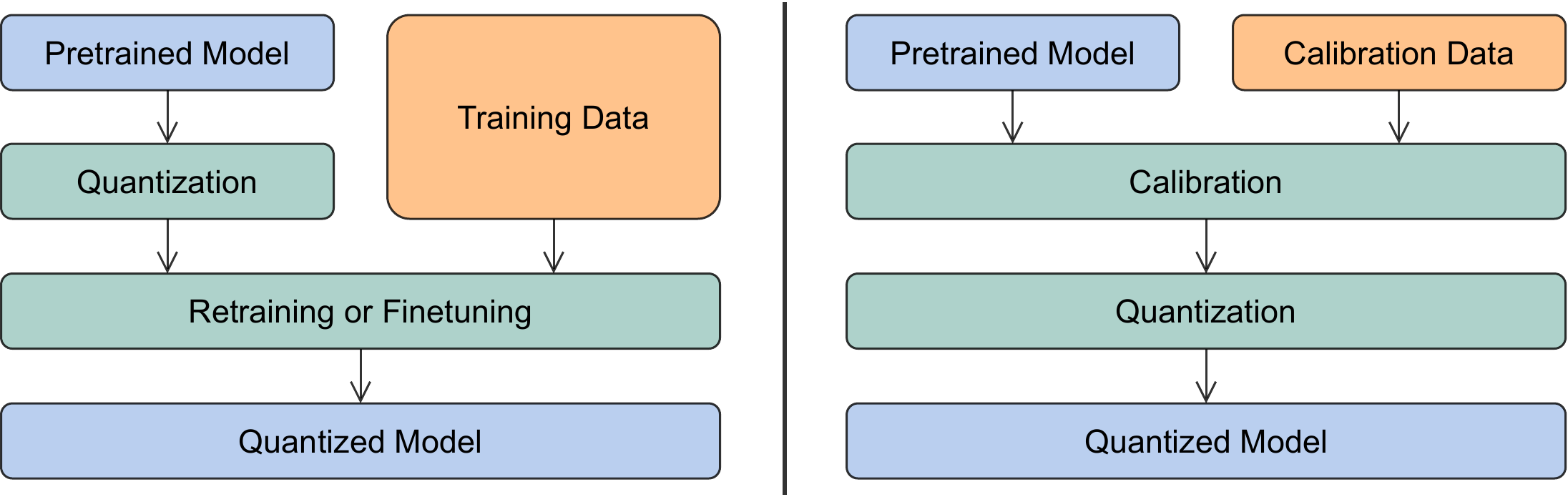}
    \end{center}
    \vspace{-5pt}
    \caption{Comparisons between post-training quantization and quantization-aware training. Different from post-training quantization, quantization-aware training integrates the quantization loss into the training loss, which allows the optimizer to minimize the quantization loss so as to improve the quantization accuracy.}
    \vspace{-5pt}
    \label{fig:post-training-quantization-vs-quantization-aware-training}
\end{figure}

\textbf{Ternarized Networks.}
In addition to binarized networks, ternarized networks \cite{li2016ternary, zhu2016trained} are another representative branch of quantized networks and have gained increasing popularity, thanks to their superior accuracy. Specifically, ternarized networks quantize the network weights from 32 bits to 2 bits, in which the 2-bit weights are constrained to $-$1, 0, and $+$1 in contrast to $\pm$1 in binarized networks. As such, ternarized networks can achieve much better accuracy than binarized networks at the cost of slightly increased computational complexity. To achieve this, \cite{li2016ternary}, as the first ternarized network, proposes to quantize the full-precision weights as follows:
\begin{equation}
\label{eq:ternary-quantization}
\begin{aligned}
    w_t =  
    \begin{cases}
    +1, & \text{if} \,\,\, w > \Delta \\
    0, & \text{if} \,\,\, |w| \leq \Delta \\
    -1, & \text{if} \,\,\, w < -\Delta
    \end{cases}
\end{aligned}
\end{equation}
where $\Delta$ is a positive constant to control the ternarization threshold. To derive the optimal ternarization threshold $\Delta^*$, \cite{li2016ternary} turns back to XNOR-Net \cite{rastegari2016xnor} and borrows the binarization estimating scheme from XNOR-Net, which introduces an adjustable scaling factor to minimize $||w - \alpha \cdot w_t||_2^2$. Finally, \cite{li2016ternary} demonstrates an empirical rule of thumb to derive $\Delta^*$ as follows:
\begin{equation}
\label{eq:ternary-quantization-solution}
    \Delta^* = 0.7 \cdot E(|w|) \approx \frac{0.7}{n} \sum_{i=1}^n |w_i|
\end{equation}
where $n$ is the number of elements within $w$. To boost the ternarization accuracy, \cite{zhu2016trained} further introduces two trained scaling coefficients $w_l^p$ and $w_l^n$ for the $l$-th layer, which are then trained using gradient descent during backward propagation. Once the training process terminates, \cite{zhu2016trained} deploys the ternarized networks on target hardware, including the trained ternarized weights and the corresponding scaling coefficients, reducing the network size by at least $\times$16 times. Subsequently, several ternarized networks \cite{alemdar2017ternary, mellempudi2017ternary, li2021trq, li2020rtn, chen2021fatnn, xu2022soft} have been proposed to further improve the ternarization accuracy. Among them, \cite{xu2022soft} demonstrates that the hard ternarization threshold $\Delta$, despite being simple and effective, often leads to sub-optimal results. To avoid this, \cite{xu2022soft} introduces the paradigm of soft ternarization threshold, which instead enables the network to automatically determine the optimal ternarization intervals to maximize its accuracy.

\newlength{\columnsepsavesavesavesavesavesavesave} 
\setlength{\columnsepsavesavesavesavesavesavesave}{\columnsep} 
\setlength{\columnsep}{6pt} 
\begin{wrapfigure}{r}{0.45\columnwidth}
    \begin{center}
        \includegraphics[width=0.45\columnwidth]{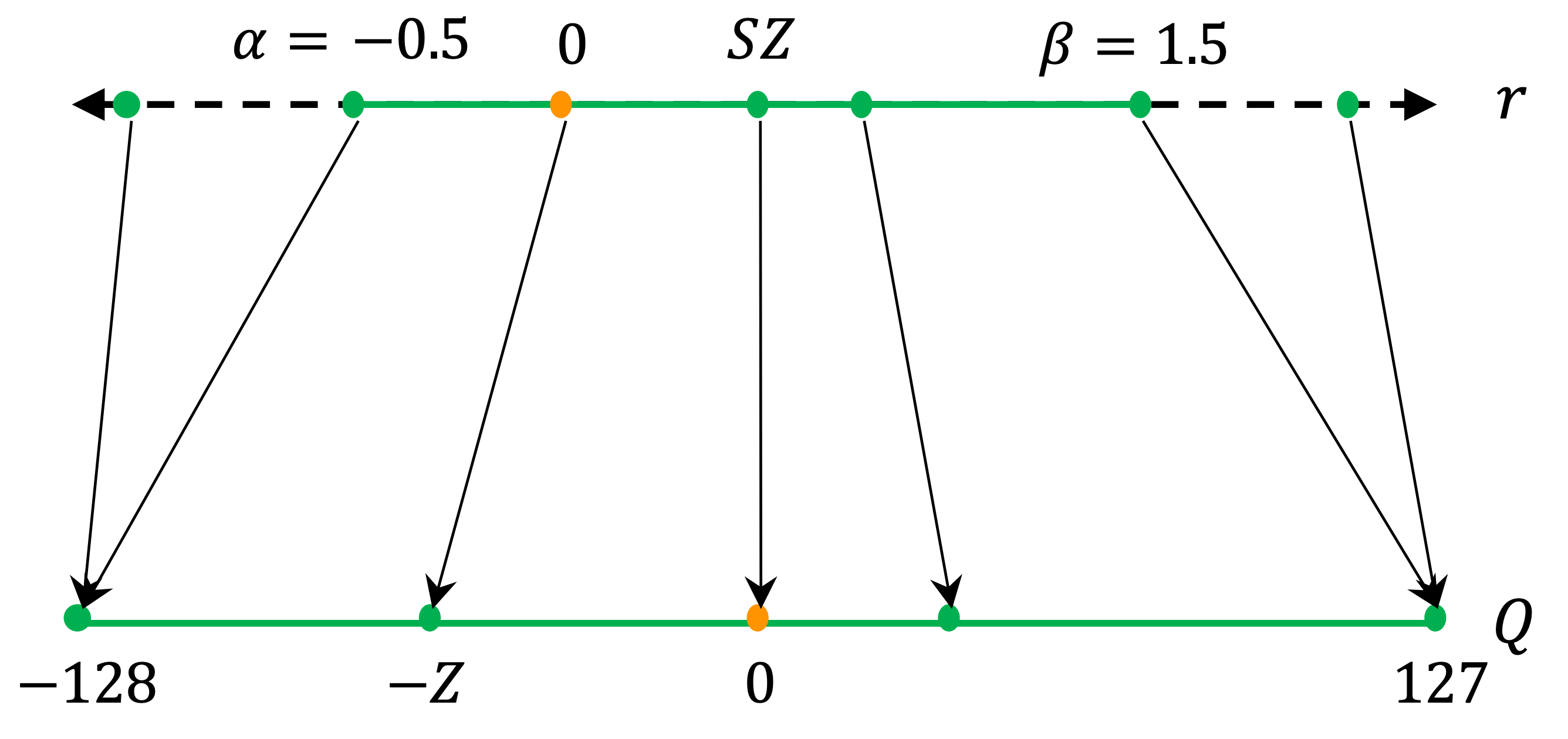}
    \end{center}
    \caption{
    TensorRT's INT8 calibration \cite{vanholder2016efficient}.
    }
    \label{fig:int8-calibration}
\end{wrapfigure}
\textbf{INT8 Quantization.}
Binarized and ternarized networks have the potential to achieve $\times$16$\sim$$\times$32 speedups, which, however, suffer from non-negligible accuracy loss, and even worse, require considerable engineering efforts to design specialized hardware for further deployment. The rationale behind this is that mainstream hardware does not support low-bit quantized networks. To overcome such limitations, an effective alternative is INT8 quantization, which trims down the network weights from 32 bits to 8 bits within the range of $[-128, 127]$. As such, INT8 quantization can lead to about $\times$4 compression in terms of the network size, and more importantly, at the cost of negligible accuracy loss \cite{vincent2011improving}. Besides, thanks to its well-suited software (e.g., Google's TensorFlow Lite and Nvidia's TensorRT), we can easily deploy INT8 quantized networks on mainstream hardware, such as mobiles, CPUs, and edge GPUs, with minimal engineering efforts \cite{nvidia-jetson, google-edgetpu}. For example, as shown in \cite{vanholder2016efficient}, TensorRT allows post-training INT8 quantization, which leverages simple weight calibration to convert pre-trained full-precision weights into 8-bit weights (see Fig.~\ref{fig:int8-calibration}) with only trivial accuracy loss. Furthermore, several follow-up works \cite{kim2021performance, zhang2023spaceevo, bhandare2019efficient, jacob2018quantization} have been recently proposed to investigate INT8 quantization and improve INT8 quantization accuracy. Among them, \cite{jacob2018quantization}, as the very first INT8 quantization work, pioneers to quantize both weights and activations with 8-bit integers to boost the inference efficiency. Besides, \cite{kim2021performance} evaluates the performance of various INT8 quantized networks on mobile GPUs, based on which \cite{kim2021performance} introduces a unified INT8 quantization framework that integrates various off-the-shelf INT8 quantization techniques, such as symmetric, asymmetric, per-layer, and per-channel INT8 quantization. Furthermore, \cite{zhang2023spaceevo} investigates the efficiency bottleneck of INT8 quantization and introduces hardware-friendly search space design to enable efficient INT8 quantization. More recently, \cite{huai2023crossbar, huai2023crimp} explore INT8 quantization to compress redundant CNNs for efficient in-memory computing infrastructures. In addition to quantizing CNNs, \cite{bhandare2019efficient} turns back to transformers and leverages INT8 quantization to quantize computation-intensive transformers in order to boost the inference efficiency for general NLP tasks. 
\setlength{\columnsep}{\columnsepsavesavesavesavesavesavesave}

\textbf{Mixed-Precision Networks.}
Mixed-precision quantization is another well-established branch of network quantization. As shown in \cite{wan2018tbn}, mixed-precision quantization allows more fine-grained quantization schemes across different weights and activations, and as a result, can usually achieve better accuracy-efficiency trade-offs than conventional fixed-precision quantization, such as binarized (1-bit), ternarized (2-bit), and INT8 (8-bit) quantization. For example, TBN \cite{wan2018tbn}, as the very first mixed-precision network, proposes to combine layer-wise ternarized inputs and binarized weights, which delivers surprisingly better accuracy-efficiency trade-offs than stand-alone binarized networks and ternarized networks. The success of TBN has motivated several subsequent mixed-precision quantization works \cite{faraone2018syq, choi2018pact} to continue improving the quantization accuracy. For example, SYQ \cite{faraone2018syq} proposes to quantize the network weights with 1/2 bits and the intermediate activation with 8 bits, whereas PACT \cite{choi2018pact} allows 2-bit activations and 2/3/4/5-bit weights. These early mixed-precision quantization works have demonstrated promising performance. Later, we will introduce automated mixed-precision quantization, which exploits automated techniques to search for the optimal bit allocation and can achieve more fine-grained quantization. Furthermore, \cite{micikevicius2017mixed, das2018mixed, jia2018highly, kuchaiev2018openseq2seq} also consider to leverage mixed-precision quantization to improve the training efficiency of full-precision networks, which can significantly accelerate the training process, and more importantly, achieve comparable accuracy to the full-precision training.

\subsubsection{Quantization Extensions and Implementations}
Below we introduce several quantization extensions and implementations, including quantization-aware training, automated mixed-precision quantization, and quantization-aware hardware accelerators.

\textbf{Quantization-Aware Training.}
Quantization-aware training refers to the technique that trains quantized networks, which is fundamentally different from post-training quantization as shown in Fig.~\ref{fig:post-training-quantization-vs-quantization-aware-training}. Note that post-training quantization can achieve satisfactory performance on early networks like AlexNet \cite{krizhevsky2012alexnet} and VGGNet \cite{simonyan2014vggnet}, which, however, suffers from significant accuracy loss when applied to more advanced lightweight networks like MobileNets \cite{howard2017mobilenets, sandler2018mobilenetv2} and ShuffleNets \cite{zhang2018shufflenet, ma2018shufflenet}. In general, quantization-aware training incorporates the quantization loss into the training loss, which then allows the optimizer to minimize the quantization loss during the training process in order to unlock better quantization accuracy than post-training quantization. In practice, \cite{jacob2018quantization}, as the seminal quantization-aware training work, proposes to quantize both weights and activations with 8-bit integers. To maximize the accuracy of INT8 quantized networks, \cite{jacob2018quantization} also introduces an effective tailored quantization-aware training approach to train the resulting INT8 quantized networks. Similar to \cite{jacob2018quantization}, \cite{zhu2020towards, zhao2021distribution} unify and improve quantization-aware training of INT8 quantized networks to minimize the accuracy degradation. To generalize quantization-aware training to train other types of quantized networks (e.g., 1-bit and 2-bit networks), a plethora of quantization-aware training works \cite{tailor2020degree, nagel2022overcoming, sakr2022optimal, youn2022bitwidth} have been subsequently proposed, which further push forward the attainable quantization accuracy.

\textbf{Automated Mixed-Precision Quantization.}
Early quantization works typically quantize all the weights and activations with the same level of precision, such as 1 bit for binarized networks and 2 bits for ternarized networks. Despite the promising performance, early uniform quantization practices suffer from sub-optimal accuracy-efficiency trade-offs. For example, as shown in TBN \cite{wan2018tbn}, mixed-precision quantization that combines layer-wise ternarized inputs and binarized weights can achieve much better accuracy-efficiency trade-offs than stand-alone binarized networks and ternarized networks. However, determining the optimal mixed-precision quantization strategy is difficult due to the large number of possible quantization combinations across different layers. To overcome such limitations, recent research has been shifted to automated mixed-precision quantization \cite{wu2018mixed, wang2019haq, dong2019hawq, yu2020search, chen2021towards, cai2020rethinking, wang2021generalizable, habi2020hmq, yang2020searching}, thanks to the tremendous success of neural architecture search (NAS) as discussed in Section~\ref{sec:automated-network-design-for-embedded-computing-systems}. Among them, \cite{wu2018mixed}, as the first automated mixed-precision quantization work, follows early differentiable NAS practices \cite{liu2019darts, wu2019fbnet} to search for the optimal layer-wise precision assignment. Furthermore, HAQ \cite{wang2019haq} leverages reinforcement learning-based search to explore the huge quantization design space with hardware feedback in the loop, which focuses on finding the optimal layer-wise precision assignment to maximize both quantization accuracy and hardware efficiency. Note that we can easily generalize recent advances in the field of NAS to further improve automated mixed-precision quantization.

\textbf{Quantization-Aware Accelerators.}
Different from INT8 quantized neural networks, binarized, ternarized, and mixed-precision quantized neural networks are not supported by mainstream hardware, such as mobiles, CPUs, and edge GPUs. This further demands the design of quantization-aware accelerators to efficiently execute low-bit quantized networks at run time. To this end, a plethora of representative quantization-aware accelerators have been recently proposed \cite{andri2016yodann, guo2018fbna, conti2018xnor, jain2020tim, scherer2021cutie, zhu2022fat, kim2020exploiting, sun2022film, lee2019energy, huang2021mixed}, including binarized networks-based accelerators \cite{andri2016yodann, guo2018fbna, conti2018xnor}, ternarized networks-based accelerators \cite{jain2020tim, scherer2021cutie, zhu2022fat}, and mixed-precision networks-based accelerators \cite{kim2020exploiting, sun2022film, lee2019energy, huang2021mixed}. These quantization-aware accelerators have demonstrated significant efficiency improvements in terms of latency, memory, area, and energy consumption in various real-world embedded scenarios.

\tikzset{
basic/.style = {draw, font=\footnotesize, rectangle},
root/.style = {basic, font=\footnotesize,  text width=2.75cm, rounded corners=2pt, thin, align=left, fill=color1!18},
parent/.style = {basic, text width=3cm, rounded corners=1.5pt, thin, align=left, fill=color1!7},
child/.style = {basic, text width=6.25cm, rounded corners=1.5pt, thin, align=left, fill=color1!1.5},
}
\forestset{
  main'/.style={
    l sep=5mm,
    anchor=west,
  },
  root'/.style={root,
    anchor=west,
    edge path={
     \noexpand\path[\forestoption{edge}]
     ($(!u.east)!.0!(!.west)$) ++(0.5,0) |- (!.west);
    },
  },
  parent'/.style={parent,
    anchor=west,
    calign=child edge,
    l sep=0.4cm
  },
}
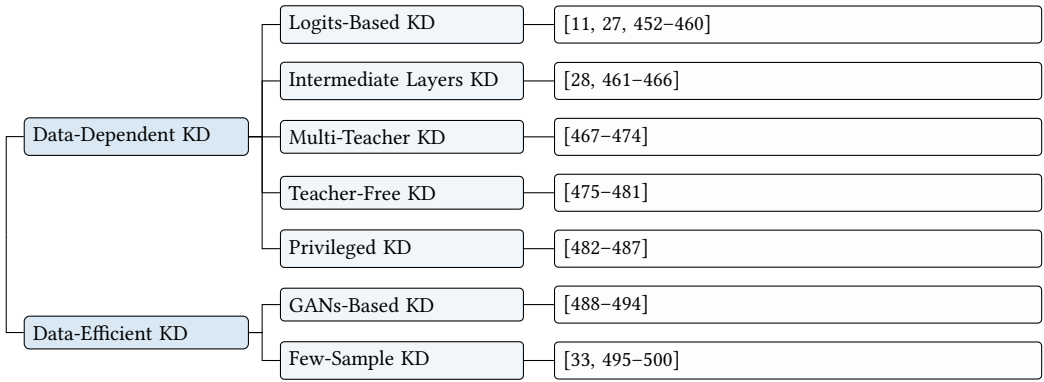
\begin{figure*}[t]
    \hspace{-0.85cm} 
    \centering
\begin{forest}
    for tree={
        forked edges,
        text centered,
        grow'=east,
        reversed=true,
        font=\footnotesize,
        rectangle, /tikz/align=left, anchor=base west, tier/.pgfmath=level(),
        rounded corners,
    }
    [, main'
    [Data-Dependent KD, root'
        [Logits-Based KD, parent'
                [
                    \cite{ba2014deep, hinton2015distilling, tian2019contrastive, hegde2020variational, wen2021preparing, cho2019efficacy, mirzadeh2020improved, beyer2022knowledge, li2017learning, xie2020self, hong2021student}
                    , child
                ]
        ]
        [Intermediate Layers KD, parent'
                [
                    \cite{romero2014fitnets, yim2017gift, kim2018paraphrasing, ahn2019variational, tung2019similarity, xu2020bert, zhou2020channel}
                    , child
                ]
        ]
        [Multi-Teacher KD, parent'
                [
                    \cite{tarvainen2017mean, you2017learning, sau2016deep, song2018collaborative, yang2020model, zhu2018knowledge, fukuda2017efficient, xiang2020learning}
                    , child
                ]
        ]
        [Teacher-Free KD, parent'
                [
                    \cite{zhang2018deep, crowley2018moonshine, zhang2019your, mobahi2020self, yun2020regularizing, ji2021refine, ge2021self}
                    , child
                ]
        ]
        [Privileged KD, parent'
                [
                    \cite{vapnik2015learning, lopez2015unifying, zhao2022progressive, tang2019retaining, wang2019adversarial, xu2020privileged}
                    , child
                ]
        ]
    ]
    [Data-Efficient KD, root'
        [GANs-Based KD, parent'
                [
                    \cite{chen2019data, fang2019data, qu2021enhancing, zhao2022dual, zhuang2022data, zhang2021data, fang2021contrastive}
                    , child
                ]
        ]
        [Few-Sample KD, parent'
                [
                    \cite{kulkarni2017knowledge, liu2019semantic, li2020few, kimura2018few, bai2020few, wang2022compressing, molchanov2022lana}
                    , child
                ]
        ]
    ]
    ]
\end{forest}
\caption{Illustration of data-dependent and data-efficient knowledge distillation (KD) works in Section~\ref{sec:network-distillation}.}
\vspace{-5pt}
\label{fig:comparisons-distillation}
\end{figure*}

\subsection{Network Distillation}
\label{sec:network-distillation}

\begin{wrapfigure}{r}{0.6\columnwidth}
    \begin{center}
        \includegraphics[width=0.6\columnwidth]{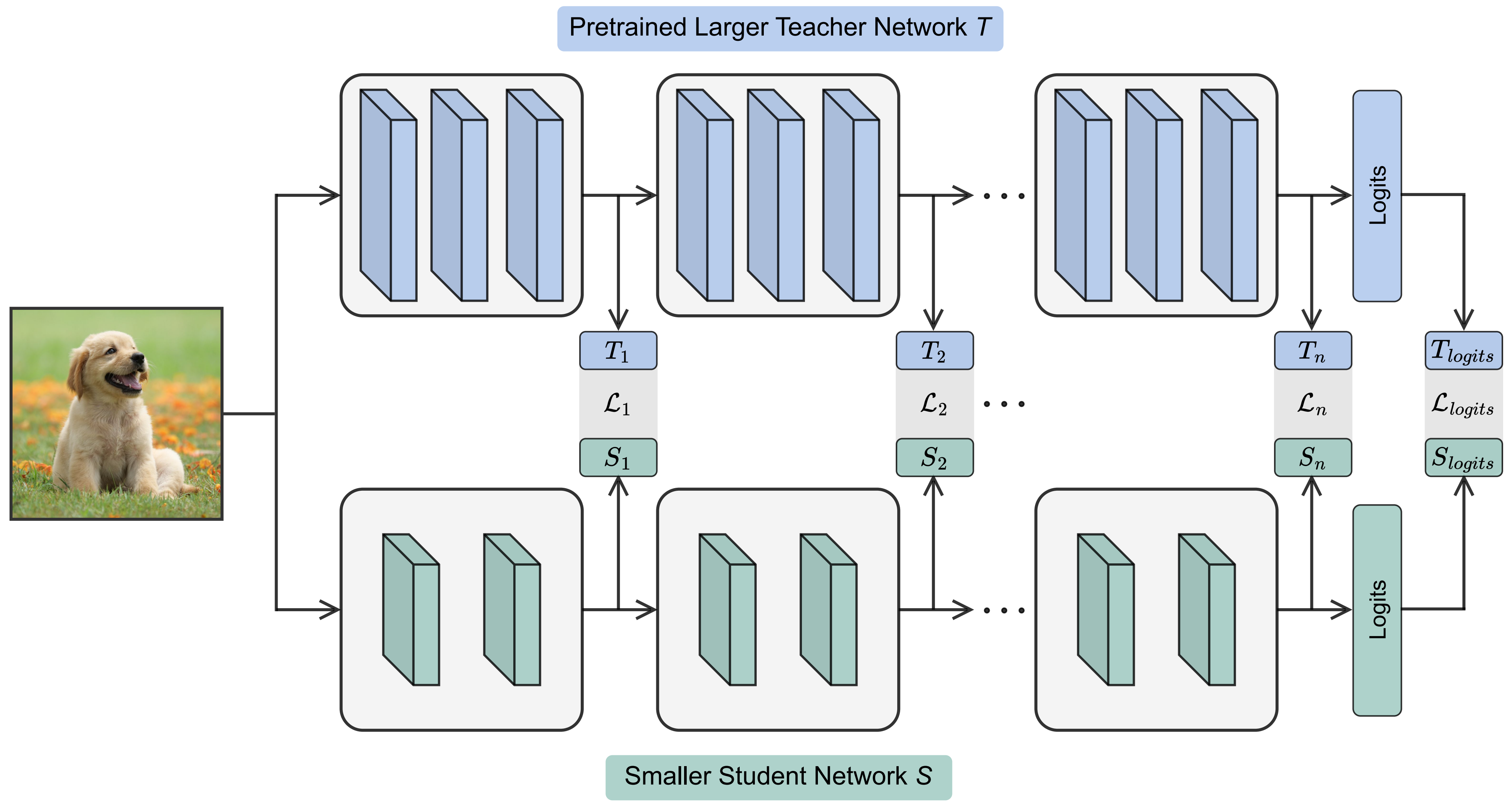}
    \end{center}
    \caption{Illustration of the teacher-student knowledge transfer process in seminal knowledge distillation techniques \cite{hinton2015distilling, romero2014fitnets}.}
    \label{fig:overview-kd}
\end{wrapfigure}

Network distillation, also referred to as knowledge distillation\footnote{We interchangeably use network distillation and knowledge distillation to refer to the distillation-based training process.}, is another well-established paradigm to further push forward the accuracy-efficiency trade-off, which is initially proposed by \cite{buciluǎ2006model} and subsequently generalized by \cite{hinton2015distilling, romero2014fitnets}. Note that knowledge distillation is a \textit{plug-and-play} training technique, which has been applied to various tasks to achieve better training performance, such as object detection \cite{chen2017learning} and language understanding \cite{sanh2019distilbert}. Different from network pruning and network quantization, which focus on improving the network efficiency without sacrificing the network accuracy as discussed in Section~\ref{sec:network-pruning} and Section~\ref{sec:network-quantization}, network distillation instead boosts the accuracy-efficiency trade-off from the accuracy perspective, which aims to improve the network accuracy without changing the network structure. In other words, unlike network pruning and network quantization that lead to simplified network structures, network distillation results in the same network but the resulting network can typically achieve higher accuracy. Specifically, as shown in \cite{buciluǎ2006model, hinton2015distilling, romero2014fitnets}, knowledge distillation refers to the training process that leverages a larger pre-trained teacher network to benefit the training process of a smaller student network (see Fig.~\ref{fig:overview-kd}), which transfers the rich and discriminative knowledge from the larger pre-trained teacher network to the smaller student network to further achieve better accuracy on target task than simply training the student network alone. Below we first present the preliminaries of knowledge distillation and then introduce recent representative data-dependent and data-efficient knowledge distillation works. These knowledge distillation works can also be found in Fig.~\ref{fig:comparisons-distillation}.

\newlength{\columnsepsavesavesavesavesavesave} 
\setlength{\columnsepsavesavesavesavesavesave}{\columnsep} 
\setlength{\columnsep}{6pt} 
\textbf{Knowledge Distillation Basics.}
In order to better understand knowledge distillation, we first elaborate on the preliminaries of knowledge distillation, which are mainly based on the most representative knowledge distillation work \cite{hinton2015distilling}. As shown in previous state-of-the-art networks \cite{howard2017mobilenets, sandler2018mobilenetv2, zhang2018shufflenet, ma2018shufflenet}, the network outputs, also referred to as the network logits, are typically fed into the softmax function to calculate the probability distribution over different categories for further prediction purposes. However, given a pre-trained teacher network, the output logits after the softmax function are discriminative but less informative, which are close to either 1 or 0 (e.g, [0.02, 0.95, 0.01, 0.01, 0.01]). This makes it difficult to directly transfer the discriminative knowledge from the pre-trained teacher network to the student network. The rationale behind this is that the student network is of smaller network size than the teacher network, thus being less capable of learning discriminative knowledge \cite{hinton2015distilling}. To mitigate this issue, \cite{hinton2015distilling} leverages the distillation temperature $T$ to soften the knowledge from the pre-trained teacher network to facilitate the teacher-student knowledge transfer process, which can be mathematically formulated as follows:
\begin{equation}
    \label{eq:knowledge-distillation-soften}
    z_i = \frac{\exp(y_i/T)}{\sum_{j=1}^n \exp(y_j/T)} \,\,\, s.t., \,\,\, i = 1, ..., n
\end{equation}
where $\{y_i\}_{i=1}^n$ denotes the output logits without softmax and $T$ is the temperature to soften the output logits $\{y_i\}_{i=1}^n$ as $\{z_i\}_{i=1}^n$. Note that Eq~(\ref{eq:knowledge-distillation-soften}) is equivalent to the standard softmax function when $T=1$. As shown in \cite{hinton2015distilling}, larger $T$ can produce softer probability distribution over different categories (e.g., $[0.1, 0.6, 0.1, 0.1, 0.1]$ when $T=5$ and $[0.2, 0.2, 0.2, 0.2, 0.2]$ when $T=+\infty$). Besides, fixing the temperature $T$ to 2 can empirically yield the best performance. Furthermore, \cite{hinton2015distilling} exploits the softened knowledge from the pre-trained teacher network to guide the training process of the student network, which can be mathematically formulated as follows:
\begin{equation}
    \label{eq:knowledge-distillation-objective}
    \mathop{\mathrm{minimize}}_{w} \,\,\, \mathcal{L}_{train}(x, w) = \mathcal{L}(y, y^*) + \alpha \cdot T^2 \cdot \mathcal{L}(y, z) \,\,\, s.t., \,\,\, y = f_w(x)
\end{equation}
where $x$ is the input data, $y^*$ is the ground-truth label, $z$ is the softened knowledge from the pre-trained teacher network, $\alpha$ is the constant to control the teacher-student distillation magnitude, and $\mathcal{L}(\cdot)$ is the standard cross entropy loss function. Apart from these, $f_w(\cdot)$ parameterizes the student network with the weight of $w$. As demonstrated in \cite{hinton2015distilling}, it is important to multiply the teacher-student distillation loss term (i.e., $\mathcal{L}(y, z)$) with $T^2$ because the teacher-student distillation loss term scales the gradient of $w$ to $1/T^2$ during the training process.
\setlength{\columnsep}{\columnsepsavesavesavesavesavesave}

With the above in mind, below we further introduce recent representative knowledge distillation practices, which are built upon \cite{hinton2015distilling} and can be roughly divided into the following two categories, including data-dependent and data-efficient knowledge distillation.

\subsubsection{Data-Dependent Knowledge Distillation}
\label{sec:data-dependent-distillation}
In this section, we introduce several representative data-dependent knowledge distillation techniques, including logits-based knowledge distillation, intermediate layers-based knowledge distillation, multi-teacher knowledge distillation, teacher-free knowledge distillation, and privileged knowledge distillation.

\textbf{Knowledge from Logits.}
Knowledge distillation from logits is one representative branch of knowledge distillation and has been widely applied to improve the network accuracy, thanks to its conceptual simplicity and surprisingly strong performance. Specifically, \cite{ba2014deep, hinton2015distilling} pioneer to leverage the logits-based knowledge from the pre-trained teacher network to facilitate the training process of the less-capable student network. As seen in \cite{hinton2015distilling}, the logits-based knowledge from the pre-trained teacher network, also referred to as soft labels, corresponds to the output of the teacher network after being fed into the softmax function to calculate the probability distribution over different categories as shown in Eq~(\ref{eq:knowledge-distillation-soften}). Subsequently, \cite{tian2019contrastive, hegde2020variational, wen2021preparing} investigate the efficacy of knowledge distillation and demonstrate that early knowledge distillation practices \cite{ba2014deep, hinton2015distilling} only yield sub-optimal results since $\alpha$ and $T$ are fixed for different teacher-student networks as shown in Eq~(\ref{eq:knowledge-distillation-objective}). Besides, \cite{cho2019efficacy, mirzadeh2020improved, beyer2022knowledge} demonstrate the accuracy of the student network may significantly degrade when there are large accuracy gaps between teacher and student. To overcome such limitations, \cite{mirzadeh2020improved} instead introduces an intermediate-sized network (i.e., teacher assistant) to facilitate the knowledge transferred from the pre-trained teacher network to the student network, thus effectively bridging the gap between powerful teacher and less-capable student. In addition to soft labels, \cite{li2017learning, xie2020self, hong2021student} demonstrate that noisy labels are also helpful to knowledge distillation, which can be leveraged to further improve the accuracy of the student network.

\textbf{Knowledge from Intermediate Layers.}
Apart from knowledge distillation from logits, knowledge distillation from intermediate layers is another representative branch of knowledge distillation, which provides more fine-grained knowledge to better guide the training process of the smaller student network. The rationale behind this is that intermediate features are also discriminative and can be combined with the final network output to further enhance the feature expressiveness as seen in \cite{luo2020person}. Specifically, \cite{romero2014fitnets} pioneers to investigate knowledge distillation from intermediate layers and introduces hint learning to improve the training process of the student network, in which hints correspond to the intermediate features of the teacher network. Compared with logits-based knowledge, knowledge from intermediate layers are often richer and more fine-grained as shown in \cite{romero2014fitnets}. Furthermore, a plethora of subsequent knowledge distillation works \cite{yim2017gift, kim2018paraphrasing, ahn2019variational, tung2019similarity, xu2020bert, zhou2020channel} have been proposed to enhance the knowledge transferred from the pre-trained teacher network to the student network, which continue to explore the rich intermediate features to facilitate the training process of the student network. For example, \cite{zhou2020channel} delves into more fine-grained channel-level knowledge distillation, leading to more fine-grained and discriminative knowledge.

\textbf{Multi-Teacher Knowledge Distillation.}
The standard knowledge distillation paradigm exploits the pre-trained knowledge from one single teacher network to guide the training process of the less-capable student network \cite{hinton2015distilling, romero2014fitnets}. Furthermore, \cite{tarvainen2017mean} demonstrates that the student network may learn richer and more discriminative knowledge from multiple teacher networks, which pushes the student network to achieve better accuracy since multiple teacher networks can provide more informative and instructive knowledge than one single teacher network. To this end, \cite{tarvainen2017mean} proposes to average the network weights of multiple teacher networks (i.e., \textit{mean teachers}) to better guide the training process of the student network. Similar to \cite{tarvainen2017mean}, several follow-up knowledge distillation works \cite{you2017learning, sau2016deep, song2018collaborative, yang2020model} propose to average the output logits of multiple pre-trained teacher networks and then exploit the averaged knowledge to enhance the training process of the student network. In addition, \cite{zhu2018knowledge, fukuda2017efficient, xiang2020learning} demonstrate that directly averaging the output logits of multiple teacher networks ignores the teacher diversity since different teacher networks may maintain different network capabilities. To avoid this, \cite{zhu2018knowledge, fukuda2017efficient, xiang2020learning} propose to actively enable and disable different teacher networks through gates during the training process to better guide the student network to learn more discriminative knowledge from different teacher networks.

\textbf{Teacher-Free Knowledge Distillation.}
Despite the promising accuracy improvement, previous knowledge distillation works \cite{hinton2015distilling, romero2014fitnets} highly rely on off-the-shelf pre-trained teacher networks, which necessitate considerable computational resources to train teacher networks. In addition, to maximize the accuracy improvement, it is also of utmost importance to design proper teacher networks, leading to additional engineering efforts. To overcome such limitations, several knowledge distillation works \cite{zhang2018deep, crowley2018moonshine, zhang2019your, mobahi2020self, yun2020regularizing, ji2021refine, ge2021self} have been recently proposed to exclude teacher networks, which instead exploit the knowledge from the student network itself to guide the training process of the student network in teacher-free manners. For example, \cite{zhang2018deep} introduces deep mutual learning, which demonstrates that the pre-trained teacher network is not necessary in the context of knowledge distillation. Instead, \cite{zhang2018deep} demonstrates that an ensemble of student networks can collaboratively learn from each other throughout the training process, and more importantly, can achieve surprisingly better training accuracy than standard knowledge distillation practices \cite{hinton2015distilling, romero2014fitnets}. This explicitly indicates that the knowledge from the student network itself can also be leveraged to improve the training process of the student network towards better training accuracy.

\textbf{Privileged Knowledge Distillation.}
Privileged knowledge distillation is a special case of knowledge distillation, where the student network has access to additional information or features that are not available to the teacher network during the training process \cite{vapnik2015learning}. In contrast to the standard knowledge distillation paradigm \cite{hinton2015distilling}, privileged knowledge distillation allows the student network to learn from both the teacher network and the additional information that is only available to the student network, which has the potential to further improve the attainable accuracy as demonstrated in \cite{vapnik2015learning}. The rationale behind privileged knowledge distillation is that the student network can leverage the additional information to improve its ability to mimic the behavior of the pre-trained teacher network. Furthermore, inspired by \cite{vapnik2015learning}, several privileged knowledge distillation works \cite{lopez2015unifying, zhao2022progressive, tang2019retaining, wang2019adversarial, xu2020privileged} have been recently proposed to improve the performance of the student network in various tasks. For example, \cite{zhao2022progressive} explores progressive privileged knowledge distillation to embrace better online action detection. In addition, \cite{xu2020privileged} introduces privileged feature distillation to improve the product recommendations of Taobao. These privileged knowledge distillation works clearly demonstrate that the student network may benefit from the additional information and knowledge to achieve better training accuracy on target task.

\subsubsection{Data-Efficient Knowledge Distillation}
\label{sec:data-efficient-distillation}
In this section, we introduce GANs-based knowledge distillation and few-sample knowledge distillation, which are of high data efficiency and can perform teacher-student distillation using a small amount of training data.

\textbf{GANs-based Knowledge Distillation.}
Despite the promising accuracy improvement, knowledge distillation is often data-driven and highly relies on sufficient training data to transfer the rich pre-trained knowledge from the teacher network to the student network, thus inevitably leading to significant engineering efforts for data preparation, such as data collection, cleaning, and labeling. As seen in the relevant literature, generative adversarial networks (GANs) have been applied to a wide range of tasks and are considered one of the most effective approaches for generating high-quality synthetic data \cite{goodfellow2014generative}. In sight of this, a plethora of GANs-based knowledge distillation works \cite{chen2019data, fang2019data, qu2021enhancing, zhao2022dual, zhuang2022data, zhang2021data, fang2021contrastive} have been recently proposed to leverage GANs to generate sufficient training data and then use the generated data to train the student network. These GANs-based knowledge distillation works have demonstrated significant data efficiency since the well-optimized generator can be used to produce a large amount of high-quality synthetic data, while at the same time achieving promising accuracy improvement on target task.

\textbf{Few-Sample Knowledge Distillation.}
In addition to GANs-based knowledge distillation, another promising direction is to perform efficient knowledge distillation to transfer the rich knowledge from the pre-trained teacher network to the student network with only a small amount of training data or only few data samples, which can also bring significant data efficiency. To achieve this, several few-sample knowledge distillation works \cite{kulkarni2017knowledge, liu2019semantic, li2020few, kimura2018few, bai2020few, wang2022compressing, molchanov2022lana} have been recently proposed. Among them, \cite{li2020few} proposes a simple yet effective solution for knowledge distillation using label-free few samples to realize both data efficiency and training/processing efficiency. Specifically, \cite{li2020few} first inserts one $1\times1$ convolutional layer at the end of each building block of the student network and then optimizes the inserted $1\times1$ convolutional layer to minimize the knowledge distillation loss, which can quickly converge using only few data samples. More recently, \cite{wang2022compressing} introduces an effective mimicking-then-replacing knowledge distillation technique to quickly train the student network with only few data samples, which maintains significant data efficiency while still achieving superior training accuracy.

\subsection{Future Envision}
In this section, we further envision several promising future trends and possible directions in the field of network compression, which are summarized as follows:
\begin{itemize}
    \item[(1)] {
    \textbf{Automated Teacher-Student Search.}
    Knowledge distillation transfers the rich knowledge from the pre-trained teacher network to the student network to facilitate the training process of the student network, which has achieved promising accuracy improvement \cite{hinton2015distilling, romero2014fitnets}. In the past, researchers empirically exploit larger networks as teacher networks and smaller networks as student networks. However, such empirical practices may lead to sub-optimal accuracy and cannot always achieve accuracy improvement. The rationale behind this is that different student networks may prefer quite different teacher networks as shown in \cite{liu2020search, dong2023diswot}. This further motivates us to design the optimal teacher-student network pair to maximize the attainable accuracy of the student network. To achieve this, one promising alternative is to leverage recent advances in the field of neural architecture search (NAS) to automatically search for the optimal teacher-student network pair.
    }
    \item[(2)] {
    \textbf{Joint Network Compression.}
    To embrace better accuracy-efficiency trade-offs, an intuitive and straightforward approach is sequential network compression, which exploits multiple network compression techniques to progressively reduce the network complexity. For example, \cite{cai2019automl} introduces a simple yet effective sequential compression pipeline, which starts with searching for an efficient network with ProxylessNAS \cite{cai2019proxylessnas} and then applies automated channel pruning \cite{he2018amc} and mixed-precision quantization \cite{wang2019haq} to further trim down the network size. However, such sequential compression pipeline has critical drawbacks, and as a result, leads to sub-optimal results. This is because the searched optimal network is not necessarily optimal for subsequent pruning and quantization. To address this, one promising future direction is joint network compression, which jointly optimizes the network structure, pruning, and quantization to yield the best accuracy-efficiency trade-off.    
    }
    \item[(3)] {
    \textbf{Federated Network Compression.}
    Federated learning is an emerging decentralized learning approach that allows multiple hardware devices to collaboratively learn the same network without sharing their raw data \cite{bonawitz2019towards}. Specifically, federated learning allows the network to be trained locally on each hardware device using its own data, in which only the network updates, rather than the raw data, are sent back to the central server for further aggregation. In sight of this, one promising future direction is federated network compression, including federated pruning, federated quantization, and federated distillation, which can significantly enhance the data privacy and protect the data security while still achieving competitive performance in terms of accuracy and training efficiency.
    }
    \item[(4)] {
    \textbf{Domain-Specific Network Compression.}
    In addition to image classification, there are also a wide range of popular downstream applications, such as object detection, tracking, and semantic segmentation, where the involved networks are still quite computation-intensive. This makes it difficult to accommodate the limited available computational resources in real-world embedded scenarios. To tackle this issue, some early practices have attempted to leverage general network compression techniques to compress domain-specific networks. For example, \cite{li2019fully, xie2020localization, chen2017learning} propose to leverage pruning \cite{xie2020localization}, quantization \cite{li2019fully}, and knowledge distillation \cite{chen2017learning} to improve the accuracy-efficiency trade-off in real-world object detection scenarios, which, however, are still under-explored and cannot simply generalize to other scenarios. Therefore, one promising future direction is domain-specific network compression, which exploits domain-specific knowledge to largely boost the network compression performance towards better accuracy-efficiency trade-offs.
    }
    \item[(5)] {
    \textbf{Mixed-Precision Training.}
    Mixed-precision training refers to the technique that trains the network with both full-precision weights and low-bit weights, which has the potential to significantly improve the training efficiency without sacrificing the accuracy. For example, PyTorch \cite{paszke2019pytorch} introduces Automatic Mixed Precision (AMP), which combines 32-bit full-precision weights with 16-bit half-precision weights during the training process. As a result, AMP achieves the same level of accuracy as the stand-alone 32-bit full-precision training, while at the same time being able to deliver about $\times$2 training speedups for convolutional networks \cite{pytorchamp}. In sight of this, one promising future direction is to leverage low-bit mixed-precision training techniques to train full-precision networks, which may aggressively push forward the training efficiency without degrading the accuracy.
    }
\end{itemize}

\section{Efficient On-Device Learning for Embedded Computing Systems}
\label{sec:efficient-on-device-learning-for-embedded-computing-systems}

\tikzset{
basic/.style = {draw, font=\footnotesize, rectangle},
root/.style = {basic, font=\footnotesize,  text width=3.8cm, rounded corners=2pt, thin, align=left, fill=color1!18},
parent/.style = {basic, text width=4.0cm, rounded corners=1.5pt, thin, align=left, fill=color1!7},
child/.style = {basic, text width=4.2cm, rounded corners=1.5pt, thin, align=left, fill=color1!1.5},
}
\forestset{
  main'/.style={
    l sep=5mm,
    anchor=west,
  },
  root'/.style={root,
    anchor=west,
    edge path={
     \noexpand\path[\forestoption{edge}]
     ($(!u.east)!.0!(!.west)$) ++(0.5,0) |- (!.west);
    },
  },
  parent'/.style={parent,
    anchor=west,
    calign=child edge,
    l sep=0.4cm
  },
}
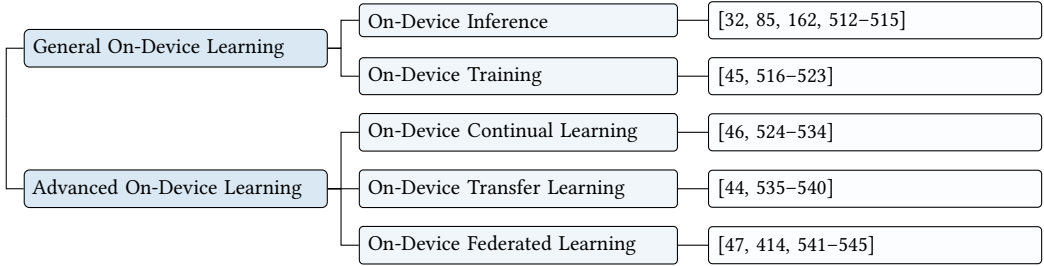
\begin{figure*}[t]
    \hspace{-0.85cm} 
    \centering
\begin{forest}
    for tree={
        forked edges,
        text centered,
        grow'=east,
        reversed=true,
        font=\footnotesize,
        rectangle, /tikz/align=left, anchor=base west, tier/.pgfmath=level(),
        rounded corners,
    }
    [, main'
    [General On-Device Learning, root'
        [On-Device Inference, parent'
                [
                    \cite{banbury2021micronets, lin2020mcunet, lin2021memory, xu2022etinynet, howard2017mobilenets, sandler2018mobilenetv2, howard2019searching}
                    , child
                ]
        ]
        [On-Device Training, parent'
                [
                    \cite{chen2016training, feng2021optimal, ding2022sketch, ye2020accelerating, yang2023efficient, liu2018dynamic, gupta2015deep, zhou2021octo, lin2022device}
                    , child
                ]
        ]
    ]
    [Advanced On-Device Learning, root'
        [On-Device Continual Learning, parent'
                [
                    \cite{van2019three, ravaglia2021tinyml, hayes2022online, pellegrini2021continual, demosthenous2021continual, xiao2022continual, kwon2021exploring, diwan2023continual, dequino2022vit, shin2020pragmatic, wang2022efficient, sundaramoorthy2018harnet}
                    , child
                ]
        ]
        [On-Device Transfer Learning, parent'
                [
                    \cite{cui2018large, kornblith2019better, mudrakarta2018k, frankle2020training, kanavati2021partial, yazdanpanah2022revisiting, cai2020tinytl}
                    , child
                ]
        ]
        [On-Device Federated Learning, parent'
                [
                    \cite{mcmahan2017communication, caldas2018expanding, lin2017deep, jacob2018quantization, zhu2021delayed, yang2022partial, qiu2022zerofl}
                    , child
                ]
        ]
    ]
    ]
\end{forest}
\caption{Comparisons of efficient on-device learning techniques that have been discussed in Section~\ref{sec:efficient-on-device-learning-for-embedded-computing-systems}.}
\vspace{-5pt}
\label{fig:comparisons-on-device-learning}
\end{figure*}

On-device learning consists of two branches, including on-device inference and training. Specifically, on-device inference refers to the process of deploying efficient pre-trained networks on local hardware devices, which allows local hardware devices to run various intelligent inference tasks, such as image classification and object detection. There have been several representative techniques \cite{lin2020mcunet, lin2021memory} to enable efficient on-device inference, which focus on either designing computation-efficient networks with less redundancy or compressing computation-intensive networks to reduce the computational complexity in order to accommodate the limited on-device computational resources. Note that this paper has discussed popular techniques for efficient on-device inference, such as efficient manual/automated network design and efficient network compression, and the readers may refer to Section~\ref{sec:manual-network-design-for-embedded-computing-systems}, Section~\ref{sec:automated-network-design-for-embedded-computing-systems}, and Section~\ref{sec:network-compression-for-embedded-computing-systems} for more details.

On the other hand, on-device training refers to the capability of local hardware to perform training tasks directly on local hardware itself without the need for remote servers \cite{cai2020tinytl}. Unlike on-device inference where the deployed network always remains static, on-device training may further enhance the deployed network over time, which allows to adapt the deployed network to new data collected from local sensors so as to achieve better accuracy. Thanks to its strong capability of protecting data privacy and ensuring data security, on-device training has become increasingly popular over the past few years in order to achieve secured embedded intelligence \cite{zhu2022device}. To this end, we, in this section, systematically discuss recent state-of-the-art on-device learning techniques (especially on-device training), including general on-device learning in Section~\ref{sec:general-on-device-learning}, on-device continual learning in Section~\ref{sec:on-device-continual-learning}, on-device transfer learning in Section~\ref{sec:on-device-transfer-learning}, and on-device federated learning in Section~\ref{sec:on-device-federated-learning}, since these on-device learning techniques feature different learning algorithms to enhance the on-device learning performance. For better understanding, we also summarize these on-device learning methods in Fig.~\ref{fig:comparisons-on-device-learning}. Note that these on-device learning techniques can typically generalize across different networks (e.g., convolutional networks and transformers). For example, we can leverage on-device federated learning to optimize both convolutional networks and transformers on multiple local hardware devices.

\subsection{General On-Device Learning}
\label{sec:general-on-device-learning}

In this section, we introduce recent state-of-the-art works about general on-device learning techniques, including efficient on-device inference and efficient on-device training.

\textbf{Efficient On-Device Inference.} 
To enable efficient on-device inference, one straightforward approach is to design tiny networks with less redundancy in order to accommodate the limited on-device computational resources. To this end, a plethora of representative tiny networks \cite{banbury2021micronets, lin2020mcunet, lin2021memory, xu2022etinynet} have been recently proposed, including MicroNets \cite{banbury2021micronets}, MCUNets \cite{lin2020mcunet, lin2021memory}, and EtinyNet \cite{xu2022etinynet}. Among them, MCUNetV1 \cite{lin2020mcunet}, as one of the early tiny networks, proposes to jointly design the lightweight tiny network using TinyNAS and the lightweight inference engine using TinyEngine, enabling ImageNet-scale inference on microcontrollers. Furthermore, MCUNetV2 \cite{lin2021memory} introduces an efficient patch-based inference pipeline to trim down the on-device memory consumption for better on-device inference since the memory consumption is the key bottleneck of on-device inference. However, different from training large networks, training tiny networks poses significant challenges as demonstrated in \cite{cai2021network}. The rationale here is that existing regularization techniques (e.g., data augmentation and dropout), despite being able to benefit the training process of large networks, may degrade the training performance of tiny networks \cite{cai2021network}. To tackle this issue, \cite{cai2021network} proposes to augment the tiny network itself rather than augmenting the input data, which shows promising accuracy improvement over the standard training scheme.

\begin{figure}[t]
    \begin{center}
    \includegraphics[width=0.9\columnwidth]{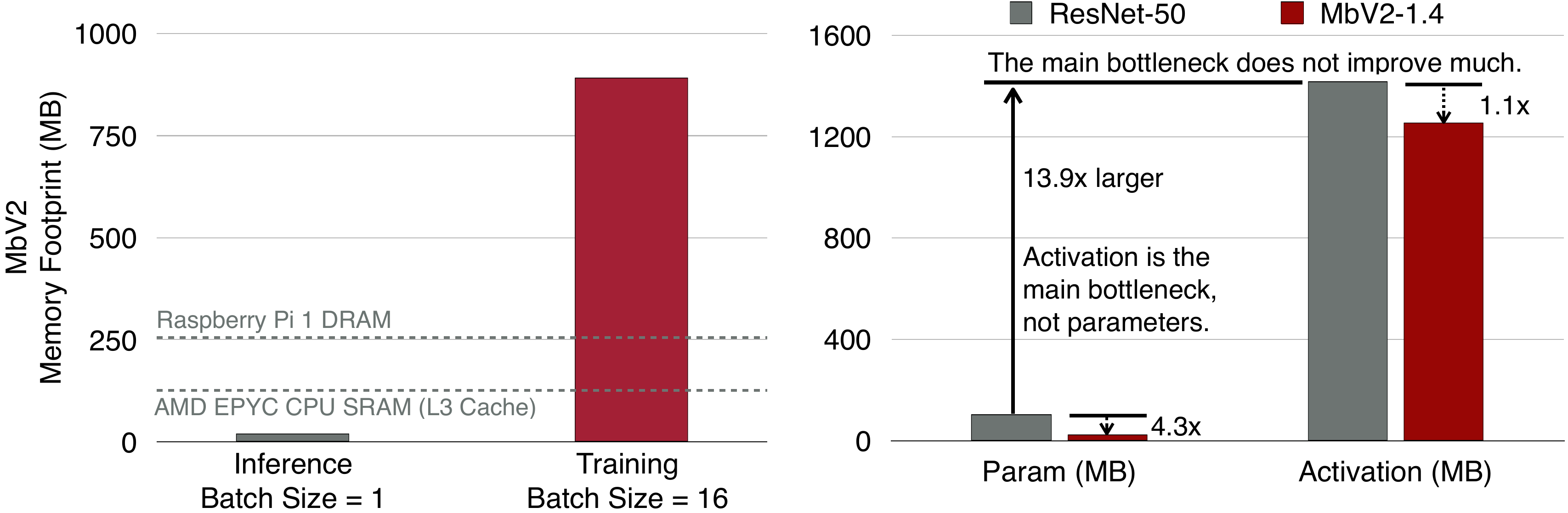}
    \end{center}
    \vspace{-5pt}
    \caption{Comparisons between on-device training and inference in terms of the memory consumption. This reveals that the activation size, instead of the parameter size, is the major bottleneck of on-device training, motivating future research to reduce the activation size for efficient on-device training. \textbf{(figure from \cite{cai2020tinytl})}}
    \vspace{-5pt}
    \label{fig:on-device-training-vs-inference}
\end{figure}

\textbf{Efficient On-Device Training.}
The key difference between on-device training and inference is that on-device training requires to save all the intermediate activations, which are used to optimize parameters using gradient descent during backward propagation. In contrast, on-device inference that only performs forward propagation does not need to save intermediate activations, which can be progressively released to reduce the memory consumption. In sight of this, on-device training suffers from non-negligible memory consumption since the activation size grows with respect to the training batch size and training typically involves a large batch size to accelerate the training process. As a result, intermediate activations arise to be the major bottleneck of on-device training as demonstrated in \cite{cai2020tinytl}. For example, under the batch size of 16, the activation size of ResNet50 \cite{he2016deep} is $\times$13.9 larger than its parameter size as shown in Fig.~\ref{fig:on-device-training-vs-inference}. To alleviate the excessive memory consumption caused by intermediate activations, there have been several representative strategies, including gradient checkpointing, activation gradient pruning, and low-bit training:
\begin{itemize}
    \item[(1)] {
    \textbf{Gradient Checkpointing.}
    Gradient checkpointing is a simple yet effective memory optimization technique, which seeks to reduce the training memory consumption at the cost of increased training time \cite{chen2016training}. To this end, gradient checkpointing reserves a minimal set of intermediate activations during forward propagation, which are then utilized to re-compute the remaining intermediate activations during backward propagation. As shown in \cite{chen2016training}, gradient checkpointing has the potential to significantly reduce the training memory consumption from $O(n)$ to $O(\sqrt{n})$, where $n$ is the number of network layers. More importantly, gradient checkpointing does not degrade the training accuracy since the training behaviors remain to be the same as the standard training scheme. Furthermore, several subsequent gradient checkpointing works generalize gradient checkpointing to allow arbitrary computation graphs \cite{feng2021optimal} and train graph neural networks \cite{ding2022sketch}.
    }
    \item[(2)] {
    \textbf{Activation Gradient Pruning.}
    Activation gradient pruning removes less important intermediate activation gradients to optimize the training memory consumption \cite{ye2020accelerating}. This relies on an empirical observation that most of the intermediate activation gradients during backward propagation are very close to zero and thus have minimal impact on gradient descent \cite{ye2020accelerating}. Therefore, pruning these very small activation gradients can effectively reduce the training memory consumption at the cost of minimal accuracy loss, which also accelerates the training process. Similar to \cite{ye2020accelerating}, \cite{yang2023efficient} proposes an efficient gradient filtering scheme, which filters similar activation gradients during backward propagation and only reserves those with unique elements to reduce the number of elements in the activation gradient maps. Apart from these, another popular approach is to build dynamic sparse computation graphs to eliminate intermediate activations in an input-dependent manner, which can also reduce the training memory consumption \cite{liu2018dynamic}.
    }
    \item[(3)] {
    \textbf{Low-Bit Training.}
    Low-bit training refers to training the given network with low-bit weights (e.g., 8 and 16 bits) rather than full-precision 32-bit weights, which has the potential to significantly reduce the training memory consumption by $\times$32 times \cite{cai2022enable}. The rationale here is that low-bit training can reduce the memory consumption for both network weights and intermediate activations. Specifically, \cite{gupta2015deep}, as an early exploration, proposes to train the given network with 16-bit weights under stochastic rounding, which leads to $\times$2 less training memory consumption than the standard full-precision 32-bit training counterpart, and more importantly, maintains comparable training accuracy. Besides, \cite{zhou2021octo} introduces an efficient INT8 training pipeline, consisting of loss-aware compensation and backward quantization, to enable tiny on-device training, thanks to the well-optimized INT8 quantization on mainstream hardware. Furthermore, \cite{lin2022device} proposes to optimize real-quantized graphs, in which an effective memory-efficient sparse update scheme and tiny training engine are integrated to achieve on-device training under 256\,KB memory. It is worth noting that low-bit training is similar to network quantization as discussed in Section~\ref{sec:network-quantization} since both leverage quantized weights to trim down the network complexity, which allows to generalize recent advanced quantization techniques to benefit low-bit on-device training.    
    }
\end{itemize}
We note that the aforementioned strategies can be also combined to further reduce the overall memory consumption during training and accelerate the training process. For example, we can easily combine gradient checkpointing and low-bit training to further reduce the training memory consumption from $O(\sqrt{n})$ to $O(\sqrt{n}/32)$, where $n$ is the number of network layers.

\subsection{On-Device Continual Learning}
\label{sec:on-device-continual-learning}

On-device continual learning, also known as on-device lifelong learning or incremental learning, is an advanced learning paradigm, which allows the deployed network to continuously learn from the newly collected data to further push forward the attainable accuracy \cite{van2019three, wang2022efficient}. This is particularly favored in real-world embedded scenarios, especially those with rich local sensors, where embedded devices can continue to collect new data through local sensors over time \cite{cai2020tinytl, lin2022device}. The newly collected data is then used to train the deployed network to unlock better performance over time. In other words, we are allowed to utilize the newly collected data to train or fine-tune the deployed network on target hardware itself, which typically leads to better accuracy on target task. In the meantime, on-device continual learning performs local training and does not need to send back the newly collected data to remote servers, which also protects data privacy and ensures data security. However, on-device continual learning, despite being able to deliver significant benefits, suffers from the catastrophic forgetting issue, which is the tendency to forget the previously learned knowledge when adapting to newly collected data \cite{van2019three}. The rationale is that on-device continual learning must adjust the pre-trained network weights in order to adapt to the newly collected data, which deteriorates the previously learned knowledge accordingly.

To alleviate the catastrophic forgetting issue, a plethora of state-of-the-art on-device continual learning works have been recently established \cite{van2019three, ravaglia2021tinyml, hayes2022online, pellegrini2021continual, demosthenous2021continual, xiao2022continual, kwon2021exploring, diwan2023continual, dequino2022vit, shin2020pragmatic, wang2022efficient, sundaramoorthy2018harnet}, which seeks to stabilize the on-device continual learning process and further push forward the attainable accuracy on target task. Among them, \cite{van2019three}, as an early exploration, investigates three common continual learning scenarios and demonstrates that it is frustratingly hard to evaluate different continual learning approaches, based on which \cite{van2019three} establishes several evaluation protocols to compare different continual learning approaches. It is worth noting that \cite{van2019three} itself does not support on-device continual learning but we can easily integrate recent advances from the lens of on-device training (see Section~\ref{sec:general-on-device-learning}) into \cite{van2019three} to further enable efficient on-device continual learning. Several subsequent on-device continual learning works \cite{ravaglia2021tinyml, hayes2022online, pellegrini2021continual, demosthenous2021continual} explore on-device continual learning on resource-constrained embedded computing systems, which have since demonstrated promising accuracy improvement. In parallel, \cite{xiao2022continual, kwon2021exploring, diwan2023continual} attempts to generalize on-device continual learning to benefit languages tasks in real-world embedded scenarios, such as environmental sound classification \cite{xiao2022continual} and automatic speech recognition \cite{diwan2023continual}. Inspired by the tremendous success of vision transformers \cite{dosovitskiy2020image}, \cite{dequino2022vit} also investigates the efficacy of on-device embedded continual learning to continuously improve the accuracy of mainstream vision transformers. Furthermore, \cite{shin2020pragmatic, wang2022efficient, sundaramoorthy2018harnet} delve deeper into on-device continual learning, which focus more on the training pipeline and introduce several on-device training enhancements to maximize the accuracy improvement, such as selective weight updates \cite{shin2020pragmatic}, weight freezing \cite{wang2022efficient}, and deep network ensembles \cite{sundaramoorthy2018harnet}.

\subsection{On-Device Transfer Learning}
\label{sec:on-device-transfer-learning}

\begin{figure}[t]
    \begin{center}
    \includegraphics[width=1.0\columnwidth]{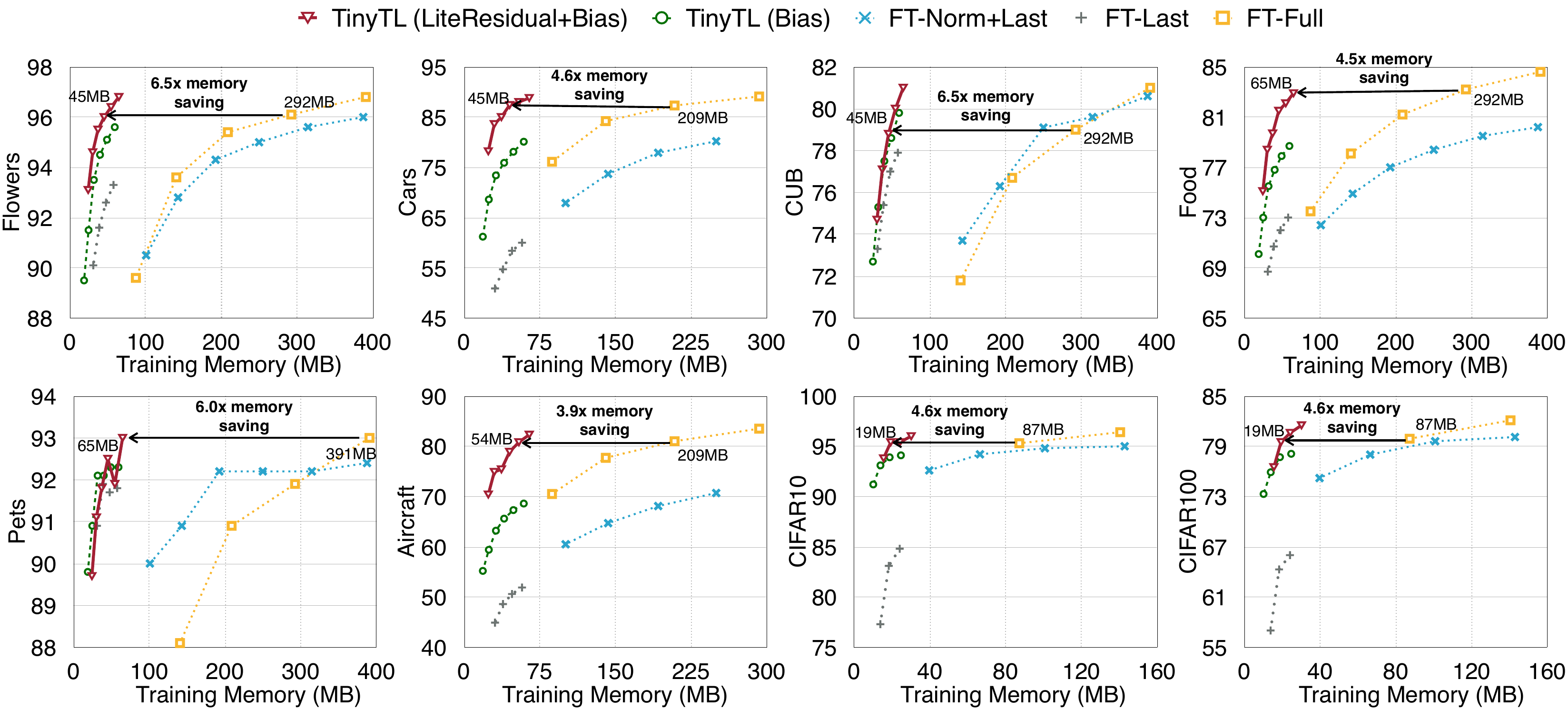}
    \end{center}
    \vspace{-5pt}
    \caption{Comparisons of various on-device transfer learning methods including TinyTL \cite{cai2020tinytl}, FT-Norm+Last \cite{mudrakarta2018k}, FT-Last \cite{kornblith2019better}, and FT-Full \cite{chatfield2014return}. Among them, TinyTL freezes the weights and only optimizes the bias modules. FT-Norm+Last fine-tunes the normalization layers and the last linear layer, whereas FT-Last fine-tunes the last linear layer and FT-Full fine-tunes the full network. \textbf{(figure from \cite{cai2020tinytl})}}
    \vspace{-5pt}
    \label{fig:on-device-transfer-learning-tinytl}
\end{figure}

As demonstrated in \cite{cai2020tinytl}, it is often different to directly train DNNs from scratch in real-world embedded scenarios, where the collected data samples are far limited. To tackle this issue, an effective alternative is on-device transfer learning, which instead fine-tunes pre-trained networks on large-scale datasets. The rationale here is that DNNs pre-trained on large-scale datasets (e.g., ImageNet \cite{deng2009imagenet}) can serve as powerful feature extractors for further transfer learning, which only fine-tunes several layers (e.g., batch normalization layers and the last layer) whereas other layers are typically frozen. Different from previous on-device learning practices as discussed in Section~\ref{sec:general-on-device-learning} and Section~\ref{sec:on-device-continual-learning}, on-device transfer learning does not require to store memory-intensive intermediate activations, and as a result, maintains significant efficiency in terms of training memory consumption as illustrated in Fig.~\ref{fig:on-device-transfer-learning-tinytl} \cite{cai2020tinytl}. Despite the promising memory efficiency, on-device transfer learning is quite challenging, which may result in poor accuracy, especially on those datasets whose data distribution is far from ImageNet \cite{cai2022enable}.

To overcome such limitations, early transfer learning practices \cite{cui2018large, kornblith2019better} propose to fine-tune all the network layers, which indeed achieve better accuracy but lead to considerable memory consumption due to memory-intensive intermediate activations. To avoid this, several subsequent transfer learning works \cite{mudrakarta2018k, frankle2020training, kanavati2021partial, yazdanpanah2022revisiting} demonstrate that it is often not necessary to fine-tune all the network layers, which indicate that fine-tuning batch normalization layers can also achieve strong accuracy on target task. This fine-tuning paradigm has the potential to significantly reduce the number of trainable parameters during the transfer learning process. In sight of this, \cite{mudrakarta2018k, frankle2020training, kanavati2021partial, yazdanpanah2022revisiting} propose to only optimize learnable parameters in batch normalization layers (see $\gamma$ and $\beta$ in Eq~(\ref{eq:batch-normalization})), whereas other learnable parameters are frozen during the transfer learning process. For example, \cite{mudrakarta2018k} leverages batch normalization layers as scale-and-bias patches and then trains the patched parameters, optionally also the last layer, whereas the remaining parameters are left unchanged. Furthermore, \cite{frankle2020training} reveals that, for those networks with sufficient depth, training only $\gamma$ and $\beta$ can reach surprisingly strong accuracy, which demonstrates the expressive power of the learnable parameters in batch normalization layers. However, fewer trainable parameters cannot directly translate to superior training memory efficiency as shown in Fig.~\ref{fig:on-device-training-vs-inference}, which may still involve a large amount of memory consumption (e.g., 326\,MB under the training batch size of 8) to store memory-intensive intermediate activations of batch normalization layers \cite{cai2020tinytl}.

To further alleviate the prohibitive training memory consumption, \cite{cai2020tinytl} introduces a simple yet effective transfer learning solution, which exhibits significant training memory efficiency. Specifically, \cite{cai2020tinytl} relies on an empirical observation that intermediate activations are only required to update network weights, whereas updating network biases does not involve intermediate activations. This observation also reveals that the training memory bottleneck comes from updating network weights rather than biases. In sight of this, \cite{cai2020tinytl} proposes to freeze network weights and only update network biases. However, freezing network weights and only updating network biases may lead to significant accuracy loss. To compensate for such accuracy loss due to freezing network weights, \cite{cai2020tinytl} introduces an effective lite residual learning scheme, which leverages generalized memory-efficient bias modules to refine memory-intensive intermediate activations. In particular, the lite residual learning scheme can improve the attainable accuracy on target task, and more importantly, at the cost of negligible memory overheads. Finally, \cite{cai2020tinytl} reduces the training memory consumption from more than 250\,MB to only 16\,MB, making it possible to explore in-memory computing infrastructures to perform memory-efficient transfer learning. Note that we can easily integrate recent advances from the lens of on-device training (see Section~\ref{sec:general-on-device-learning}) into the aforementioned transfer learning works towards boosted on-device transfer learning performance.

\subsection{On-Device Federated Learning}
\label{sec:on-device-federated-learning}

On-device federated learning is an advanced decentralized learning paradigm, which enables efficient training on a large corpus of decentralized data residing on local client devices like mobile phones and allows multiple local client devices to jointly train the given network without explicitly sharing their raw data \cite{mcmahan2017communication, huai2022collate}. In practice, on-device federated learning has the potential to significantly accelerate the training process when the number of client devices evolves. Besides, on-device federated learning is also one instance of the more general approach of \textit{"bringing the neural network to the data"} rather than \textit{"bringing the data to the neural network"}, and as a result, addresses the fundamental problems of data privacy, security, and ownership \cite{bonawitz2019towards}. This is particularly favored in real-world embedded scenarios, where embedded devices themselves can continue to collect new data through local sensors. Thanks to the above practical benefits, on-device federated learning has garnered increasing attention from both academia and industry \cite{qiu2022zerofl}. In the past decade, on-device federated learning has been utilized to empower a plethora of real-world intelligent applications, such as mobile keyboard content suggestions \cite{hard2018federated}, medical image analysis \cite{adnan2022federated}, and smart healthcare infrastructures \cite{antunes2022federated}. As demonstrated in \cite{mcmahan2017communication}, standard on-device federated learning practices typically consist of the following five iterative steps:
\begin{itemize}
    \item[(1)] {
    \textbf{Initialization.} 
    On-device federated learning begins with the randomly initialized network, namely the global model, which is shared among local client devices. At the early learning stage, the global model is sent to all the local client devices from the centralized server, in which each local client device receives the same copy of the global model.
    }
    \item[(2)] {
    \textbf{Local Training.}
    Once local client devices receive the global model, they start to perform local on-device training, which treats the global model as the local model and then trains the local model using the locally collected data. Note that the locally collected data only resides on the local client device itself and is not shared among other client devices.
    }
    \item[(3)] {
    \textbf{Model Update.}
    After local on-device training terminates, each local client device requires to generate the respective model update scheme, which should essentially reflect what the local client device has learned from the locally collected data. These model update schemes, instead of the locally collected data, are then sent back to the centralized server for further aggregation, which effectively eliminates data leakage and protects data privacy.
    }
    \item[(4)] {
    \textbf{Aggregation.}
    The centralized server receives model update schemes from all the local client devices, after which the centralized server aggregates the received model update schemes to further produce an improved global model.
    }
    \item[(5)] {
    \textbf{Distribution.}
    The centralized server then distributes the improved global model to all the local client devices, and the above steps repeat until convergence.
    }
\end{itemize}
Despite being able to deliver superior learning performance across various real-world embedded tasks, on-device federated learning also suffers from critical limitations, especially from the perspective of data transmission \cite{bonawitz2019towards}, posing significant challenges to generalize on-device federated learning to benefit real-world intelligent embedded applications. Different from the centralized server that is equipped with high-end network infrastructures, local client devices in real-world embedded scenarios are often low-end with less-capable network infrastructures. In such a case, it may be time-consuming to (1) distribute the global model from the centralized server to local client devices and (2) send back model update schemes from local client devices to the centralized server for further aggregation. To overcome such limitations, a plethora of advanced federated learning techniques have been recently established to accommodate the limited data bandwidth of local client devices \cite{mcmahan2017communication, caldas2018expanding, lin2017deep, jacob2018quantization, zhu2021delayed, yang2022partial, qiu2022zerofl}, which primarily focus on reducing the total data bits transferred between the remote centralized server and local client devices, such as federated averaging \cite{mcmahan2017communication}, gradient compression \cite{caldas2018expanding, lin2017deep}, quantization \cite{jacob2018quantization}, delayed gradient averaging \cite{zhu2021delayed}, partial variable training \cite{yang2022partial}, and local training sparsity \cite{qiu2022zerofl}. Note that we can easily integrate recent advances from the lens of general on-device training techniques (see Section~\ref{sec:general-on-device-learning}) into the aforementioned federated learning works to further enhance on-device federated learning.

\subsection{Future Envision}
In this section, we further envision several promising future trends and possible directions in the field of on-device learning, which are summarized as follows:
\begin{itemize}
    \item[(1)] {
    \textbf{Offline On-Device Federated Learning.}
    As discussed in Section~\ref{sec:on-device-transfer-learning}, on-device federated learning highly relies on the centralized server for updating local models, which requires stable internet requirements for data movements between local devices and remote server, and as a result, may suffer from inferior on-device learning efficiency due to the communication overheads between local devices and remote server, especially when the internet connectivity is limited or unavailable. Therefore, one promising future trend is offline on-device federated learning, which excludes the remote centralized server and exploits local devices themselves to perform learning tasks. In particular, offline on-device federated learning has the potential to significantly boost the on-device learning efficiency.
    }
    \item[(2)] {
    \textbf{Personalized On-Device Learning.}
    As discussed in Section~\ref{sec:general-on-device-learning}, on-device learning exhibits strong local personalization, which distinguishes itself from the global training counterpart. In practice, personalized on-device learning can bring two-fold benefits. On the one hand, personalized on-device learning allows local devices to directly learn from local users to provide user-tailored AI solutions, which can thus protect the data privacy since the collected data do not need to be transferred to the remote cloud. On the other hand, personalized on-device learning can achieve better learning accuracy since it can continue to collect rich new personalized training data from local users. Therefore, future research should leverage this unique capability to further provide more highly personalized on-device learning solutions, where local devices can actively and quickly adapt themselves to users' diverse needs so as to deliver user-tailored services, such as personalized voice assistants.
    }
    \item[(3)] {
    \textbf{Robust On-Device Learning.}
    On-device learning, despite being able to achieve promising success, still suffers from critical limitations, such as poor adversarial robustness \cite{huang2021robustness}. This is particularly important in real-world embedded computing systems, such as embedded visual sensing \cite{song2021deepmtd, song2019moving}, where the environments may dynamically change over time. This further makes local on-device learning more vulnerable to adversarial attacks, especially those unseen adversarial attacks, which may significantly degrade the on-device learning performance even when encountering simple adversarial attacks \cite{huang2021robustness}. To overcome such limitations, future research should focus on developing robust on-device learning techniques featuring novel adversarial training algorithms, such as \cite{ma2025dynamic, ma2025mobile, wu2025demo}, which can achieve competitive on-device learning performance while also maintaining superior adversarial robustness against well-engineered adversarial attacks or even unseen adversarial attacks.
    }
    \item[(4)] {
    \textbf{Efficient On-Device Learning Ecosystems.}
    As discussed in Section~\ref{sec:general-on-device-learning}, on-device learning has gained increasing popularity from both academia and industry, thanks to its strong capability to ensure data privacy and security. In sight of this, future research should also develop efficient on-device learning ecosystems, including software and hardware frameworks, to further support the development, deployment, and management of on-device learning applications, making it easier for developers to create and optimize models for various on-device learning purposes. For example, \cite{lin2022device}, as one of the most representative on-device learning methods, leverages quantization to trim down the training memory consumption. However, mainstream embedded computing systems do not support low-bit training, making it difficult to benefit mainstream embedded computing systems. 
    }
\end{itemize}

\section{Efficient Large Language Models for Embedded Computing Systems}
\label{sec:large-language-models-for-embedded-computing-systems}

In the past few years, large language models (LLMs), such as GPT-3 \cite{brown2020language} and GPT-4 \cite{openai2023gpt4}, have achieved impressive success across various real-world language processing tasks \cite{bai2024beyond}. However, the strong learning capability of LLMs also comes at the cost of excessive computational complexity. For example, OpenAI's GPT-3 \cite{brown2020language}, as one of the most representative LLMs, consists of 175 billion parameters. More recently, LLMs continue to evolve with ever-increasing model sizes in order to achieve state-of-the-art performance \cite{chowdhery2023palm, le2023bloom}. These further make it challenging to deploy LLMs on modern embedded computing systems. To this end, we, in this section, first introduce the preliminaries on LLMs and then discuss recent state-of-the-art advances from the perspective of efficient LLMs, including efficient LLM architecture design in Section~\ref{sec:efficient-llm-architecture-design}, efficient LLM compression techniques in Section~\ref{sec:efficient-llm-compression}, and efficient LLM system design in Section~\ref{sec:efficient-llm-system-design}. We also summarize the above state-of-the-art advances about efficient LLMs in Fig.~\ref{fig:comparisons-llms}. Finally, in Section~\ref{sec:llm-future-envision}, we envision several promising future directions in the field of efficient LLMs.

\subsection{Preliminaries on LLMs}
\label{sec:preliminaries-on-llms}

Large language models (LLMs) are emerging machine learning models, which are dedicated to understanding, generating, and interacting with human language through leveraging extensive textual data. In practice, LLMs are typically built upon transformer with both encoder and decoder \cite{vaswani2017attention} and heavily rely on self-attention mechanisms to measure the significance of different words in the given sentence, regardless of their positional relationships. Thanks to their strong capability to interpret rich information, LLMs can exhibit remarkable performance across a wide range of language processing tasks, such as text summarization, translation, question answering, and conversational response generation. As discussed in \cite{bai2024beyond}, recent state-of-the-art LLMs can be divided into three main categories according to their inherent architectures, including encoder-only architecture, decoder-only architecture, and encoder-decoder architecture as follows:
\begin{enumerate}
    \item [(1)]
    \textbf{Encoder-Only Language Models.}
    Encoder-only language models typically focus on transforming the given input text into continuous representations, which can capture and reflect the context of the given input text. These encoder-only language models are usually used for real-world language processing tasks that require understanding or embedding of the given input text, such as sentence classification, named entity recognition, and extractive question answering, where the output does not need to be sequential or generated texts. For example, BERT \cite{devlin2018bert} is one of the most representative encoder-only language models that features masked language modeling during training, which enables the model itself to understand the context from both directions (i.e., left and right context).
    \item [(2)]
    \textbf{Decoder-Only Language Models.}
    Decoder-only language models typically focus on generating texts based on the given input text, which can interpret the context of the given input text. These decoder-only language models are usually used for real-world language processing tasks where generating texts is required, such as text generation and language modeling. For example, GPT-3 \cite{brown2020language} is one of the most representative decoder-only language models that features advanced auto-regressive training, which can learn to accurately predict the next word in sequence from all the previous words.
    \item [(3)]
    \textbf{Encoder-Decoder Language Models.}
    Encoder-decoder language models, also known as sequence-to-sequence (seq2seq) models, typically consist of two parts, including (1) the encoder to process the given input text and encode it into feature representations and (2) the decoder to explore the above feature representations to generate an output sequence. The encoder-decoder architecture is versatile and suitable for real-world language processing tasks that require to transform the given input text into different formats, such as language translation, summarization, and dialogue systems. For example, T5 \cite{raffel2020exploring} is one of the most representative encoder-decoder language models, which formulates the given task as a text-to-text transformation problem and converts the given input text to the target output text. In parallel, BART \cite{lewis2019bart} is particularly effective in generative and comprehensive tasks, thanks to its bidirectional encoder and auto-regressive decoder.
\end{enumerate}

\tikzset{
basic/.style = {draw, font=\footnotesize, rectangle},
root/.style = {basic, font=\footnotesize,  text width=3.4cm, rounded corners=2pt, thin, align=left, fill=color1!18},
parent/.style = {basic, text width=3.4cm, rounded corners=1.5pt, thin, align=left, fill=color1!7},
child/.style = {basic, text width=5.2cm, rounded corners=1.5pt, thin, align=left, fill=color1!1.5},
}
\forestset{
  main'/.style={
    l sep=5mm,
    anchor=west,
  },
  root'/.style={root,
    anchor=west,
    edge path={
     \noexpand\path[\forestoption{edge}]
     ($(!u.east)!.0!(!.west)$) ++(0.5,0) |- (!.west);
    },
  },
  parent'/.style={parent,
    anchor=west,
    calign=child edge,
    l sep=0.4cm
  },
}
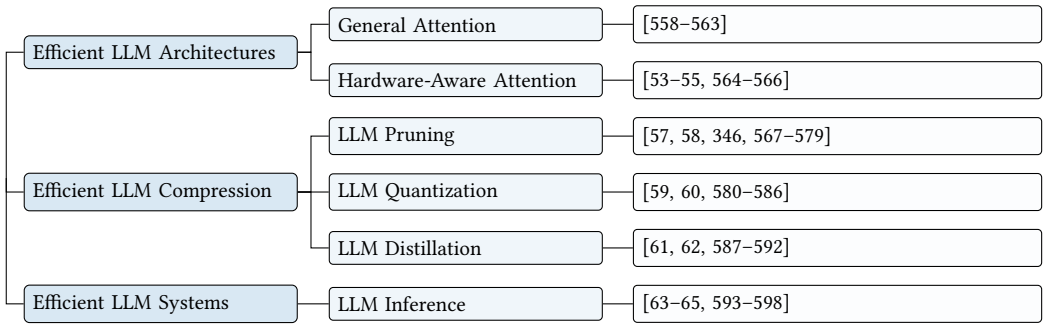
\begin{figure*}[t]
    \hspace{-0.85cm} 
    \centering
\begin{forest}
    for tree={
        forked edges,
        text centered,
        grow'=east,
        reversed=true,
        font=\footnotesize,
        rectangle, /tikz/align=left, anchor=base west, tier/.pgfmath=level(),
        rounded corners,
    }
    [, main'
    [Efficient LLM Architectures, root'
        [General Attention, parent'
                [
                    \cite{vyas2020fast, xiong2021nystromformer, zandieh2023kdeformer, ma2022mega, alberti2023sumformer, gupta2023flurka}
                    , child
                ]
        ]
        [Hardware-Aware Attention, parent'
                [
                    \cite{dao2022flashattention, dao2023flashattention, kwon2023efficient, ham20203, ham2021elsa, yang2023gated}
                    , child
                ]
        ]
    ]
    [Efficient LLM Compression, root'
        [LLM Pruning, parent'
                [
                    \cite{sun2023simple, frantar2023sparsegpt, ashkboos2024slicegpt, mirzadeh2023relu, li2023sparse, thangarasa2023spdf, ansell2024scaling, kurtic2023sparse, ma2023llm, an2024fluctuation, kurtic2024ziplm, chen2023lorashear, xia2023sheared, chen2024compressing, men2024shortgpt, kim2024shortened}
                    , child
                ]
        ]
        [LLM Quantization, parent'
                [
                    \cite{xiao2023smoothquant, lin2023awq, dettmers2023spqr, wei2023outlier, lee2023owq, chee2024quip, shao2023omniquant, li2024evaluating, guo2023olive}
                    , child
                ]
        ]
        [LLM Distillation, parent'
                [
                    \cite{wang2022self, peng2023instruction, wu2023lamini, jiang2023lion, gu2023minillm, liang2023less, agarwal2023gkd, kim2024token}
                    , child
                ]
        ]
    ]
    [Efficient LLM Systems, root'
        [LLM Inference, parent'
                [
                    \cite{sheng2023flexgen, wang2023tabi, aminabadi2022deepspeed, wu2023fast, borzunov2022petals, jin2024s, patel2023splitwise, zhong2024distserve, du2024liger}
                    , child
                ]
        ]
    ]
    ]
\end{forest}
\caption{Overview of efficient LLM architectures, LLM compression techniques, and LLM systems in Section~\ref{sec:large-language-models-for-embedded-computing-systems}.}
\vspace{-5pt}
\label{fig:comparisons-llms}
\end{figure*}

\subsection{Efficient LLM Architectures}
\label{sec:efficient-llm-architecture-design}

As discussed in Section~\ref{sec:preliminaries-on-llms}, recent LLMs are typically built upon transformer \cite{vaswani2017attention} and heavily rely on self-attention mechanisms to interpret the significance of different words in the given sentence, regardless of their positional relationships. However, self-attention mechanisms, despite their strong capability for language processing, also introduce considerable computational complexity. As pointed out in \cite{keles2023computational}, the quadratic time and memory complexity of self-attention mechanisms may significantly slow down the pre-training, inference, and fine-tuning stages of LLMs. To optimize the prohibitive computational complexity of LLMs, recent state-of-the-art efficient LLMs often focus on exploring computation-efficient self-attention mechanisms.

\textbf{General Efficient Attention.}
Some recent works \cite{vyas2020fast, xiong2021nystromformer, zandieh2023kdeformer, ma2022mega, alberti2023sumformer, gupta2023flurka} focus on exploring computation-efficient attention mechanisms to optimize the quadratic computational complexity of vanilla self-attention \cite{vaswani2017attention}. Among them, \cite{vyas2020fast} introduces clustered attention, which groups different queries into clusters and computes the attention just for the centroids rather than computing the attention for every query. To further improve the above approximation, \cite{vyas2020fast} also employs the computed clusters to identify the keys with the highest attention per query and computes the exact key-query dot products. \cite{xiong2021nystromformer} draws insights from the Nystrom method and proposes to approximate the standard self-attention mechanism in linear complexity, which thus can enable applications to longer sequences with even thousands of tokens. \cite{zandieh2023kdeformer} demonstrates that the complexity bottleneck of self-attention mainly comes from the computation of partition functions in the denominator of the softmax function and the multiplication of the softmax matrix with the matrix of values. To this end, \cite{zandieh2023kdeformer} features an efficient kernel density estimation (KDE) solver to resolve the above complexity bottleneck via sub-sampling based fast matrix products, which can approximate the attention in sub-quadratic time with provable spectral norm bounds. \cite{ma2022mega} introduces an efficient single-head gated attention mechanism with exponential moving average to incorporate inductive bias of position-aware local dependencies into the position-agnostic attention mechanism, which can exhibit linear time and space complexity while causing minimal performance loss. \cite{alberti2023sumformer} introduces an efficient universal approximation for self-attention, which can exhibit linear time and space complexity and also reveal the theoretical insights behind existing efficient transformers (e.g., Linformer \cite{wang2020linformer}) with linear time and space complexity. \cite{gupta2023flurka} introduces an efficient attention approximation mechanism featuring fused low-rank kernel approximation, which can provide sizable runtime performance gains and is also of high quality.

\textbf{Hardware-Aware Efficient Attention.}
In parallel to the above efficient attention approximation works, some recent works \cite{dao2022flashattention, dao2023flashattention, kwon2023efficient, ham20203, ham2021elsa, yang2023gated} focus on exploring efficient hardware-aware attention mechanisms, which can exhibit considerable efficiency improvement on modern hardware systems. Among them, FlashAttention \cite{dao2022flashattention} features an efficient IO-aware exact attention algorithm, which explores tiling to reduce the total number of memory reads and writes between GPU high bandwidth memory and GPU on-chip SRAM. However, FlashAttention is still not nearly as fast as optimized matrix-multiply (GEMM) operations, which is mainly due to the sub-optimal work partitioning between different thread blocks and warps on GPUs. To tackle this issue, FlashAttention-2 \cite{dao2023flashattention} further introduces an improved work partitioning scheme, which (1) tweaks the algorithm to reduce the number of non-matmul FLOPs, (2) parallelizes the attention computation workloads across different thread blocks, and (3) distributes the work between warps to reduce communications through shared memory. FLASHLINEARATTENTION \cite{yang2023gated} dives deeper into I/O-awareness and introduces an effective hardware-efficient algorithm for linear attention, which trades off memory movement against parallelizability and can even be faster than FlashAttention-2. PagedAttention \cite{kwon2023efficient} draws insights from the operating system's solution to memory fragmentation and sharing through virtual memory with paging and further divides the request's key-value (KV) cache into different blocks, each of which contains the attention keys and values of a fixed number of tokens. A3 \cite{ham20203} demonstrates that implementing attention mechanisms using matrix-vector multiplication is often sub-optimal and further proposes to accelerate attention mechanisms with joint algorithmic approximation and hardware specialization. Similar to A3, ELSA \cite{ham2021elsa} features an effective approximation scheme to significantly reduce the amount of computation workloads by efficiently filtering out relationships that are unlikely to affect the final output.

\subsection{Efficient LLM Compression}
\label{sec:efficient-llm-compression}

In addition to designing LLMs with efficient architectures, another promising direction is to explore efficient LLM compression techniques to optimize the computational complexity of existing computation-intensive LLMs. With this in mind, we, in this section, further discuss recent state-of-the-art compression techniques for LLMs, including efficient LLM pruning in Section~\ref{sec:efficient-llm-pruning}, efficient LLM quantization in Section~\ref{sec:efficient-llm-quantization}, and efficient LLM distillation in Section~\ref{sec:efficient-llm-distillation}.

\subsubsection{Efficient LLM Pruning}
\label{sec:efficient-llm-pruning}

Pruning is one of the most effective strategies to optimize the computational efficiency of LLMs, which removes the less important parameters of LLMs while incurring minimal accuracy loss. Recent state-of-the-art LLM pruning methods can be divided into two main categories, including non-structured LLM pruning and structured LLM pruning as follows:
\begin{enumerate}
    \item [(1)]
    \textbf{Non-Structured LLM Pruning.}
    Non-structured LLM pruning removes the less important LLM weights/connects, which can yield more aggressive compression ratios than structured pruning while also exhibiting strong accuracy \cite{sun2023simple, frantar2023sparsegpt, ashkboos2024slicegpt, mirzadeh2023relu, li2023sparse, thangarasa2023spdf, ansell2024scaling, kurtic2023sparse}. For example, SparseGPT \cite{frantar2023sparsegpt} shows that LLMs can be pruned to at least 50\% sparsity in one shot without any retraining, and more importantly, at minimal accuracy loss. In parallel, Wanda \cite{sun2023simple} proposes to prune the less important weights with the smallest magnitudes multiplied with the corresponding output activations on a per-output basis. More importantly, both SparseGPT and Wanda can generalize to semi-structured pruning \cite{ashkboos2024slicegpt, li2023sparse} towards better hardware parallelism, which can deliver realistic on-device inference speedups with the support of some existing deep learning libraries (e.g., cuSPARSElt \cite{mishra2021accelerating} and TVM \cite{chen2018tvm}). Furthermore, \cite{mirzadeh2023relu} advocates for reinstating ReLU activation in LLMs and explores sparse patterns in ReLU-based LLMs, which shows that ReLU activation can effectively reduce LLM inference computation overheads up to three times. In practice, non-structured pruning has been widely employed to enhance the pre-training and fine-tuning process of LLMs towards better pre-training and fine-tuning process \cite{thangarasa2023spdf, ansell2024scaling, kurtic2023sparse}.
    \item [(2)]
    \textbf{Structured LLM Pruning.}
    In contrast to non-structured LLM pruning, structured LLM pruning can achieve realistic inference speedups on target hardware, which, however, also suffers from more aggressive accuracy loss than non-structured pruning. To tackle this dilemma, recent state-of-the-art non-structured LLM pruning methods \cite{ma2023llm, an2024fluctuation, kurtic2024ziplm, chen2023lorashear, xia2023sheared, chen2024compressing, men2024shortgpt, kim2024shortened} typically feature an additional fine-tuning stage to further recover the attainable accuracy of the pruned LLM. For example, LLM-Pruner \cite{ma2023llm} employs structural pruning to selectively remove non-critical coupled structures according to their gradient information, which can preserve the majority of the LLM's functionality while optimizing its computational efficiency. Furthermore, LLM-Pruner recovers the performance of the pruned LLM using another state-of-the-art tuning technique (i.e., LoRA \cite{hu2021lora}), which merely takes 3 hours with 50K data. Similar to LLM-Pruner, ZipLM \cite{kurtic2024ziplm} iteratively identifies and removes LLM components with the worst loss-runtime trade-off, which can end up with efficient LLMs and also generalize across various runtime constraints. In addition, LoRAShear \cite{chen2023lorashear} first creates the dependency graphs over LoRA modules, which then proceeds progressive structured pruning on LoRA adaptors and enables inherent knowledge transfer. To further recover the lost information during pruning, LoRAShear also introduces an effective fine-tuning scheme with dynamic data adaptors to narrow down the performance gap between the pruned LLM and the non-pruned LLM. More recently, several LLM layer pruning methods \cite{men2024shortgpt, kim2024shortened, chen2024compressing} demonstrate that LLM's layers are also redundant and thus can be removed to largely enhance the inference efficiency of LLMs at minimal accuracy loss. For example, ShortGPT \cite{men2024shortgpt} and Shorted-LLaMA \cite{kim2024shortened} propose to remove the less important LLM layers according to their layer importance score, whereas LLM-Streamline \cite{chen2024compressing} proposes to replace the less important LLM layers with more lightweight ones.
\end{enumerate}

\subsubsection{Efficient LLM Quantization}
\label{sec:efficient-llm-quantization}

Recent state-of-the-art LLM quantization techniques \cite{xiao2023smoothquant, lin2023awq, dettmers2023spqr, wei2023outlier, lee2023owq, chee2024quip, shao2023omniquant, li2024evaluating, guo2023olive} focus on reducing the weights of LLMs from higher to lower bits (e.g., from 32 bits to 8 bits or even 1 bit), which can substantially enhance the inference efficiency of LLMs at the cost of slight accuracy loss. Among them, SmoothQuant \cite{xiao2023smoothquant} introduces an efficient training-free post-training quantization solution to enable 8-bit weight and 8-bit activation quantization for LLMs. Given that weights are easy to quantize while activations are not, SmoothQuant also smooths the activation outliers by offline mitigating the quantization difficulty from activations to weights with an equivalent mathematical transformation. Similar to SmoothQuant, AWQ \cite{lin2023awq} introduces an efficient hardware-friendly quantization approach for low-bit LLM weight-only quantization, which is built upon an interesting observation that weights are not equally important and reserving only 1\% of salient weights can greatly reduce the quantization error. In light of this, AWQ proposes to search for the optimal per-channel scaling scheme that protects the salient weights by observing the activations rather than weights. SpQR \cite{dettmers2023spqr} introduces a new compressed format for efficient LLM quantization, which can enable near-lossless compression of LLMs across various model scales while maintaining comparable compression levels to previous quantization methods. Specifically, SpQR first identifies and isolates the outlier weights that may cause particular-large quantization errors, after which SpQR stores them in high precision while compression all other weights in 3-4 bits. OS+ \cite{wei2023outlier} features the channel-wise shifting for asymmetry and the channel-wise scaling for concentration since these operations can be seamlessly migrated into the subsequent quantization modules while maintaining strict equivalence. OS+ also introduces a fast and stable scheme to calculate effective shifting and scaling values, which can further achieve better quantization burden balance towards better quantization performance.

Furthermore, OWQ \cite{lee2023owq} introduces an efficient outlier-aware weight quantization strategy, which aims to minimize LLM's memory footprint through low-bit quantization. OWQ prioritizes a small subset of structured weights that are sensitive to quantization and stores them in higher bits, while applying highly tuned quantization to the remaining dense weights. QuIP \cite{chee2024quip} introduces quantization with incoherence process, which consists of two independent stages, including (1) an adaptive rounding stage to minimize the pre-defined quadratic proxy objective and (2) an efficient pre- and post-processing stage to ensure weight and Hessian incoherence via multiplication by random orthogonal matrices. OmniQuant \cite{shao2023omniquant} introduces an omnidirectionally calibrated quantization technique for LLMs, which consists of two novel components including learnable weight clipping (LWC) and learnable equivalent transformation (LET). LWC modulates the extreme weight values by optimizing the clipping threshold and LET eliminates the activation outliers by shifting the challenge of quantization from activations to weights. Both LWC and LET can be seamlessly integrated into an effective differentiable optimization framework featuring block-wise error minimization for both weight-only and weight-activation quantization. \cite{li2024evaluating} further dives deeper into LLM quantization and analyzes the effect of LLM quantization with comprehensive experiments to evaluate current state-of-the-art LLM quantization techniques, which systematically summarizes the effect of LLM quantization, provides recommendations to apply LLM quantization techniques, and points out future directions of LLM quantization. In contrast to the above LLM quantization works that focus on efficient LLM quantization algorithms, OliVe \cite{guo2023olive} presents an algorithm/architecture co-designed solution to explore efficient quantized LLMs, which features an outlier-victim pair (OVP) quantization scheme and handles outlier values locally with low hardware overheads and high performance gains. This enables an efficient hardware-aligned OVP encoding scheme, which can be integrated into existing hardware accelerators (e.g., systolic arrays and tensor cores) towards more efficient quantized LLMs for generative inference.

\subsubsection{Efficient LLM Distillation}
\label{sec:efficient-llm-distillation}

Another promising direction is to leverage the pre-trained knowledge from large LLMs to enhance the training or fine-tuning process of small LLMs, which can allow small LLMs to maintain as strong performance as large LLMs while exhibiting superior efficiency. As discussed in \cite{bai2024beyond}, recent LLM distillation methods can be divided into two main categories, including black-box LLM distillation and white-box LLM distillation as follows:
\begin{enumerate}
    \item [(1)]
    \textbf{Black-Box LLM Distillation.}
    In the context of black-box distillation, the teacher LLM's parameters are not available for the student LLM and the student LLM can only see the final output from the teacher LLM. In practice, black-box distillation typically features those commercial LLMs (e.g., GPT-3 \cite{brown2020language} and GPT-4 \cite{openai2023gpt4}) as the teacher and leverages the predictions from the teacher to further enhance the training or fine-tuning process of small student LLMs \cite{wang2022self, peng2023instruction, wu2023lamini, jiang2023lion}. For example, Self-Instruct \cite{wang2022self} first generates a large number of instructions, input, and output sequences from GPT-3 using its APIs, after which Self-Instruct filters the invalid or similar ones before using them to fine-tune the original GPT-3 model. Finally, Self-Instruct can achieve an absolute improvement of 33\% over the original GPT-3 model on Super-NaturalInstructions. Similar to Self-Instruct, \cite{peng2023instruction} uses GPT-4 to first generate rich instruction-following data pairs and then uses the generated data pairs to fine-tune small LLaMA models to improve their performance.
    \item [(2)]
    \textbf{White-Box LLM Distillation.}
    In the context of white-box distillation, the teacher LLM's parameters are available for the student LLM and the student LLM can also see the hidden intermediate output from the teacher LLM. More recently, with the emergence of open-source LLMs, white-box distillation has become more popular and more valuable for the LLM community since the student LLM can potentially benefit from the hidden states of the teacher LLM towards better distillation performance \cite{gu2023minillm, liang2023less, agarwal2023gkd, kim2024token}. Among them, MiniLLM \cite{gu2023minillm} first replaces the forward Kullback-Leibler divergence (KLD) objective with reverse KLD, which can prevent the student LLM from overestimating the low-probability regions of the teacher distribution. Next, MiniLLM introduces an effective optimization approach to learn the above reverse KLD objective, which can enhance the student LLM to generate high-quality responses. TED \cite{liang2023less} presents an effective task-aware layer-wise distillation strategy, which features task-aware filters to align the hidden states of teacher and student at each layer. The above filters can select the knowledge from the hidden states that are useful for target tasks, which can further reduce the knowledge gap between teacher and student LLMs. GKD \cite{agarwal2023gkd} proposes to train the student LLM on its self-generated output sequences along with the feedback from the teacher LLM on such self-generated sequences. In addition, GKD also offers the flexibility to employ alternative loss functions between teacher and student, which can enhance the distillation performance of the student even when the student lacks the expressivity to mimic the teacher's distribution. More recently, \cite{kim2024token} introduces token-scaled logit distillation for quantization-aware training of LLMs, which can effectively mitigate the overfitting issue and also largely enhance the distillation process from the teacher predictions and the ground truths.
\end{enumerate}

\subsection{Efficient LLM Systems}
\label{sec:efficient-llm-system-design}
\vspace{+3pt}

In parallel to the rapid development of efficient LLM algorithms, a plethora of efficient LLM systems and infrastructures have also recently emerged \cite{sheng2023flexgen, wang2023tabi, aminabadi2022deepspeed, wu2023fast, borzunov2022petals, jin2024s, patel2023splitwise, zhong2024distserve, du2024liger}, which further optimize the generative inference efficiency of LLMs from the perspective of efficient system-level implementations. Among them, FlexGen \cite{sheng2023flexgen} features an efficient high-throughput generation engine for running LLMs on one single GPU with limited memory, which can be flexibly configured under various hardware resource constraints by aggregating memory and computation from the GPU, CPU, and disk. After solving a linear programming problem, FlexGen can also search for efficient patterns to store and access tensors. Tabi \cite{wang2023tabi} features an inference system with an efficient multi-level inference engine, which can serve queries using small models and optional LLMs for demanding applications. Tabi is particularly optimized for discriminative models (not generative LLMs) in a serving framework, which uses the calibrated confidence score to determine whether to directly return the accurate results of small models or further re-route them to LLMs. DeepSpeed \cite{aminabadi2022deepspeed} presents a comprehensive system solution for efficient transformer inference, which consists of (1) a multi-GPU inference engine to minimize the runtime latency while maximizing the runtime throughput of both dense and sparse transformers when they fit into aggregate GPU memory and (2) a heterogeneous inference engine that leverages CPU and NVMe memory in addition to GPU memory and computation to enable high inference throughput with large models which do not fit into aggregate GPU memory. FastServe \cite{wu2023fast} presents an efficient distributed inference serving system for efficient LLM inference, which (1) exploits the auto-regressive pattern of LLM inference to enable preemption at the granularity of each output token and (2) explores preemptive scheduling to minimize job completion time with a novel skip-join multi-level feedback queue scheduler. Petals \cite{borzunov2022petals} features an efficient collaborative system for runtime inference and fine-tuning of LLMs through joining the resources of multiple parties. In contrast to concurrent LLM inference engines, Petals also natively exposes hidden states of the served model, allowing to train and share custom model extensions based on efficient fine-tuning schemes.
\vspace{+4pt}

Furthermore, S$^3$ \cite{jin2024s} demonstrates that designing an inference system with prior knowledge of the output sequence can largely increase the runtime inference throughput of LLMs. Therefore, in order to increase the runtime inference throughput of LLMs, S$^3$ proposes to (1) first predict the output sequence length, (2) then schedule generation queries based on the prediction to increase runtime resource utilization and throughput, and (3) finally handle mispredictions. Thanks to the prior knowledge of the output sequence, S$^3$ can achieve much better inference throughput than early LLM systems. Splitwise \cite{patel2023splitwise} proposes to split the two phases of typical LLM inference workloads on to different hardware, which thus allows to use well-suited hardware for each inference phase and provision independent computational resources for each inference phase. This strategy can improve the runtime resource utilization across different hardware. Splitwise also optimizes the state transfer across different hardware using the fast back-plane interconnects in today's GPU clusters to further increase the runtime LLM inference throughput. DistServe \cite{zhong2024distserve} proposes to disaggregate the prefill and decoding computation to enhance the runtime serving performance of LLMs, which assigns the prefill and decoding computation workloads to different GPUs and thus largely eliminates the prefill-decoding interference towards better runtime inference throughput. DistServe also optimizes the above two phases according to the serving cluster's bandwidth to minimize the communication overheads caused by the disaggregation. Liger \cite{du2024liger} features an efficient distributed collaborative inference system for LLMs, which can achieve low inference latency at high throughput on multiple GPUs. In addition, to achieve high parallelism and throughput, Liger also introduces an efficient scheduling strategy to effectively schedule the computation and communication kernels across different input requests onto multiple streams of multiple GPUs.

\subsection{Future Envision}
\label{sec:llm-future-envision}
In this section, we further envision several promising future trends and possible directions in the field of efficient LLMs, which are summarized as follows:
\begin{enumerate}
    \item [(1)]
    \textbf{AutoML for Efficient LLMs.}
    Recent state-of-the-art efficient LLMs are typically built upon manual heuristics, which, despite their efficacy, often require considerable human expertise and engineering efforts. In light of this, one promising future direction is to automatically explore efficient LLMs using automated machine learning (AutoML) techniques \cite{zoph2016neural}. For example, given an efficient LLM, we can leverage AutoML techniques to automatically search for its tailored efficient system implementation towards the optimal on-device inference speedup. Similarly, we can also leverage AutoML techniques to automatically search for its tailored pruning or quantization strategy towards the optimal accuracy-efficiency trade-off. This has the potential to largely push forward the frontier of efficient LLM designs.
    \item [(2)]
    \textbf{Alternative Structures for Efficient LLMs.}
    Recent state-of-the-art LLMs heavily rely on the self-attention mechanism in transformer \cite{vaswani2017attention}, which, however, suffers from quadratic time and memory complexity and greatly slows down the pre-training, inference, and fine-tuning stages of LLMs \cite{keles2023computational}. To tackle this dilemma, several alternative structures have recently emerged (e.g., RWKV \cite{peng2023rwkv}, Mamba \cite{gu2023mamba}, and RetNet \cite{sun2023retentive}), which exhibit optimized computational efficiency and also allow researchers to perform efficient language modeling tasks without transformers. For example, RetNet \cite{sun2023retentive} introduces the recurrent representation to enable low-cost inference, which improves the decoding throughput, runtime latency, and GPU memory without sacrificing the language modeling performance. In light of this, one promising future direction is to explore more efficient alternative structures for LLMs, which may deliver considerable efficiency gains over existing transformer-based LLMs without sacrificing the language modeling performance.
    \item [(3)]
    \textbf{Hardware-Aware Benchmarks for Efficient LLMs.}
    Recent state-of-the-art efficient LLMs are typically optimized in terms of the number of parameters or FLOPs. However, these theoretical complexity metrics cannot accurately reflect the runtime performance on target hardware (e.g., latency and energy). This further makes it challenging to fairly compare different efficient LLMs in terms of their runtime inference efficiency on target hardware. In light of this, one promising future direction is to design hardware-aware benchmarks for efficient LLMs, which may include different hardware performance metrics (e.g., latency and energy) across different hardware systems.
    \item [(4)]
    \textbf{Infrastructures for Efficient LLMs.}
    Recently, there have been a large number of works on efficient LLM compression including LLM pruning and LLM quantization, which, however, often require specialized hardware accelerators and thus cannot achieve realistic on-device inference speedups on modern embedded computing systems. For example, non-structured LLM pruning can remove the less important weights to explore highly sparse LLMs with aggressive compression ratios. However, the resulting sparse LLMs cannot achieve realistic on-device inference speedups due to the irregular network sparsity \cite{ma2023llm}. Another recent work \cite{guo2023olive} has also explored to accelerate quantized LLMs and achieved promising performance. However, these are far from enough for real-world large-scale deployments. Similarly, some other works \cite{xiong2025learning, chen2023mugnoc, chen2022lamp, chen2021marco} also have the potential to optimize the communication overheads towards more efficient LLM deployment. In light of this, one promising future direction is to design specialized software and hardware infrastructures to further optimize LLMs for efficient on-device inference.
\end{enumerate}
 
\section{Deep Learning Frameworks for Embedded Computing Systems}
\label{sec:deep-learning-frameworks-for-embedded-computing-systems}

In the past few years, DNNs have been achieving tremendous success in a myriad of real-world intelligent embedded computing scenarios, such as on-device speech recognition \cite{park2018fully, he2020real}, object detection and tracking \cite{xu2019dac, zhang2020skynet}, autonomous vehicles \cite{baidya2020vehicular, zeng2022enabling}, etc. In the meantime, a series of customized software \cite{tensorflow2015-whitepaper, paszke2019pytorch, jia2014caffe, chen2015mxnet, ketkar2017introduction, thakkar2019introduction, ma2019paddlepaddle, bigdl-1} and hardware frameworks \cite{nvidia-jetson, intel-ncs, google-edgetpu, google-coral, hikey-970, orange-pi} have also been developed to facilitate the deployment of DNNs on embedded computing systems. Therefore, we, in this section, further discuss recent popular deep learning software and hardware frameworks that bring deep learning to embedded computing systems to embrace ubiquitous embedded intelligence.

\begin{table}[t]
\resizebox{\textwidth}{!}{%
\begin{tabular}{l|c|c|c|c|c|c}
\toprule[0.175em]
Software & Created by & Year & Programming Languages & Computation Graph & Training & Maintenance \\ \hline
TensorFlow \cite{tensorflow2015-whitepaper} & Google & 2015 & \makecell[c]{Python, C++, \\ Java, and JavaScript} & Static and Dynamic &  \ding{51} & \ding{51}  \\ \hline
PyTorch \cite{paszke2019pytorch} & \makecell[c]{Facebool \\ (now Meta)}& 2016 & Python and C++ & Dynamic & \ding{51} & \ding{51}   \\ \hline
Caffe \cite{jia2014caffe} & Berkeley & 2014 & C++ & Static & \ding{51} & \ding{55}            \\ \hline
MXNet \cite{chen2015mxnet} & Amazon & 2015 & \makecell[c]{Python, C++, R, Java, \\ Julia, JavaScript, \\ Scala, Go, and Perl} & Static and Dynamic &  \ding{51}  & \ding{55}           \\ \hline
Keras \cite{ketkar2017introduction} & Personal & 2015 & Python & Static and Dynamic & \ding{51}  & \ding{51}           \\
\hline
CoreML \cite{thakkar2019introduction} & Apple & 2017 & \makecell[c]{Python, Swift, \\ and Objective-C} & Static &   \ding{55} & \ding{51}          \\ \hline
PaddlePaddle \cite{ma2019paddlepaddle} & Baidu & 2016 & \makecell[c]{Python and C++} & Static and Dynamic & \ding{51}  & \ding{51}         \\ \hline
BigDL \cite{bigdl-1} & Intel & 2017 & Python and Scala &   Dynamic  &  \ding{51} & \ding{51}      \\
\toprule[0.175em]
\end{tabular}%
}
\caption{Illustration of deep learning software frameworks that have been discussed in Section~\ref{sec:deep-learning-software-frameworks}. Note that we refer to the framework under active maintenance if there are new releases with the previous six months.}
\vspace{-5pt}
\label{tab:deep-learning-software}
\end{table}

\subsection{Deep Learning Software Frameworks}
\label{sec:deep-learning-software-frameworks}

In this section, we introduce popular deep learning software frameworks that have been widely used to develop deep learning solutions for embedded computing systems, including TensorFlow \cite{tensorflow2015-whitepaper}, PyTorch \cite{paszke2019pytorch}, Caffe \cite{jia2014caffe}, MXNet \cite{chen2015mxnet}, Keras \cite{ketkar2017introduction}, CoreML \cite{thakkar2019introduction}, PaddlePaddle \cite{ma2019paddlepaddle}, and BigDL \cite{bigdl-1}. The covered deep learning software frameworks are summarized in Table~\ref{tab:deep-learning-software}.

\textbf{TensorFlow} \cite{tensorflow2015-whitepaper} is an open-source deep learning software framework developed by Google, which is released in 2015 and has since become one of the most popular deep learning software frameworks for training and deploying DNNs. In practice, TensorFlow, especially TensorFlow Lite, allows developers to easily build and deploy DNNs in a wide range of embedded computing systems, including mobile phones, microcontrollers (MCUs), Raspberry Pi, TPUs, and edge GPUs. In the meantime, TensorFlow also supports various real-world applications, ranging from image and speech recognition to natural language processing and predictive analytics. With its flexible architecture and vast pre-trained models, TensorFlow has been deemed as one of the most important tools for researchers and developers in the field of deep learning.

\textbf{PyTorch} \cite{paszke2019pytorch} is an open-source deep learning software framework that is widely used for training and deploying deep neural networks. It is developed by Facebook (now known as Meta) and is released in 2016. One of the key features is the dynamic computation graph, which allows developers to change the computation behavior of DNNs on the fly. This feature distinguishes PyTorch from other deep learning software frameworks (e.g., TensorFlow) that only support static computation graph. In addition, PyTorch has a number of high-level features that make it easier to build more complex DNNs. For example, the TorchVision package provides various useful tools and pre-trained models for image and video processing. With its dynamic computation graph, ease of use, and range of high-level features, PyTorch has become an essential software framework for training and deploying DNNs in the deep learning community.

\textbf{Caffe} \cite{jia2014caffe} is a popular deep learning software framework developed by Berkeley and released in 2014, which has gained increasing popularity due to its speed, modularity, and ease of use. One of the key features is its ability to deal with large datasets with millions of images. Another important feature is its modularity, which allows developers to add or remove components with ease. Besides, Caffe includes a large library that contains hundreds of pre-trained models, which can be used to quickly build deep learning applications, such as image classification, object detection, and segmentation. In addition to its powerful features, Caffe has a user-friendly interface, allowing developers to train and deploy DNNs without extensive knowledge of deep learning. With its powerful features and user-friendly interfaces, Caffe has become an invaluable tool for researchers and developers and inspired subsequent deep learning software frameworks.

\textbf{MXNet} \cite{chen2015mxnet} is an open-source deep learning software framework to train and deploy DNNs, which is developed by Amazon and released in 2015. One of the key technical merits is its distributed training capability, which allows to train DNNs across multiple computation nodes, and more importantly, in a computationally efficient manner. Besides, MXNet also supports multiple programming languages, such as Python, C++, R, and Julia, which further increases the accessibility to researchers and developers with diverse skill levels. Apart from these, MXNet's integration with other deep learning software frameworks and tools, such as Apache Spark and Apache Flink, is another important feature that facilitates the integration of deep learning into existing data processing pipelines. Thanks to its scalability, flexibility, and efficiency, MXNet has become a popular option for developers and researchers in the deep learning community.

\textbf{Keras} \cite{ketkar2017introduction} is an open-source deep learning software framework written in Python, which provides high-level APIs for building and training efficient DNN solutions. It is developed by François Chollet in 2015 and is now maintained by a community of developers. In particular, Keras has been integrated into TensorFlow, and starting from TensorFlow 2.0, Keras has become the default API to build DNN solutions in TensorFlow. Specifically, one of the key features of Keras is its modularity, which allows developers and researchers to easily construct and customize DNNs. Furthermore, Keras also allows users to productize DNN solutions on modern mobile phones like iOS and Android, on the web, or on the Java virtual machine. Last but not least, Keras allows to train DNN solutions in an efficient distributed manner on clusters of multiple GPUs and TPUs. The aforementioned strengths make Keras increasingly popular in both industry and academia.

\textbf{CoreML} \cite{thakkar2019introduction} is an open-source deep learning software framework developed by Apple in 2017, which aims to integrate DNNs into Apple commercial products, such as iPhone, iPad, and Apple Watch. In addition to supporting extensive DNNs with over 30 layer types, CoreML also covers standard machine learning models, such as tree ensembles, support vector machine, and generalized linear models. Furthermore, another key feature of CoreML is its ability to directly run DNNs on the device, without the need for cloud-based inference. Besides, CoreML provides a range of optimization techniques, such as quantization and pruning, to reduce the complexity of DNNs. It is worth noting that this is particularly important for modern mobile devices, which typically have limited storage and computational resources. Also, CoreML, built on top of advanced technologies like Metal and Accelerate, seamlessly takes advantage of CPUs and GPUs to provide the maximum inference performance at run time. These technical features make CoreML the first choice for developing efficient DNN solutions on Apple commercial products.

\textbf{PaddlePaddle} \cite{ma2019paddlepaddle}, also known as Paddle, is an open-source deep learning software framework developed by Baidu, which has been released to benefit the deep learning community since 2016. Specifically, PaddlePaddle is designed to be an industrial platform with advanced technologies and rich features that cover core deep learning frameworks, basic model libraries, end-to-end tools, and service platforms. In particular, PaddlePaddle is originated from industrial practices with dedication and commitment to industrialization. It has been adopted by a wide range of sectors, including manufacturing, agriculture, and enterprise service. With the industrial benefits, PaddlePaddle has motivated an increasing number of developers and researchers to commercialize AI.

\textbf{BigDL} \cite{bigdl-1} is an open-source deep learning software framework that runs on top of Apache Spark. It is developed by Intel and released in 2017. The goal of BigDL is to provide a high-performance, scalable, and easy-to-use platform, especially distributed deep learning. To this end, BigDL includes a comprehensive set of features that cover various deep learning applications, including image classification, object detection, and natural language processing. One of the key features of BigDL is its ability to take full advantage of distributed computing resources, such as CPU, GPU, and FPGA clusters, to accelerate the training of DNNs. Besides, BigDL is seamlessly integrated with Apache Spark, which enables users to leverage the distributed computing capability of Spark for data preprocessing and postprocessing. This integration also makes it possible to build end-to-end deep learning pipelines that span from data ingestion to model deployment.

\subsection{Deep Learning Hardware Frameworks}
\label{sec:deep-learning-hardware-frameworks}

In this section, we introduce popular embedded hardware platforms that are designed to run powerful DNNs in embedded scenarios without cloud-based assistance, including Nvidia Jetson \cite{nvidia-jetson}, Intel Neural Compute Stick \cite{intel-ncs}, Google Edge TPU \cite{google-edgetpu}, Google Coral Dev Board \cite{google-coral}, Huawei HiKey 970 \cite{hikey-970}, and Orange Pi AI Stick Lite \cite{orange-pi}. The covered deep learning hardware frameworks are summarized in Table~\ref{tab:deep-learning-hardware}.

\begin{table}[t]
\resizebox{\textwidth}{!}{%
\begin{tabular}{l|c|c|c|c|c|c}
\toprule[0.175em]
Hardware & RAM & Storage & Power & Performance & Price  & \makecell[c]{Supported Deep \\ Learning Software} \\ \hline
Nvidia Jetson TX2 \cite{nvidia-jetson} & 8\,GB LPDDR4 & 32\,GB eMMC 5.1 & 7.5\,W $\sim$ 15\,W & 1.33\,TFLOPS & \$399 & \makecell[c]{TensorFlow, PyTorch, \\ Caffe, Keras, and MXNet} \\ \hline
Nvidia Jetson Nano \cite{nvidia-jetson} & 4\,GB LPDDR4 & 16\,GB eMMC 5.1 & 5\,W $\sim$ 10\,W & 0.472\,TFLOPS & \$99 & \makecell[c]{TensorFlow, PyTorch, \\ Caffe, Keras, and MXNet} \\ \hline
Nvidia Jetson AGX Xavier \cite{nvidia-jetson} &  32\,GB LPDDR4x & 32\,GB eMMC 5.1 & 10\,W $\sim$ 30\,W & 32\,TOPS & \$1,099 & \makecell[c]{TensorFlow, PyTorch, \\ Caffe, Keras, and MXNet} \\ \hline
Nvidia Jetson Xavier NX \cite{nvidia-jetson} &  8\,GB LPDDR4x & 16\,GB eMMC 5.1 & 10\,W $\sim$ 20\,W & 21\,TOPS & \$399 & \makecell[c]{TensorFlow, PyTorch, \\ Caffe, Keras, and MXNet} \\ \hline
Nvidia Jetson AGX Orin \cite{nvidia-jetson} &  32\,GB LPDDR5 & 64\,GB eMMC 5.1 & 15\,W $\sim$ 40\,W & 275\,TOPS & \$1,999 & \makecell[c]{TensorFlow, PyTorch, \\ Caffe, Keras, and MXNet} \\ \hline
Nvidia Jetson Orin NX \cite{nvidia-jetson} &  16\,GB LPDDR5 & 32\,GB eMMC 5.1 & 10\,W $\sim$ 25\,W & 100\,TOPS & \$599 & \makecell[c]{TensorFlow, PyTorch, \\ Caffe, Keras, and MXNet} \\ \hline
Intel Neural Compute Stick \cite{intel-ncs} & N/A & N/A & 0.5\,W $\sim$ 1.5\,W & 0.1\,TFLOPS & \$79 & \makecell[c]{TensorFlow, Caffe, \\ and MXNet} \\ \hline
Google Edge TPU \cite{google-edgetpu} & N/A & N/A & 2\,W & 4\,TOPS & \$75 & \makecell[c]{TensorFlow and \\ TensorFlow Lite} \\ \hline
Google Coral Dev Board \cite{google-coral} & 1\,GB LPDDR4 & 8\,GB eMMC 5.1 & 1\,W $\sim$ 6\,W & 4\,TOPS & \$149 & \makecell[c]{TensorFlow and \\ TensorFlow Lite} \\ \hline
Huawei HiKey 970 \cite{hikey-970} & 6\,GB LPDDR4 & 64\,GB UFS 2.1 & 6\,W $\sim$ 12\,W & 1.88\,TOPS & \$299 & \makecell[c]{TensorFlow, PyTorch, \\ and Caffe} \\ \hline
Orange Pi AI Stick Lite \cite{orange-pi} & N/A & N/A & 1\,W $\sim$ 2\,W & 4\,TOPS & \$69 & \makecell[c]{TensorFlow, PyTorch, \\ and Caffe} \\ 
\toprule[0.175em]
\end{tabular}%
}
\caption{Illustration of deep learning software frameworks that have been discussed in Section~\ref{sec:deep-learning-hardware-frameworks}. Note that the price here refers to the initial price during the product launch, which is subject to fluctuations over time.}
\vspace{-5pt}
\label{tab:deep-learning-hardware}
\end{table}

\textbf{Nvidia Jetson} \cite{nvidia-jetson} is a series of embedded system-on-modules (SoMs) designed by Nvidia for running advanced deep learning workloads, especially the inference of DNNs. Specifically, Nvidia Jetson consists of Nvidia Jetson TX2, Nvidia Jetson Nano, Nvidia Jetson AGX Xavier, Nvidia Jetson Xavier NX, Nvidia Jetson AGX Orin, and Nvidia Jetson Orin NX. To accelerate deep learning workloads, Nvidia Jetson runs on top of Nvidia's CUDA parallel computing architectures and features an integrated system-on-chip (SoC) with a powerful Nvidia GPU, a multi-core CPU, and various high-speed interfaces, including Ethernet, USB, HDMI, and CSI/DSI. More importantly, Nvidia Jetson is compatible with various deep learning software frameworks, including TensorFlow, PyTorch, Caffe, Keras, and MXNet. Thanks to its advanced architecture design and powerful interfaces, Nvidia Jetson is able to support a wide range of embedded deep learning applications to accommodate different resource and performance requirements.

\textbf{Intel Neural Compute Stick} \cite{intel-ncs} is a small, low-power, and cost-effective embedded hardware designed to run deep learning workloads without cloud-based assistance, which is developed by Movidius (now acquired by Intel). Neural Compute Stick (NCS) is a small USB device that can be connected to a host computer or embedded computing system. Specifically, NCS features the Myriad 2 Vision Processing Unit (VPU), which is optimized for the inference of DNNs. Besides, NCS is integrated with various high-speed interfaces, including USB 3.0 and Wi-Fi. Meanwhile, developers can use the Intel Movidius SDK, which provides a set of tools for developing, testing, and deploying DNNs on NCS. Furthermore, NCS supports various deep learning software frameworks, including TensorFlow, Caffe, and MXNet. Thanks to its significant flexibility and cost efficiency, NCS makes it possible to deploy advanced DNNs in a wide range of embedded scenarios.

\textbf{Google Edge TPU} \cite{google-edgetpu} is a custom-built ASIC chip to accelerate deep learning workloads on resource-constrained edge computing systems. Specifically, Google Edge TPU is designed to seamlessly work together with TensorFlow Lite, a lightweight version of TensorFlow, and is optimized for the inference of DNNs towards enhanced inference efficiency. It is worth noting that Google Edge TPU itself cannot work alone, and similar to Intel Neural Compute Stick, it must be connected to other embedded computing systems, such as Raspberry Pi 4 and Google Coral Dev Board, to deliver deep learning solutions. In particular, Google Edge TPU is capable of performing up to two trillion floating-point operations per second (TFLOPS) and four trillion operations per second (TOPS) using only two watts of power. This further allows us to build and deploy powerful deep learning solutions on embedded computing systems with limited computational resources. Thanks to its easy integration with other embedded computing systems, powerful performance, and significant efficiency, Google Edge TPU has gained increasing popularity in the deep learning community for deploying deep learning solutions on embedded computing systems.

\textbf{Google Coral Dev Board} \cite{google-coral} is a single-board computer designed for building embedded deep learning applications. Specifically, it features an on-board Google Edge TPU, which is a custom-built chip to run TensorFlow Lite models with high performance and low power consumption. The Coral Dev Board has various built-in interfaces, including Audio, Wi-Fi, Bluetooth, Ethernet, and USB 3.0, which enable it to be connected to other embedded computing systems. Besides, the Coral Dev Board is integrated with 1\,GB LPDDR4 RAM, 8\,GB eMMC 5.1 flash memory, and a MicroSD slot for additional storage. The Coral Dev Board also comes with pre-installed software tools, including TensorFlow Lite, Google Edge TPU API, and various sample applications. These allow users to easily and quickly start building their deep learning solutions. Thanks to its powerful Google Edge TPU, various useful interfaces, and software tools, the Coral Dev Board has become increasingly popular for developing and deploying embedded deep learning solutions.

\textbf{Huawei HiKey 970} \cite{hikey-970} is a high-performance single-board embedded computer designed by Huawei. Specifically, the HiKey 970 features a powerful neural processing unit (NPU) to accelerate various deep learning workloads. The HiKey 970 is also integrated with 6\,GB LPDDR4 LPDDR4 RAM and 64\,GB UFS 2.1 flash memory, while at the same time allowing MicroSD extension for additional storage. Besides, the HiKey 970 supports various high-speed interfaces, including Ethernet, USB 3.0, and PCIe 3.0. Furthermore, the HiKey 970 is compatible with popular deep learning software frameworks, such as TensorFlow, PyTorch, and Caffe, allowing users to easily and quickly build on-board deep learning solutions. Thanks to its powerful NPU, rich memory and storage, and extensive connectivity options, the HiKey 970 is suitable for a wide range of intelligent embedded applications, such as robotics, autonomous vehicles, and smart home devices.

\textbf{Orange Pi AI Stick Lite} \cite{orange-pi} is a tiny and cost-effective USB stick designed for small to medium-sized deep learning workloads. It is equipped with a single-core Cortex-A7 processor and a neural processing unit (NPU) that provides hardware acceleration. Note that, similar to the Intel Neural Compute Stick and Google Edge TPU, the Orange Pi AI Stick Lite cannot work alone, which must be connected to a host device using the on-device USB 3.0 interface. Furthermore, the Orange Pi AI Stick Lite supports various deep learning software frameworks, including TensorFlow, PyTorch, and Caffe. Thanks to its cost efficiency, the Orange Pi AI Stick Lite is suitable for embedded computing systems to deal with small to medium-sized deep learning workloads.

\subsection{Future Envision}
In this section, we envision the future trends and possible directions of deep learning software and hardware infrastructures, which are summarized as follows:
\begin{itemize}
    \item[(1)] {
    \textbf{Integration with Emerging Technologies.}
    In the future, we should consider to develop deep learning software and hardware, which can be seamlessly integrated with emerging technologies. For example, quantum computing \cite{wang2022quantumnat, wang2022quest, wang2022qoc} has the potential to deliver significant speedup and computational capability, which can accelerate the training and inference of DNNs. Therefore, it is of paramount importance to explore the potential of integrating deep learning software and hardware with emerging technologies to unlock new possibilities and new advances in various real-world scenarios.
    }
    \item[(2)] {
    \textbf{Democratization of Deep Learning.}
    The democratization of deep learning \cite{ahmed2020democratization} has emerged as a prominent trend in the deep learning era, with the explicit goal of making deep learning software and hardware more accessible to a wider range of developers and researchers. Therefore, in order to alleviate the technical barrier to entry for building efficient embedded deep learning solutions, we, in the future, should continue to develop more user-friendly deep learning software and hardware frameworks, democratizing the benefits and advances of deep learning in real-world embedded scenarios.
    }
    \item[(3)] {
    \textbf{Development of Specialized Hardware.}
    Conventional embedded computing systems typically focus on optimizing and accelerating the training and inference of traditional convolutional networks, neglecting recent advances in the deep learning era. Among them, Vision Transformer (ViT) \cite{dosovitskiy2020image} is the most representative one, which has opened up a new direction and has been challenging the dominant role of traditional convolutional networks in various real-world vision applications, such as image classification \cite{dosovitskiy2020image, liu2021swin, liu2022swin}, object detection \cite{li2022exploring, fang2021yolos, minderer2022simple}, semantic segmentation \cite{strudel2021segmenter, chen2021transunet, gu2022multi, kirillov2023segmentanything}, and video analysis \cite{liu2022video, arnab2021vivit, neimark2021video}. Therefore, in order to unleash the promise of ViT and its variants, we should also develop specialized embedded computing systems to accelerate the family of ViT rather than only focusing on accelerating complicated convolutional networks.
    }
    \item[(4)] {
    \textbf{Development of More Powerful Hardware.}
    As seen in recent advanced DNNs \cite{liu2022convnet, woo2023convnext, liu2022more}, the network complexity has continued to explode, and as a result, continues to enlarge the computational gap between computation-intensive DNNs and resource-constrained embedded computing systems. In parallel, the large language models (LLMs), such as ChatGPT \cite{openai2023gpt4}, have been achieving impressive success in various neural language processing (NLP) tasks, such as language generation, language translation, question answering, etc., at the cost of pushing forward the network complexity to another unseen level, significantly enlarging the computational gap. These further demonstrate the necessity of innovating more powerful yet cost-effective embedded computing systems to further bridge the aforementioned computational gap, especially from the hardware perspective.
    }
    \item[(5)] {
    \textbf{Development of Infrastructures for On-Device Training.}
    In the past, the convention in the deep learning community is to (1) first train DNNs on powerful GPUs or remote cloud and (2) then deploy the pre-trained DNNs on local embedded computing systems for further inference at run time. Compared with this convention, the emerging paradigm of on-device training enables the pre-trained DNNs to adapt to the new data collected from the local sensors or by the users \cite{lin2022device}. As such, the users can benefit from customized DNNs without having to transfer the collected data to the remote cloud, thereby significantly protecting the data privacy and security \cite{lin2022device}. Nonetheless, conventional embedded computing systems are typically optimized for inference and do not support efficient on-device training due to the training memory bottleneck during the training process \cite{cai2020tinytl, lin2022device}. This motivates us to further develop efficient infrastructures, including specialized deep learning software and hardware, to effectively accommodate future on-device training demands.
    }
\end{itemize}

\section{Deep Learning Applications for Embedded Computing Systems}
\label{sec:deep-learning-applications-for-embedded-computing-systems}

In the previous sections, we have extensively discussed recent advances towards ubiquitous embedded intelligence from various perspectives of efficient deep learning networks, algorithms, software, and hardware. In this section, we further elaborate on recent popular intelligent deep learning applications in real-world embedded scenarios, spanning from vision to NLP tasks. Note that these intelligent embedded applications highly rely on efficient deep networks and efficient deep learning algorithms that have been extensively discussed in the previous sections.

\subsection{Computer Vision Applications}
\label{sec:computer-vision-applications}

Computer vision is an emerging field that focuses on interpreting and understanding visual information from real-world environments, such as image and video, spanning from image classification \cite{simonyan2014vggnet} to downstream vision tasks, such as object detection \cite{liu2016ssd}, tracking \cite{kristan2015visual}, and segmentation \cite{li2014secrets}. Below we discuss recent popular intelligent embedded vision applications.

\newlength{\columnsepsavesavesavesavesavesavesavesave} 
\setlength{\columnsepsavesavesavesavesavesavesavesave}{\columnsep} 
\setlength{\columnsep}{6pt} 
\begin{wrapfigure}{r}{0.525\columnwidth}
    \begin{center}
        \includegraphics[width=0.525\columnwidth]{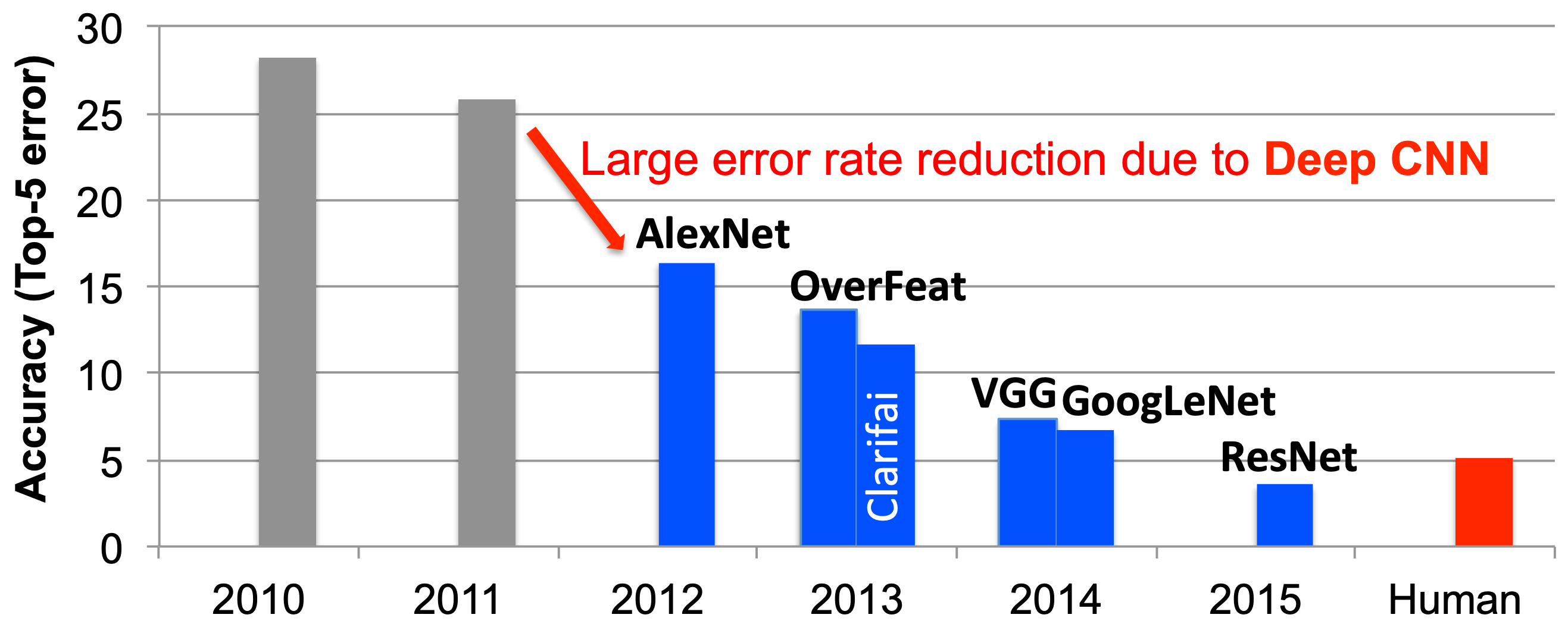}
    \end{center}
    \caption{Milestones of early convolutional networks \cite{sze2017efficient}.}
    \label{fig:cnn-milestones}
\end{wrapfigure}
\textbf{Image Classification.}
Image classification, also referred to as image recognition, is the most fundamental vision task, which focuses on recognizing the input image based on its visual information \cite{simonyan2014vggnet}. This enables various intelligent applications in real-world embedded computing systems, such as mobile phones and IoT sensors, allowing these embedded computing systems to automatically recognize objects, scenes, or patterns within the given image \cite{cai2019device}. For example, face recognition \cite{boragule2023device}, person re-identification \cite{luo2020person}, and hand gesture recognition \cite{sung2021device} have been widely integrated into mainstream embedded computing systems, such as mobile phones, ATMs, and intelligent cameras, for the purpose of identity authentication. In practice, image classification typically features deep convolutional networks, such as VGGNet \cite{simonyan2014vggnet}, ResNet \cite{he2016deep}, and DenseNet \cite{huang2017densely}, thanks to their strong capabilities to capture rich visual information, especially for large-scale datasets like ImageNet \cite{deng2009imagenet}. For example, as shown in Fig.~\ref{fig:cnn-milestones}, AlexNet \cite{krizhevsky2012alexnet}, as the first of its kind, demonstrates the possibility of leveraging convolutional layers to learn discriminative features from vision inputs, which exhibits significantly better recognition performance on ImageNet than previous well-established non-convolutional networks, such as multi-layer perceptrons (MLPs) and other learning-based techniques. Furthermore, ResNet \cite{he2016deep} investigates the training collapse of deep convolutional networks and introduces a simple yet effective deep residual learning paradigm, which allows us to significantly increase the network depth for stronger learning capabilities and also marks the booming development of the deep learning era. As a result, ResNet, for the first time, achieves better recognition performance on ImageNet than humans thanks to its significant network depth as shown in Fig.~\ref{fig:cnn-milestones}.
\setlength{\columnsep}{\columnsepsavesavesavesavesavesavesavesave}

\textbf{Downstream Vision Applications.}
Downstream vision applications typically refer to the practical and specific usage, where the results or outputs from other fundamental vision tasks, such as image classification, are applied to deal with real-world challenges. Popular downstream vision applications in practice include but not limited to object detection \cite{liu2016ssd}, object tracking \cite{kristan2015visual}, object segmentation \cite{li2014secrets}, image super-resolution \cite{liu2021evsrnet, wu2022compiler}, image restoration \cite{shi2023memory}, pose estimation \cite{grishchenko2022blazepose}, image captioning \cite{yao2017boosting, you2016image}, augmented reality (AR) and virtual reality (VR) \cite{fu2023gen}, and video-related analysis \cite{tang2019coin}. Among them, \cite{shi2023memory} features memory-oriented structured pruning to optimize the on-device memory consumption during runtime image restoration, which can accommodate the limited memory and storage requirements in real-world embedded scenarios. These downstream vision applications have evolved to be ubiquitous in real-world embedded scenarios, which serve as important components towards ubiquitous embedded intelligence. For example, object detection and tracking have been widely used in recent autonomous vehicles \cite{schwarting2018planning} to detect other vehicles and surveillance systems \cite{elharrouss2021review} to detect suspicious persons or activities. These downstream vision applications have also been widely applied to other real-world embedded scenarios, such as smart cities and intelligent healthcare \cite{wang2022development}. To further facilitate the development of intelligent applications, several powerful tools have been recently proposed. For example, Precog \cite{drolia2017precog} introduces an efficient object detection infrastructure to enable real-time object detection on resource-constrained embedded computing systems, such as Raspberry Pi, which also features YOLOv3 \cite{redmon2018yolov3} to achieve superior on-device object detection accuracy.

\textbf{From CNNs to Vision Transformers.}
More recently, vision transformers (ViTs) \cite{dosovitskiy2020image} and its variants have demonstrated surprisingly strong performance in various vision tasks, including but not limited to image classification \cite{dosovitskiy2020image, liu2021swin, liu2022swin}, object detection \cite{li2022exploring, fang2021yolos, minderer2022simple}, semantic segmentation \cite{strudel2021segmenter, chen2021transunet, gu2022multi, kirillov2023segmentanything}, and video analysis \cite{liu2022video, arnab2021vivit, neimark2021video}, which continue to push forward the state-of-the-art performance over their convolutional counterparts across various vision tasks. Specifically, \cite{dosovitskiy2020image}, as the very first vision transformer, proposes to divide the input image into a series of smaller image patches (e.g., 8, 16, and 32 patches), each of which is then fed into the transformer-based encoder to learn discriminative features. The learned discriminative features are further aggregated and fed into the classification layer to make predictions as shown in Fig.~\ref{fig:vit-overview}. However, despite their strong performance across various vision tasks, ViTs and its variants often exhibit inferior on-device efficiency \cite{han2022vision} since they are typically more difficult to parallelize on resource-constrained embedded computing systems than their convolutional counterparts and thus inevitably suffer from considerable resource underutilization as pointed out in \cite{maaz2023edgenext}. To overcome such limitations, a plethora of resource-efficient vision transformers have recently flourished and we refer the interested readers to Section~\ref{sec:manual-transformers} for more details about recent representative resource-efficient vision transformers. We emphasize that significant efforts are still required in order to further alleviate the on-device efficiency bottleneck and also unleash the promise of modern vision transformers, which are of paramount importance to bring powerful vision transformers to the less capable embedded computing systems towards ubiquitous embedded intelligence.

\subsection{Natural Language Processing Applications}
\label{sec:natural-language-processing-applications}

In parallel to vision tasks, natural language processing (NLP) is another representative application that has been widely deployed in real-world embedded scenarios to explore auditory and textual inputs, which has largely revolutionized how embedded computing systems interact with users and their surroundings \cite{anwar2020natural}. In practice, embedded computing systems ranging from traditional IoT systems to wearable systems, and autonomous systems are transitioning from simple responsive systems to more proactive and interactive systems, which can comprehend context and also anticipate users' needs based on their linguistic inputs. To this end, below we introduce several representative NLP applications in real-world embedded scenarios.
\begin{itemize}
    \item[(1)]
    \textbf{Sentiment Analysis.}
    Recent intelligent embedded computing systems, such as wearable devices and intelligent healthcare infrastructures, have largely featured sentiment analysis, which can effectively capture users' physiological status through language interactions \cite{zhou2021sentiment}. This also allows more comprehensive understanding of users' emotional well-being, which paves the way for future holistic health ecosystem solutions \cite{shah2020mental}.
    \item[(2)]
    \textbf{Automatic Speech Recognition.}
    Automatic speech recognition has gained increasing interest in real-world embedded scenarios, such as autonomous vehicles and smart homes, which can largely facilitate complicated functions using vocal commands. This can mitigate manual interactions and also enhance safety and user convenience \cite{dong2020rtmobile, raj2022reduced}.
    \item[(3)]
    \textbf{Conversational Agents.}
    The conversational agents have been playing an important role in recent intelligent embedded computing systems, such as home automation systems \cite{cho2019once} and interactive assistant systems \cite{siri}. These intelligent conversational agents maintain strong abilities to comprehend and interpret users' commands, preferences, and behavioral patterns towards better intelligent services in subsequent interactions.
    \item[(4)]
    \textbf{Speech-to-Text/Text-to-Speech Synthesis.}
    The integration of text-to-speech (TTS) \cite{luo2021lightspeech} and speech-to-text (STT) \cite{wang2020fairseq} marks an important milestone in enriching human-computer interactions, especially for wearable systems, such as mobile phones and intelligent translation devices, which can largely facilitate human-computer interactions. Specifically, TTS can synthesize digital text into speech to provide auditory feedback to users and vice versa for STT, both of which are particularly important in hands-free environments.
    \item[(5)]
    \textbf{Real-Time Translation.}
    The emergence of real-time translation has served as an effective technique to eliminate cross-language barriers. More recently, real-time translation has been widely integrated into real-world embedded scenarios, especially wearable communication devices, which can largely facilitate cross-language interactions \cite{hou2019signspeaker, ren2020simulspeech}.
\end{itemize}
To summarize, the integration of NLP and embedded computing systems is more than simple technical enhancements. Instead, it is an important paradigm shift towards ubiquitous embedded intelligence. It can enable real-world embedded computing systems to understand and interpret not only short commands but also longer contexts and conversations, which can further ensure seamless and enriched interfaces between humans and embedded computing systems.

\subsection{Future Envision}
In this section, we envision some future trends and possible directions of intelligent embedded applications, which are summarized as follows:
\begin{itemize}
    \item[(1)] 
    \textbf{LLMs-Enabled Embedded Applications.}
    Large language models (LLMs) starting from GPT-3 \cite{brown2020language} have attracted considerable interest from both academia and industry, thanks to their surprisingly strong performance across various language tasks. Among them, ChatGPT \cite{openai2020chatgpt}, as one of the most representative LLMs-enabled applications, has achieved promising performance improvement over humans across diverse domains of knowledge. Nonetheless, modern LLMs, despite their promise, require a huge amount of computational resources for both training and inference, making it challenging to deploy powerful LLMs on resource-constrained embedded computing systems. Therefore, modern LLMs can only be deployed on remote GPU servers and provide remote services to local users through network connectivity. This, however, is often less convenient and also involves data security/privacy concerns. To overcome such limitations, a plethora of works have been recently proposed to compress computation-intensive LLMs towards better on-device inference efficiency. For example, SmoothQuant \cite{xiao2023smoothquant} and AWQ \cite{lin2023awq} pioneer to quantize the weights of powerful LLMs from higher bits to lower bits in order to reduce their prohibitive computational complexity, making it possible to run powerful LLMs on resource-constrained embedded computing systems. These are also important milestones to bring LLMs to real-world embedded computing systems towards ubiquitous embedded intelligence. 
    \item[(2)]
    \textbf{Multi-Modal Embedded Applications.}
    Modern embedded applications largely focus on one single modality, either from the perspective of vision or language processing. Nonetheless, recent embedded computing systems typically feature various advanced sensors, which can simultaneously collect rich data from multiple modalities, including but not limited to visual, auditory, and tactile information. In practice, the most important benefit of these multi-modal embedded applications is their strong abilities to provide comprehensive understanding of real-world dynamic environments using comprehensive information collected from different modalities. This also has the potential to significantly boost the attainable accuracy on target task and also greatly improve the reliability in real-world dynamic environments. For example, visual information can be easily augmented with other modalities, such as radar and lidar, which can be jointly leveraged to deliver better and safer driving experiences in autonomous vehicles \cite{chowdhuri2019multinet}. However, despite the promising benefits, the development of multi-modal embedded applications is also challenging. On the one hand, the real-time synchronization of diverse data modalities may require significant computational resources. On the other hand, the development of multi-modal embedded applications introduces additional complexity for data alignment, calibration, and fusion, which may also require more advanced software algorithms to ensure real-time processing.
\end{itemize}

\section{Conclusion}
\label{sec:conclusion}

In this survey, we focus on summarizing recent efficient deep learning infrastructures for embedded computing systems towards ubiquitous embedded intelligence, spanning \textbf{from training to inference}, \textbf{from manual to automated}, \textbf{from convolutional neural networks to transformers}, \textbf{from transformers to vision transformers}, \textbf{from vision models to large language models}, \textbf{from software to hardware}, and \textbf{from algorithms to applications}. To this end, we discuss recent efficient deep learning infrastructures for embedded computing systems from the lens of (1) efficient manual network design for embedded computing systems, (2) efficient automated network design for embedded computing systems, (3) efficient network compression for embedded computing systems, (4) efficient on-device learning for embedded computing systems, (5) efficient large language models for embedded computing systems, (6) efficient deep learning software and hardware for embedded computing systems, and (7) efficient intelligent applications for embedded computing systems. Furthermore, we also envision promising future directions and trends to enable more efficient and ubiquitous embedded intelligence. We believe this survey can shed light on future research and allow researchers to quickly and smoothly get started in this emerging field.

\bibliographystyle{unsrt}
\bibliography{reference}

\end{document}